\documentclass[opre,nonblindrev]{informs3} 

\DoubleSpacedXI 

\usepackage{endnotes}
\let\footnote=\endnote

\usepackage{natbib}
 \bibpunct[, ]{(}{)}{,}{a}{}{,}%
 \def\bibfont{\small}%
\TheoremsNumberedThrough     
\ECRepeatTheorems

\EquationsNumberedThrough    

\usepackage{hyperref, bbm, nicefrac, makecell, svg, subcaption}
\usepackage{algorithm, algorithmicx, float}
\usepackage[noend]{algpseudocode}
\def\arginf{\mathop{\rm arg\,inf}}%
\begin{document}


\RUNAUTHOR{Dogan et. al.}

\RUNTITLE{Learning to Incentivize with Unobserved Rewards}

\TITLE{Learning to Incentivize Perfect-Knowledge Agents with Unobserved Rewards}

\ARTICLEAUTHORS{%
\AUTHOR{Ilgin Dogan}
\AFF{Department of Industrial Engineering and Operations Research, University of California, Berkeley, Berkeley, California 94720. Current affiliation: INSEAD, Fontainebleau, France, \EMAIL{ilgin.dogan@insead.edu}} 
\AUTHOR{Zuo-Jun Max Shen}
\AFF{Department of Industrial Engineering and Operations Research, University of California, Berkeley, Berkeley, California 94720, \EMAIL{maxshen@berkeley.edu}}
\AUTHOR{Anil Aswani}
\AFF{Department of Industrial Engineering and Operations Research, University of California, Berkeley, Berkeley, California 94720, \EMAIL{aaswani@berkeley.edu}}
} 

\ABSTRACT{%
Motivated by a number of real-world applications from domains like healthcare and sustainable transportation, in this paper we study a scenario of repeated principal-agent games within a multi-armed bandit (MAB) framework, where: the principal gives a different incentive for each bandit arm, the agent picks a bandit arm to maximize its own expected reward plus incentive, and the principal observes which arm is chosen and receives a reward (different than that of the agent) for the chosen arm. Designing policies for the principal is challenging because the principal cannot directly observe the reward that the agent receives for their chosen actions, and so the principal cannot directly learn the expected reward using existing estimation techniques. As a result, the problem of designing policies for this scenario, as well as similar ones, remains mostly unexplored. In this paper, we construct a policy that achieves a low regret (i.e., square-root regret up to a log factor) in this scenario for the case where the agent has perfect-knowledge about its own expected rewards for each bandit arm.  We design our policy by first constructing an estimator for the agent's expected reward for each bandit arm. Since our estimator uses as data the sequence of incentives offered and subsequently chosen arms, the principal's estimation can be regarded as an analogy of online inverse optimization in MAB's. Next, we construct a policy that we prove achieves a low regret by deriving finite-sample concentration bounds for our estimator. We conclude with numerical simulations demonstrating the applicability of our \textcolor{black}{framework to personalized adherence-improving incentive design for long-term statin treatments}.

}%


 \HISTORY{}

\maketitle

%


\section{Introduction}
System designers frequently use the idea of providing incentives to stakeholders as a powerful means of steering the stakeholders for their own benefit. Operations management includes many such examples, such as offering performance-based bonuses to ride-hailing drivers, providing monetary incentives to patients for medical adherence, quality-contingent bonus payments for workers in crowdsourcing platforms, and vertical collaboration between shippers and carriers in transportation planning. In many real-world settings, the problem of designing efficient incentives can be posed as a repeated principal-agent problem where a principal (i.e., system designer) designs sequential incentive policies to motivate an agent (i.e., stakeholder) to convey certain behaviors that eventually serve the goal of maximizing the principal's cumulative net reward. Typically, there is an element of information asymmetry in these systems which arises between the principal and the agent in the form of either \emph{adverse selection} (i.e., hidden information) or \emph{moral hazard} (i.e., hidden actions) \citep{bolton2004contract}. For instance, in the context of employment incentives designed by an employer, the hidden information in an adverse selection setting could be the level of productivity of an employee whereas a hidden action in the moral hazard setting could be the total effort level of the employee. More generally, the hidden information in the adverse selection setting can be seen as an unknown ``type'' or ``preferences'' of the agent that directly affects the action chosen by the agent, which in turn determines both the agent's reward and the principal's reward. These situations require specification of the agent's private information and the distributional-knowledge that the principal has concerning that information. 

Existing literature on repeated principal-agent models mostly studies the moral hazard setting, with a more recent focus on the problem of estimating agent's unknown model parameters under hidden actions (e.g., \citealt{ho2016}, \citealt{kaynar2022estimating}). On the other hand, the adverse selection setting is mostly studied either for single-period static games \citep{navabi2018optimal, chade2019disentangling, gottlieb2022simple} or else for the repeated dynamic games where restrictive assumptions are made on, for example, dimension of the agent's action space, utility function of the agent, and relationship between principal's rewards and agent's unknown type (e.g., \citealt{halac2016optimal, esHo2017dynamic, maheshwari2022inducing}). Furthermore, the estimation and learning problem has not previously been explored under adverse selection. However, system designers in practice require more generic and richer dynamic approaches that leverage data on past incentives and observed actions without necessarily imposing a specific structure on the reward model or type distribution of the agent. 

Our main goal in this paper is to open a new window to repeated principal-agent models under adverse selection from the perspective of statistical learning theory. In particular, we consider an unexplored setting of adverse selection where the principal can only observe the history of the agent's actions but is uninformed about the associated rewards of the agent. To enhance the practical relevance of our approach, we design a generic and simple model. We assume that the agent has the perfect knowledge of their reward model and picks the reward-maximizing action based on the incentives provided by the principal at each period. Under this repeated \emph{unobserved rewards} setting, we are mainly interested in studying the following two research questions:   
\begin{enumerate}
    \item How to compute a statistically consistent estimator for a non-parametric reward model of the agent? 
    \item How to design data-driven and adaptive incentives that will attain low regret to the principal? 
\end{enumerate}

\subsection{Motivating Applications}
\subsubsection{Personalized Incentive Design for Medical Adherence}\label{subsubsec:medical}
The problem of patients not following medication dosing instructions is recognized as a major and widespread issue around the world. Such lack of adherence to a medication regime leads to not only poor health outcomes but also substantial financial costs \cite{osterberg2005adherence}. According to \cite{world2003adherence}, medication non-adherence is observed 50\% of the time, which may increase up to 80\% for relatively asymptomatic diseases such as hypertension \citep{brown2016medication}. Research reveals various reasons for this problem including individual-level factors (e.g., medication side effects), social factors (e.g., cultural beliefs), and economic factors (e.g., transportation costs to clinics) \citep{world2003adherence, bosworth2010medication, long2011patient}. To overcome some of these concerns, incentive programs that provide financial rewards to the patients are commonly employed and shown to effectively improve medical adherence. There is a related literature in medicine and economics on examining the effects of these monetary incentives using empirical analyses \citep{lagarde2007conditional, gneezy2011and} and in operations management on quantitatively designing the financial incentives for different market contexts \citep{aswani2018inverse, ghamat2018contracts, guo2019impact, suen2022design}. 

The design of financial incentives throughout a medication regime with finite length adequately features a repeated principal-agent problem under the hidden rewards setting introduced in this paper. Given their personal preferences and characteristics (i.e., type), the patient (i.e., agent) exhibits certain adherence behaviors in order to maximize their total utility which is comprised of benefits obtained through the improvements in their health conditions, costs incurred due adherence, and incentives offered by the healthcare provider. On the other hand, the goal of the healthcare provider (i.e., principal) is to maximize the clearance rate, that is the rate at which the infected patient is recovered, by designing motivating payments to improve their adherence actions. This payment design problem is nontrivial due to scarce clinical resources and the information asymmetry between the provider and the patient. Although the healthcare provider can often fully observe the patient's adherence actions, the realized utilities of the patient often stands as a private information to the provider. Because our data-driven incentive design framework presented is based on a generic model without restrictive technical assumptions, it could be easily leveraged to overcome the problem of medical non-adherence. 

\color{black}
\subsubsection{Deforestation Incentives for Payment for Ecosystem Services}\label{subsubsubsec:PES}
According to the UN Food and Agriculture Organization, deforestation since 1990 is estimated to have reached 420 million hectares worldwide, with an annual rate of around 10 million hectares every year since then, the majority of which belongs to tropical forests \citep{forest}. However, the rate of forest loss is reported to have declined substantially in the last five years due to various interventions, such as Payment for Ecosystem Services (PES) \citep{busch2017drives, science}. PES programs establish a positive economic incentive mechanism by which governments and non-governmental organizations contract owners of natural resources (i.e., local landowners or service providers, such as private forest owners and farmers) to incentivize them to prevent environmental degradation, such as deforestation \citep{warnes2023area}. 

From the perspective of payment providers, the goal of these incentive mechanisms is to maximize the environmental benefits obtained from the amount of conserved forests with the minimal amount of payments. However, a prevalent obstacle in effectiveness of these contracts involves the information disparities between program designers and the forest owners \citep{salzman2018global}. Typically, the forest owners hold private information about their opportunity costs and the amount of deforestation they would choose without any incentives \citep{engel2016devil}. Only a few papers in the literature have studied the contract design problem in the PES context by using a principal-agent framework in which the agent has a privately-known baseline conservation level for a known initial forest area \citep{mason2013additionality, li2023simple}. These studies characterize the optimal contracts for static one-shot games between providers and forest owners, while leaving the analysis open for dynamic performance-based incentive policies. To bridge this gap, we suggest that PES incentives should incorporate the effects of learning from forest owners' conservation actions over time. Our sequential \emph{hidden rewards} model can be effectively extended to formalize this information asymmetric setup, and the proposed data-driven incentive design framework can help reveal the true willingness of forest owners for their land use.
\color{black}

\subsubsection{Sustainable and Collaborative Transportation Planning with Backhauling} \label{subsubsec:backhaul}
Backhauling is a way of improving the efficiency of shipping vehicles by providing pickup loads for them on their way back to the origin depot. It has been widely applied in logistics operations to reduce both the transportation costs of companies and negative environmental impacts due to fuel consumption and pollutant emissions \citep{early_2011, juan2014, turkensteen2017combining}. In the context of collaborative transportation in a supply chain network, backhauling is a complex, yet powerful, tool for achieving green closed-loop logistics. Due to the hierarchical relationship between shippers (i.e., companies) and carriers in a transportation network, it is often studied as a form of vertical collaboration in which companies create integrated outbound-inbound routes -- instead of dedicated delivery and dedicated pickup routes -- and provide incentives (i.e., side payments) to carriers to induce these routes \citep{ergun2007reducing, audy2012framework, marques2020integrated, santos2021bilevel}. 

These existing approaches focus on solving the shipper's single-period static routing and pricing problems by using techniques mostly from optimization theory. However, in practice, shippers face these decisions and interact with carriers dynamically and repeatedly at every shipment request. Therefore, there is clearly a need for designing the vertical collaboration between a shipper and a carrier as a sequential learning and decision-making process. In that regard, this incentive design problem can be formulated as a repeated game between a principal (shipper) and an agent (carrier) under adverse selection. The goal of the shipper is to initiate the use of pre-planned integrated routes for their linehaul and backhaul customers to minimize their total transportation costs, whereas the carrier aims to maximize their total profits from the selected routes. At the end of each shipment request, the shipper observes the set of routes chosen by the carrier after the provided incentives while the total profit obtained by the carrier stands as invisible information to the shipper -- which makes it more challenging for the shipper to predict and orient the carrier's future selections. Taking into account all these features, the repeated \emph{hidden rewards} model and the adaptive incentive policy proposed here can explicitly account the goals and interactions of both parties and yield effective incentive plans by leveraging the available data over a given contract period.  

\subsection{Main Contributions and Outline}
We next present an outline of our paper and our main methodological contributions in more detail.
\begin{description}
\item[\textit{\textbf{Consistent estimator.}}] In Section \ref{subsec:model}, we provide the details of the principal-agent setting introduced above. Then, we introduce a novel estimator for a non-parametric reward model of a reward-maximizing agent with finite set of actions in Section \ref{subsec:estimator}. Our estimator is formulated exactly as a linear optimization model that estimates the expected rewards of all actions without assuming any \textcolor{black}{specific form of the distribution function}. In accordance with the \emph{hidden rewards} setting, the only input to our estimator is the data on past incentives and past actions chosen by the agent. In Section \ref{subsec:identifiabilityconsistency}, we give results proving identifiability and finite-sample statistical consistency of the proposed estimator. Essentially, we prove probability bounds on the diameter of the random polytope defined by the feasible space of our estimator in each time period.
\item[\textit{\textbf{Data-driven and low-regret incentives.}}] Section \ref{subsec:epsilongreedy} describes a practical and computationally efficient $\epsilon$-greedy policy for the design of the principal's adaptive incentives over a finite time horizon of length $T$. By utilizing the finite-sample concentration bounds derived for our estimator, we compute the regret of the proposed policy with respect to an oracle incentive policy that maximizes the principal's expected net reward at each time step under the perfect knowledge of all reward expectations. Section \ref{subsec:regret} presents a rigorous regret bound of order $O(\sqrt{T \log T})$ for the repeated principal-agent models with hidden agent rewards. 
\item[\textit{\textbf{Discussion and Numerical results.}}] Our approach assumes that the agent's decisions are consistent with a fixed vector of reward expectations. \textcolor{black}{We further explore scenarios where the agent is knowledgeable about the principal's model and can indeed extract higher information rents by basing their actions on a fixed reward vector that differs from the true one. In Section \ref{sec:agent}, we provide a discussion from the perspective of the reward-maximizing agent and argue that our incentive scheme is designed to maximize the principal's cumulative expected net reward subject to the inherent rent collected by the agent.} To support our theoretical results and demonstrate our data-driven incentive framework, we also conduct simulation experiments on an instance of the \textcolor{black}{personalized medical adherence incentives} problem outlined earlier. In Section \ref{sec:numerical}, we share the details of our experimental setting and numerical results. 
\end{description}
Lastly, we conclude in Section \ref{sec:conclusion} by discussing future work that might be steered by our analyses in this paper. We include the proofs for all theoretical results provided in the main text in Appendices.

\subsection{Related Literature}
\begin{description}
\item[\textit{\textbf{Repeated Principal-Agent Models}}.] 
There is a rich and extensive literature on principal-agent models in economics \citep{holmstrom1979moral, grossman1983, hart1987theory} and in operations management \citep{martimort2009theory}. For repeated models, most existing studies focuses on the moral hazard setting \citep{radner1981monitoring, rogerson1985repeated, spear1987repeated, abreu1990toward, plambeck2000performance, conitzer2006, sannikov2008, sannikov2013}. Several of them study the problem of estimating the agent's model when actions are hidden \citep{vera2003structural, misra2005salesforce, misra2011structural, ho2016, kaynar2022estimating}. On the other hand, related work on the design of incentives under the adverse selection setting is relatively scarce. In many of them, the agent's type (e.g., level of effort or probability of being successful) is considered as an additional, unknown information on top of a moral hazard setting \citep{dionne85, banks1993adverse, gayle2015identifying, williams2015solvable, halac2016optimal, esHo2017dynamic}. Only a few of these works study the estimation problem for the hidden type setting, and they use statistical estimation methods such as least squares approximation \citep{leezenios2012}, minimization of a sum of squared criterion function \citep{gayle2015identifying}, and simulation-based maximum likelihood estimation \citep{aswani2019data, mintz2023behavioral}. However, the adverse selection setting studied in these papers comes with limiting assumptions such as the assumption that the agent's type parameter belongs to a discrete set. 

Our work differs from these studies in several ways. Although the \textit{unobserved rewards} setting has various application areas, we are not aware of any other paper studying this novel and non-trivial dynamic principal-agent model. The estimation problem we consider in this setting involves estimating the reward expectation values which belong to a bounded continuous space. Differently from the existing work summarized above, we solve a practical linear program and follow a set-based estimation approach to estimate these continuous mean rewards. Furthermore, regarding the incentive design problem, these past papers do not consider the exploration-exploitation trade-off faced by the principal, and hence, they are not able to provide guarantees on how close to optimal their solutions are. In this paper, we take a sequential learning approach to compute adaptive and efficient incentives for the principal and perform a regret analysis for the considered repeated adverse selection models.  

\item[\textit{\textbf{Multi-Armed Bandits for Incentive Design}}.] 
A related line of research from sequential decision-making includes the use of a multi-armed bandit (MAB) framework for mechanism design. MAB's are widely applied to dynamic auction design problems which are closely related with the incentive design in dynamic principal-agent problems \citep{Nazerzadeh2008DynamicCM, Devanur2009ThePO, jain2014multiarmed,  NIPS2014, biswas2015truthful, ho2016, braverman2019multi, bhat2019optimal, abhishek2020designing, shweta2020multiarmed, han2020learning, simchowitz2021, wang2022, gao2022combination}. 

The principal's problem in our repeated game between the principal and agent under unobserved rewards is directly applicable to the MAB framework. At each iteration of the game, the principal offers a set of incentives corresponding to the set of arms (i.e., actions) in the agent's model and generates a random reward through the arm selected by the agent. As the interaction between these two parties proceeds, the principal faces a trade-off between consistently learning the unknown reward expectations of each agent arm (i.e., \textit{exploring} the agent's arm space through adverse incentives to encourage selection of different arms) and maximizing their own cumulative net reward (i.e., \textit{exploiting} arms estimated to yield the highest expected rewards by motivating the agent to select them with the minimal incentives). For this reason, the MAB framework is useful in effectively navigating the principal's exploration-exploitation trade-off while designing data-driven incentives.

\item[\textit{\textbf{Inverse Optimization}}.] Inverse optimization is a framework for inferring parameters of an optimization model from the observed solution data that are typically corrupted by noise \citep{ahuja2001inverse, heuberger2004inverse}. More recent work in this area probes into estimating the model of a decision-making agent by formulating the agent's model as a linear or a convex optimization problem in offline settings (where data are available a priori) \citep{keshavarz2011imputing, bertsimas2015data, esfahani2018data, aswani2018inverse, chan2019inverse, chan2022inverse} or in online settings (where data arrive sequentially) \citep{barmann2018online, dong2018generalized, dong2020inverse, maheshwari2023convergent}. Different from these studies, we do not assume any specific structure of the agent's decision-making problem, but instead we consider a reward-maximizing agent with finite action space. This case of estimating the non-parametric model of a reward-maximizing agent is also addressed by \cite{kaynar2022estimating}, who study the offline static setting of the principal-agent problem under moral hazard. A key distinction is that we study the online dynamic setting of the repeated principal-agent problem under adverse selection. In accordance with the unobserved rewards setting that we examine, we design an estimator for the expected rewards of the agent's arms, whose only input is the data of reward-maximizing arms in response to the provided incentives in the past. In that respect, the principal's estimation problem under the sequential unobserved rewards setting can be regarded as an analogy of online inverse optimization in MAB's. Moreover, to prove consistency of the principal's estimator in this setting, we build upon initial ideas of statistics with set-valued functions \citep{aswani2019statistics}.
\end{description}

\subsection{Mathematical Notation}
We first specify our notational conventions throughout the paper. All vectors are denoted by boldfaced lowercase letters. A vector $\mathbf{x}$ whose entries are indexed by a set $\mathcal{M} = [1, \ldots, M]$ is defined as $\mathbf{x} = (x_m)_{m \in \mathcal{M}}$. If each entry $x_m$ belongs to a set $\mathcal{X}$, then we have $\mathbf{x} \in \mathcal{X}^M$. The $\ell_\infty$-norm of the vector $\mathbf{x}$ is defined by $\|\mathbf{x} \|_\infty = \max (|x_1|, \ldots, |x_M|)$. Further, the cardinality of a set $\mathcal{X}$ is denoted by $|\mathcal{X}|$, and $\mathbbm{1}(\cdot)$ denotes the indicator function that takes value $1$ when its argument $(\cdot)$ is true, and $0$ otherwise. Lastly, the notations  $\mathbf{0}_n$ and $\mathbf{1}_n$ are used for the all-zeros and all-ones vectors of size $n$, respectively, and $\mathbb{P}(\cdot)$ is used for probabilities. 
\section{The Repeated Game and Principal's Estimation}\label{sec:estimator}
We start this section by introducing our repeated principal-agent model under adverse selection and continue by presenting our novel estimator along with the associated statistical results. 
\subsection{The Repeated Game with Hidden Rewards}\label{subsec:model}
We consider a repeated play between a principal and an agent over a finite time horizon $\mathcal{T} = [1, \ldots, T]$. At each time step $t \in \mathcal{T}$, the principal offers a vector of incentives $\boldsymbol{\pi}_t = (\pi_{t,a})_{a \in \mathcal{A}}$ corresponding to the set of all possible actions of the agent $\mathcal{A} = \{1, \ldots, n\}$. Then, the agent takes the action $i_t(\boldsymbol{\pi}_t)$ with the maximum expected total reward given the incentives $\boldsymbol{\pi}_t$, i.e.,
\begin{equation}
    i_t(\boldsymbol{\pi}_t) := \argmax \ (\mathbf{r}^0 + \boldsymbol{\pi}_t) = \argmax_{a \in \mathcal{A}} \left(r^0_a  + \pi_{t,a} \right)
\end{equation}
where $\mathbf{r}^0 = (r^0_a)_{a \in \mathcal{A}}$ is the true vector of expected rewards of the agent and is only known by the agent. We assume that $r^0_a, \forall a \in \mathcal{A}$ belongs to a compact set $\mathcal{R} = [R_{\min}, R_{\max}] \subset \mathbb{R}$ where $R_{\max} - R_{\min} \geq 1$. Based on the action chosen by the agent, the principal collects a stochastic reward outcome denoted by $\mu_{t, i_t(\boldsymbol{\pi}_t)} \sim \mathbb{F}_{\theta^0_{i_t(\boldsymbol{\pi}_t)}, i_t(\boldsymbol{\pi}_t)}$ with expectation $\theta^0_{i_t(\boldsymbol{\pi}_t)} \in \Theta$ where $\Theta$ is a known compact set. The true mean reward vectors $\mathbf{r}^0$ and $\boldsymbol{\theta}^0 = (\theta^0_a)_{a \in \mathcal{A}}$ are unknown by the principal. The principal can only observe the selected action $i_t(\boldsymbol{\pi}_t)$ and their own net reward realization $\mu_{t, i_t(\boldsymbol{\pi}_t)} - \color{black} \pi_{t, i_t(\boldsymbol{\pi}_t)}$\color{black}.

\textcolor{black}{We emphasize that the agent’s utility model is deliberately kept generic to ensure the wide relevance of the estimation problem under study. This formulation can accommodate a broad class of models, including those that incorporate the agent’s risk preferences over random rewards. Its practical adaptability is illustrated in our numerical experiments in Section \ref{sec:numerical}.}

In this setting, to ensure that our research problems are well-posed, it suffices to assume that the range of the incentives that the principal is able to provide to the agent covers the range of the agent's reward expectations. 
\begin{assumption}\label{assm1}
The incentives $\pi_{t,a}$, $\forall a \in \mathcal{A}$ belongs to a compact set $\mathcal{C} = [\underline{C}, \overline{C}]$ where $\underline{C} = R_{\min}$  and $\overline{C} = R_{\max} + \gamma$ for some constant $0 < \gamma \leq R_{\max} - R_{\min} - 1$. 
\end{assumption}
Because the principal's goal is to provide incentives that will drive the agent's decisions, this assumption ensures that the magnitudes of the incentives are large enough to have an effect on the relative order of the actions with respect to their total rewards after adding the incentives.  

\subsection{The Estimator}\label{subsec:estimator}
Due to the information asymmetry in our repeated adverse selection model, the learning process of the principal comprises estimating the agent's expected reward vector $\mathbf{r}^0$ by solely watching the actions maximizing the total reward vector $\mathbf{r}^0 + \boldsymbol{\pi}_\tau$ in the past time periods $\tau \leq t$. Our fundamental observation of this estimation problem is that the differences of pairs of entries of $\mathbf{r}^0$ is crucial for the statistical analysis, not the individual values of the entries. With this observation on hand, we must first discuss an ambiguity in this problem before formulating our estimator. 

Consider two different estimates of the mean reward vector, $\mathbf{r}' \in \mathcal{R}^n$ and $\mathbf{r}'' = \mathbf{r}' + k \mathbf{1}_n$, where $k$ is any constant scalar such that $\mathbf{r}'' \in \mathcal{R}^n$. For a given incentive vector $\boldsymbol{\pi}_\tau$, the principal will not be able to distinguish between $\mathbf{r}'$ and $\mathbf{r}''$ in the considered affine space since both estimates will yield the same maximizer action, that is $\argmax \ (\mathbf{r}' + \boldsymbol{\pi}_\tau) = \argmax \ (\mathbf{r}'' + \boldsymbol{\pi}_\tau)$. To overcome this issue of identifiability, we can remove one redundant dimension from the considered estimation problem by setting all the differences of pairs of $\mathbf{r}$'s entries with respect to a reference point 0. 
\begin{definition} For a mean reward vector $\mathbf{r} = (r_1, r_2, \ldots, r_n) \in \mathcal{R}^n$, we define $\mathbf{s}$ as the normalized mean reward vector that is without loss of generality defined by $\mathbf{s}:= \mathbf{r} - r_1 \mathbf{1}_n = (0, r_2 - r_1, \ldots, r_n -r_1)$ and belongs to the compact set $\mathcal{S}^n = [R_{\min} - R_{\max}, R_{\max}- R_{\min}]^n$. 
\end{definition}
This dimensionality reduction allows us to decrease our degrees of freedom and derive the identifiability result for our estimator. We further note that the maximizer action $i_\tau(\boldsymbol{\pi}_\tau)$ for the total expected reward vector $\mathbf{r}^0 + \boldsymbol{\pi}_\tau$ is also the maximizer for $\mathbf{s}^0 + \boldsymbol{\pi}_\tau$. Therefore, we will define our estimator and conduct our theoretical analyses with respect to the normalized reward vector $\mathbf{s}$. 

Next, we formalize our estimator for $\mathbf{s}^0$. Let $\boldsymbol{\Pi}_t = \{\boldsymbol{\pi}_1, \ldots, \boldsymbol{\pi}_{t-1}\}$ be the sequence of incentives offered by the principal and $I_t(\boldsymbol{\Pi}_t) = \{i_1(\boldsymbol{\pi}_1), \ldots, i_{t-1}(\boldsymbol{\pi}_{t-1})\}$ be the sequence of actions chosen by the agent up to time $t$. Then, the principal's estimate $\widehat{\mathbf{s}}_t\left(I_t(\boldsymbol{\Pi}_t), \boldsymbol{\Pi}_t\right)$ at time $t$ for the agent's normalized mean reward vector $\mathbf{s}^0$ is formulated as
    \begin{align} 
        \widehat{\mathbf{s}}_t\left(I_t(\boldsymbol{\Pi}_t), \boldsymbol{\Pi}_t\right) \in &\argmin \hspace{0.5em}  0 \allowdisplaybreaks  \\
            &\hspace{0.2em} \mathrm{s.t.} \hspace{0.5em} s_{i_\tau(\boldsymbol{\pi}_\tau)}  + \pi_{\tau,i_\tau(\boldsymbol{\pi}_\tau)} \geq s_a  + \pi_{\tau,a} && \quad \forall a \in \mathcal{A}, \ \tau = 1, \ldots, t-1  \allowdisplaybreaks  \\
            &\hspace{1.9em} s_1 = 0, \ s_a \in \mathcal{S} && \quad \forall a \in \mathcal{A} \allowdisplaybreaks  
    \end{align} 
This optimization problem can be regarded as the feasibility version of the set-membership estimation problem \citep{schweppe1967, hespanhol2020statistical}. {\color{black} It is important to clarify the distinction between our estimator and the classical models studied in two related literatures: the ranking and selection (R\&S) and MABs. In R\&S, the goal is to use sampling to identify the alternative with the largest mean \citep{hong2021review}, while in MABs the focus is on sequential learning to maximize cumulative reward. In contrast, our setting requires learning the relative ordering of means across the entire set of arms. This poses a significant theoretical challenge, as the difficulty of the principal’s learning problem grows with the size of the arm space. Consequently, every alternative action matters equally in our framework, unlike in the classical R\&S or MAB formulations.} 

We can reformulate it by defining the loss function 
\begin{align}
    L\left(\mathbf{s}, I_t(\boldsymbol{\Pi}_t), \boldsymbol{\Pi}_t\right) = \sum_{\tau = 1}^{t-1} \ell\left(\mathbf{s}, i_\tau(\boldsymbol{\pi}_\tau), \boldsymbol{\pi}_\tau\right) \label{eq:lossfunc}
\end{align}
which is the sum of $t-1$ extended real-valued functions given by
\begin{align}
    \ell\left(\mathbf{s}, i_\tau(\boldsymbol{\pi}_\tau), \boldsymbol{\pi}_\tau\right) = \left\{\begin{array}{ll}
    0, & \mathrm{if} \hspace{0.5em} s_{i_\tau(\boldsymbol{\pi}_\tau)}  + \pi_{\tau,i_\tau(\boldsymbol{\pi}_\tau)} \geq s_a  + \pi_{\tau,a}, \ \forall a \in \mathcal{A} \allowdisplaybreaks  \\
    + \infty, & \mathrm{otherwise } 
    \end{array}\right\}.
\end{align}
Now, we reformulate our feasibility estimator as 
\begin{equation} 
\label{prblm:estimator}
  \begin{alignedat}{4}
           \widehat{\mathbf{s}}_t\left(I_t(\boldsymbol{\Pi}_t), \boldsymbol{\Pi}_t\right) \in & \argmin_{s_1 = 0, \ s_a \in \mathcal{S}, \forall a \in \mathcal{A}} L\left(\mathbf{s}, I_t(\boldsymbol{\Pi}_t), \boldsymbol{\Pi}_t\right)
  \end{alignedat}
\end{equation}
Note that we may use the simplified notation $\widehat{\mathbf{s}}_t$ throughout the paper for conciseness. We next present the results of our statistical analysis for the estimator (\ref{prblm:estimator}). 

\subsection{Identifiability and Consistency}\label{subsec:identifiabilityconsistency}
The convergence behavior of the sequence of estimates $\widehat{\mathbf{s}}_t$ depends on a characterization of the loss function known as an \emph{identifiability condition} \citep{van2000asymptotic} that ensures the loss function is minimized uniquely by the true vector $\mathbf{s}^0$. We start our consistency analysis by proving the identifiability of our estimator (\ref{prblm:estimator}). The identifiability of our estimation problem requires characterizing the set of incentive vectors that distinguishes between $\mathbf{s}^0$ and an incorrect estimate $\mathbf{\widehat{s}}_{t}$. We first provide some intermediate results in Propositions \ref{prop:iden1} -- \ref{prop:iden3} and then formalize the final identifiability result for our estimator in Proposition \ref{prop:iden4}.

Let $\mathcal{N}(\mathbf{s}^0, \beta) \subset \mathcal{S}^n$ be an open neighborhood centered around $\mathbf{s}^0$ with diameter $\beta > 0$ such that $\mathcal{N}(\mathbf{s}^0, \beta) := \{\mathbf{s} : \| \mathbf{s} - \mathbf{s}^0 \|_\infty \leq \beta\}$, and consider the compact set $\mathcal{F} := \mathcal{S}^n \setminus \mathcal{N}(\mathbf{s}^0, \beta)$. We define an open ball $\mathcal{B}(\mathbf{s}^j, d) := \{\mathbf{s} : \| \mathbf{s} - \mathbf{s}^j \|_\infty < d\}$ centered around a vector $\mathbf{s}^j$ with diameter $d > 0$.  Since $\mathcal{F}$ is compact, for some finite $q > 0$ and $d < \beta$, there is a finite subcover $\{\mathcal{B}(\mathbf{s}^j, d) : \mathbf{s}^j \in \mathcal{F}\}_{j = 1}^q$ of a collection of open balls covering $\mathcal{F}$. Given a normalized reward vector $ \mathbf{s} \in \mathcal{B}(\mathbf{s}^j, d), j \in \{1, \ldots, q\}$, our arguments in the following propositions will be based on the following indices:
\begin{itemize}
    \item $K := \argmax_{a \in \mathcal{A}} s_{a}$ (the set of indices corresponding to the highest value entries in $\mathbf{s}$)
    \item $K^0 := \argmax_{a \in \mathcal{A}} s^0_a$ (the set of indices corresponding to the highest value entries in $\mathbf{s}^0$)
    \item $b \in \argmax_{a \in \mathcal{A}} |s^0_a - s_{a}|$ (the index of an entry with the highest absolute value in $\mathbf{s}^0 - \mathbf{s}$)
\end{itemize}
\begin{proposition} \label{prop:iden1} Suppose that $K^0 \cap K = \emptyset$ for a given vector $\mathbf{s} \in \mathcal{B}(\mathbf{s}^j, d), j \in \{1, \ldots, q\}$, and that the principal chooses each incentive $\pi_{t,a}$ uniformly randomly from the compact set $\mathcal{C}$, that is $\pi_{t,a} \sim U(\underline{C}, \overline{C}), \forall a \in \mathcal{A}$, at time $t \in \mathcal{T}$. Then,
 \begin{align}
        &\mathbb{P}\left(\ell\left(\mathbf{s}, i_t(\boldsymbol{\pi}_t), \boldsymbol{\pi}_t\right) = +\infty \right) \geq 
        \left(\frac{1}{2} -\frac{\left(\overline{C} - \underline{C} - s^0_{\kappa^0} +  s^0_{\kappa}\right)^2 }{2(\overline{C}-\underline{C})^2} \right)\left(1 - \frac{\gamma + \beta - d}{\overline{C} - \underline{C}} \right)^2  \left( \frac{\gamma}{\overline{C} - \underline{C}} \right)^{n-2}  \label{eq:identf-1}
\end{align}
for any $\kappa \in K, \ \kappa^0 \in K^0$, and $\gamma$ as introduced in Assumption \ref{assm1}. 
\end{proposition}
\begin{proposition} \label{prop:iden2} Suppose that $K^0 \cap K \neq \emptyset$, $b \notin K^0 \cap K$ for a given vector $\mathbf{s} \in \mathcal{B}(\mathbf{s}^j, d), j \in \{1, \ldots, q\}$, and that $\pi_{t,a} \sim U(\underline{C}, \overline{C}), \forall a \in \mathcal{A}$ at time $t \in \mathcal{T}$. Let $\omega = \sup_{\mathbf{s} \in \mathcal{B}(\mathbf{s}^j, d)} \max_{a \in \mathcal{A}} \{|s^0_a|, |s_a| \}$ be the largest absolute value among the entries of $\mathbf{s}^0$ and of all the vectors in $\mathcal{B}(\mathbf{s}^j, d)$. Then, 
 \begin{align}
        &\mathbb{P}\left(\ell\left(\mathbf{s}, i_t(\boldsymbol{\pi}_t), \boldsymbol{\pi}_t\right) = +\infty \right) \geq 
        \frac{\beta^2}{(\overline{C}-\underline{C})^2} \left(1 - \frac{\gamma + \omega}{\overline{C} - \underline{C}} \right)^2 \left( \frac{\gamma}{\overline{C} - \underline{C}}\right)^{n-2}. \label{eq:identf-2} \allowdisplaybreaks 
\end{align} 
\end{proposition}
\begin{proposition} \label{prop:iden3}
Suppose that $K^0 \cap K \neq \emptyset$, $b \in K^0 \cap K$ for a given vector $\mathbf{s} \in \mathcal{B}(\mathbf{s}^j, d), j \in \{1, \ldots, q\}$, and that $\pi_{t,a} \sim U(\underline{C}, \overline{C}), \forall a \in \mathcal{A}$ at time $t \in \mathcal{T}$. Then, 
 \begin{align}
        \mathbb{P}\left(\ell\left(\mathbf{s}, i_t(\boldsymbol{\pi}_t), \boldsymbol{\pi}_t\right) = +\infty \right) \geq 
        \frac{\beta^2}{(\overline{C}-\underline{C})^2} \left(1 - \frac{\gamma + \beta - d}{\overline{C} - \underline{C}} \right) \left(1 - \frac{\gamma + \omega}{\overline{C} - \underline{C}} \right) \left( \frac{\gamma}{\overline{C} - \underline{C}}  \right)^{n-2} \label{eq:identf-3} \allowdisplaybreaks 
\end{align}
for the constant $\omega$ defined in Proposition \ref{prop:iden2}. 
\end{proposition}
Propositions \ref{prop:iden1} -- \ref{prop:iden3} analyze three mutually exclusive cases for a given reward vector $\mathbf{s}$ and the true reward vector $\mathbf{s}^0$. In all cases, these results show that as the distance $\beta$ between the considered vector $\mathbf{s}$ and the true vector $\mathbf{s}^0$ increases, the probability that the estimator (\ref{prblm:estimator}) will be able to differentiate these two vectors is also increasing proportional to $\beta^2$, and that this probability of invalidating an incorrect estimate is always strictly positive. In other words, they state that the unknown mean reward vector $\mathbf{s}^0$ can be learned from the input data collected by offering randomly chosen incentives that explore the agent's action space. Proposition \ref{prop:iden4} combines these results to show that our adverse selection model satisfies an identifiability property required for a precise inference on the agent's hidden rewards. 

\begin{proposition} \label{prop:iden4} \textbf{\textit{(Identifiability)}}
At time $t \in \mathcal{T}$, suppose that $\pi_{t,a} \sim U(\underline{C}, \overline{C}), \forall a \in \mathcal{A}$. Then, for any normalized reward vector $ \mathbf{s} \in \mathcal{F}$, we have 
 \begin{align}
        \mathbb{P}\left(\ell\left(\mathbf{s}, i_t(\boldsymbol{\pi}_t), \boldsymbol{\pi}_t\right) = +\infty \right) \geq 
        \alpha \beta^2 \label{eq:identf-4} \allowdisplaybreaks 
\end{align}
for some constant $\alpha > 0$.
\end{proposition}
Theorem \ref{thm:concen} presents the finite-sample concentration behavior for our estimator with respect to the loss function (\ref{eq:lossfunc}). The main sketch of the proof of Theorem \ref{thm:concen} follows by the existence of the finite subcover $\{\mathcal{B}(\mathbf{s}^j, d) : \mathbf{s}^j \in \mathcal{F}\}_{j = 1}^q$ of an open covering of $\mathcal{F}$ and by using the result of Proposition \ref{prop:iden4} for each of the open balls in this subcover. Then, the final inequality is obtained by using volume ratios to bound the covering number $q$. The complete proof is given in Appendix \ref{appendix1}. The intuition behind the upper bound given in (\ref{prop:deter-finitesample}) is that the learning rate of the principal's estimator depends on the number of time periods at which the principal is exploring the action space of the agent.
\begin{theorem} \label{thm:concen}
Let $\eta(1, t)$ be the number of time steps that the principal chooses each incentive $\pi_{t,a}$ uniformly randomly from the compact set $\mathcal{C}$ up to time $t$, that is 
$\eta(1, t) = \left|\Lambda(1, t)\right|$ where $\Lambda(1, t)= \{\tau: 1 \leq \tau \leq t-1, \ \pi_{\tau,a} \sim U(\underline{C}, \overline{C}), a \in \mathcal{A} \}$. Then, we have
\begin{align}
    \mathbb{P}\left(\inf_{\mathbf{s} \in \mathcal{F}}  L\left(\mathbf{s}, I_t(\boldsymbol{\Pi}_t), \boldsymbol{\Pi}_t\right) < +\infty\right) \leq 
    \color{black} \exp \left(- \alpha (\eta(1, t)-1) \beta^2 - n \log \beta + n \log (2R_{\max} - 2R_{\min}) \right) \label{prop:deter-finitesample} \allowdisplaybreaks 
\end{align}
where $\mathcal{F} = \{\mathbf{s} \in \mathcal{S}^n: \| \mathbf{s} - \mathbf{s}^0 \|_\infty > \beta\}$ as before. 
\end{theorem}
This theorem is useful because it allows us to derive our finite-sample concentration inequality with respect to the distance between our estimates $\widehat{\mathbf{s}}_t$ and the true reward vector $\mathbf{s}^0$. We conclude this section with an alternative statement of Theorem \ref{thm:concen}. 
\begin{corollary}  \textbf{\textit{(Finite-Sample Concentration Bound)}} \label{cor:concen}
The principal's estimator (\ref{prblm:estimator}) satisfies 
\begin{align}
    \mathbb{P} \left(\|\mathbf{s}^0 - \widehat{\mathbf{s}}_t \|_\infty > \beta \right) \leq 
    \color{black} \exp \left(- \alpha (\eta(1, t)-1) \beta^2 - n \log \beta + n \log (2R_{\max} - 2R_{\min}) \right) 
\end{align} 
for any $\beta > 0$. 
\end{corollary}
Recall that the radius of a polytope is the maximum distance between any two points in it. Then, because both $\widehat{\mathbf{s}}_t$ and $\mathbf{s}^0$ are feasible solutions to (\ref{prblm:estimator}), this corollary can be also interpreted as a probability bound on the radius of the random polytope defined by the constraints of our estimation problem. 
\section{Principal's Adaptive Incentive Framework}\label{sec:learning}
In this section, we develop an adaptive incentive policy that yields an efficient regret bound for the principal in the repeated adverse selection game described in Section \ref{subsec:model}. As per the considered model setting, the principal needs to learn their own expected rewards $\boldsymbol{\theta}^0$ in addition to the agent's rewards. Because the principal can fully observe the reward outcomes $\mu_{t, i_t(\boldsymbol{\pi}_t)}$ that they get through the agent's decisions, we consider an unbiased estimator under the following assumption about the principal's reward distribution family.  
\begin{assumption}\label{assm2}
The principal's rewards $\mu_{t,a}$'s for an arm $a \in \mathcal{A}$ are independent and follow a sub-Gaussian distribution $\mathbb{F}_{\theta_a, a}$ for all $\theta_a \in \Theta$.  
\end{assumption}
This assumption states that the rewards $\mu_{t,a}$ and $\mu_{t',a}$ collected by the principal at any two time points $t, t'$ where the agent chooses arm $a$ are independent from each other. Assumption \ref{assm2} is a mild assumption that is commonly encountered in many MAB models. 

Let $T(a,t) = \left|\{\tau \in \mathcal{T} : \tau \leq t-1, i_\tau(\boldsymbol{\pi}_\tau) = a   \} \right|$ be the number of time points that the agent selects arm $a$ up to time $t$. Then, the principal's estimator for $\theta^0_a, \forall a \in \mathcal{A}$ is given by
\begin{align}
    \widehat{\theta}_{t,a} = \frac{1}{T(a,t)}\sum_{\tau = 1}^{t-1} \mu_{\tau,a} \mathbbm{1} \left(i_\tau(\boldsymbol{\pi}_\tau) = a\right) \label{eq:theta-hat}
\end{align}
which is the sample mean of the principal's reward outcomes for agent's arm $a$ up to time $t$. If the principal's reward distribution $\mathbb{F}_{\theta_a^0, a}$ for any $a \in \mathcal{A}$ is an exponential family distribution where the sufficient statistic is equal to the random variable itself, such as Bernoulli, Poisson, and the multinomial distributions, then $\widehat{\theta}_{t,a}$ corresponds to the maximum likelihood estimator for $\theta^0_a$.

\subsection{Principal's $\epsilon$-Greedy Algorithm}\label{subsec:epsilongreedy}
We develop an $\epsilon$-greedy algorithm that integrates the principal's estimation problem and the incentive design problem in a practical learning framework. The pseudocode of the principal's $\epsilon$-greedy algorithm is given in Algorithm \ref{alg:epsilon}. 

\begin{algorithm*}
\caption{Principal's $\epsilon$-Greedy Algorithm}
\label{alg:epsilon} 
\begin{algorithmic}[1]
\State Set: $m \geq 4$, $\alpha > 0$ 
\For{$t \in [1, \ldots, n]$} \label{alg-initial1}
\State Set: $\boldsymbol{\pi}_t = (\pi_{t,a})_{a \in \mathcal{A}}$ where $\pi_{t, a} = \overline{C}$ for $a = t$ and $\pi_{t, a} = 0$ for all $a \neq t$
\If {$t \geq 2$} $\widehat{\theta}_{t,  i_{t-1}(\boldsymbol{\pi}_{t-1})} = \mu_{t-1, i_{t-1}(\boldsymbol{\pi}_{t-1})}$ \label{alg-initiallast}
\EndIf
\State Observe: $i_t(\boldsymbol{\pi}_t) = \argmax\limits_{a \in \mathcal{A}} \left(s^0_a  + \pi_{t,a}\right)$ and $\mu_{t, i_t(\boldsymbol{\pi}_t)}$
\EndFor
\For{$t \in [n+1, \ldots, T]$} 
\State Compute: $\widehat{\theta}_{t, i_{t-1}(\boldsymbol{\pi}_{t-1})} \in \frac{1}{T(i_{t-1}(\boldsymbol{\pi}_{t-1}), t)} \sum\limits_{\tau = 1}^{t-1} \mu_{\tau,i_{t-1}(\boldsymbol{\pi}_{t-1})} \mathbbm{1} (i_\tau(\boldsymbol{\pi}_\tau) = i_{t-1}(\boldsymbol{\pi}_{t-1})) $ 
\State Set: $\epsilon_t = \min \big\{1, \nicefrac{m}{t}\big\}$   \label{alg-m}
\State Sample: $x_t \sim \mathrm{Bernoulli} (\epsilon_t)$
\If {$x_t = 1$}
\State Sample: $\pi_{t,a} \sim \mathcal{U}\left(\underline{C}, \overline{C}\right)$ for all $a \in \mathcal{A}$ \label{alg-explore1}
\State Set: $\boldsymbol{\pi}_t = (\pi_{t,a})_{a \in \mathcal{A}}$ \label{alg-explorelast}
\Else {} 
\State Compute: $\beta_t = \sqrt{\frac{\log (\eta(1, t)-1)}{\alpha (\eta(1, t)-1)}}$ where $\eta(1, t) = \big| \{\tau: x_\tau = 1, n + 1 \leq \tau \leq t-1\} \big|$ \label{alg:beta}
\State Compute: $\widehat{\mathbf{s}}_t \in \argmin \left\{L\left(\mathbf{s}, I_t(\boldsymbol{\Pi}_t), \boldsymbol{\Pi}_t\right) \big | s_1 = 0, s_a \in \mathcal{S}, \forall a \in \mathcal{A}\right\}$ \label{alg-exploit1}
\For{$j \in \mathcal{A}$}  \label{alg-steps1}
\State Compute: $\widetilde{V}(j, \widehat{\mathbf{s}}_t; \widehat{\boldsymbol{\theta}}_t) = \widehat{\theta}_{t,j} - \left(\max\limits_{a \in \mathcal{A}} \widehat{s}_{t,a}\right) + \widehat{s}_{t,j} - 2\beta_t$
\EndFor
\State Compute: $j^*_t = \argmax\limits_{j \in \mathcal{A}} \widetilde{V}(j, \widehat{\mathbf{s}}_t; \widehat{\boldsymbol{\theta}}_t)$
\State Set: $c_{j^*_t}(\widehat{\boldsymbol{\theta}}_t, \widehat{\mathbf{s}}_t) = \left(\max\limits_{a \in \mathcal{A}} \widehat{s}_{t,a}\right) - \widehat{s}_{t,j^*_t} + 2\beta_t$ and $c_{a}(\widehat{\boldsymbol{\theta}}_t, \widehat{\mathbf{s}}_t) = 0 $ for all $a \neq j^*_t$ \label{alg-stepslast}
\State Set: $\boldsymbol{\pi}_t = (c_{a}(\widehat{\boldsymbol{\theta}}_t, \widehat{\mathbf{s}}_t))_{a \in \mathcal{A}}$ \label{alg-exploitlast}
\EndIf
\State Observe: $i_t(\boldsymbol{\pi}_t) = \argmax\limits_{a \in \mathcal{A}} \left(s^0_a  + \pi_{t,a}\right)$ and $\mu_{t, i_t(\boldsymbol{\pi}_t)}$ \label{alg-action}
\EndFor
\end{algorithmic}
\end{algorithm*}

During the first $n = |\mathcal{A}|$ time periods, the principal makes the agent select each of the $n$ actions once so that the principal will be able to record a reward observation and have an initial estimate of $\theta^0_a$ for all $a \in \mathcal{A}$. To achieve this, the principal offers the maximum possible incentive ($\overline{C}$) for the desired action which is sufficient to make it the agent's reward-maximizer action by Assumption \ref{assm1}. After this initialization period, at each time point $t \in [n+1, \ldots, T]$, the algorithm first updates the estimate of $\theta^0$ for the most recently played action $i_{t-1}(\boldsymbol{\pi}_{t-1})$, and then samples a Bernoulli random variable $x_t$ based on the exploration probability $\epsilon_t$. If $x_t = 1$, then the algorithm performs a pure exploration step by simply choosing an incentive vector $\boldsymbol{\pi}_t = (\pi_{t,a})_{a \in \mathcal{A}}$ where each component $\pi_{t,a}$ is selected uniformly randomly from the compact set $\mathcal{C}$. On the other hand, if $x_t = 0$, then the principal performs a greedy exploitation by first updating their estimate $\widehat{\mathbf{s}}_t$ for the unknown mean rewards of the agent by solving the estimation problem (\ref{prblm:estimator}). Next, the principal computes the vector of incentives $\mathbf{c}(\widehat{\boldsymbol{\theta}}_t, \widehat{\mathbf{s}}_t)$ that maximizes their estimated expected net reward at time $t$. The expected net reward of the principal is computed by subtracting the provided incentive at that time step from the expected reward that the principal will collect through the action which will be chosen by the agent. However, since the agent's true rewards are unknown, the principal cannot exactly know in advance the action that will be chosen by the agent after the provided incentives. Therefore, the principal tries to incentivize the agent to select the action that is estimated to maximize the principal's expected net reward at that time step. This is ensured by incrementing the calculated incentive by an additional amount {\color{black} ($2\beta_t$)} to account for the uncertainty in the estimates of the agent's expected rewards. 

For that purpose, using $\widehat{\boldsymbol{\theta}}_t$ and $\widehat{\mathbf{s}}_t$, the principal first estimates the minimum incentives (denoted by $(\widetilde{c}^j_a)_{a \in \mathcal{A}}$) required to make the agent pick an action $j \in \mathcal{A}$ and the corresponding expected net reward value (denoted by $\widetilde{V}(j, \widehat{\mathbf{s}}_t; \widehat{\boldsymbol{\theta}}_t)$) that will be observed after action $j$ is taken by the agent:
\begin{align}
    &\widetilde{c}^j_j = \left(\textstyle \max\limits_{a \in \mathcal{A}} \widehat{s}_{t,a}\right) - \widehat{s}_{t,j} + 2\beta_t \label{eq:incentive1}  \allowdisplaybreaks  \\
    &\widetilde{c}^j_a = 0, \quad \forall a \in \mathcal{A}, \ a \neq j  \label{eq:incentive2}  \allowdisplaybreaks  \\
    &\widetilde{V}(j, \widehat{\mathbf{s}}_t; \widehat{\boldsymbol{\theta}}_t) = \widehat{\theta}_{t, j} - {\color{black} \widetilde{c}^j_j} =  \widehat{\theta}_{t, j} - \left(\max\limits_{a \in \mathcal{A}} \widehat{s}_{t,a}\right) + \widehat{s}_{t,j} - 2\beta_t \allowdisplaybreaks \label{eq:incentive3}
\end{align}  
where $\beta_t >0, \forall t$. After computing these values for every action $j \in \mathcal{A}$, the principal chooses the set of incentives corresponding to action $j^*_t$ that brings the highest $\widetilde{V}(j, \widehat{\mathbf{s}}_t; \widehat{\boldsymbol{\theta}}_t)$ value. The chosen vector of incentives is denoted by $\mathbf{c}(\widehat{\boldsymbol{\theta}}_t, \widehat{\mathbf{s}}_t) = (c_a(\widehat{\boldsymbol{\theta}}_t, \widehat{\mathbf{s}}_t))_{a \in \mathcal{A}}$ such that $c_{j^*_t}(\widehat{\boldsymbol{\theta}}_t, \widehat{\mathbf{s}}_t) = \left(\textstyle \max_{a \in \mathcal{A}} \widehat{s}_{t,a}\right) - \widehat{s}_{t,j^*_t} + 2\beta_t$ and $c_a(\widehat{\boldsymbol{\theta}}_t, \widehat{\mathbf{s}}_t) = 0, \forall a \neq j^*_t$ where $j^*_t \in \argmax_{j \in \mathcal{A}} \widetilde{V}(j, \widehat{\mathbf{s}}_t; \widehat{\boldsymbol{\theta}}_t)$. \textcolor{black}{We recall that the incremental amount $2\beta_t$ is interpreted as our confidence interval for the principal’s estimation errors. We show that the design of these exploitation incentives is purposeful: by adding this increment, we guarantee that the targeted arm $j^*_t$ receives an incentive \textit{beyond the confidence interval}, so that $j^*_t$ becomes the agent’s reward-maximizing action with high probability. We formalize this property in Proposition \ref{prop4regret} in the next subsection.}

At the end of each time period, the principal provides the selected incentives $\boldsymbol{\pi}_t$ to the agent and observes the reward-maximizer arm $i_t(\boldsymbol{\pi}_t)$ chosen by the agent. As a result, the principal receives a net reward of $\mu_{t,i_t(\boldsymbol{\pi}_t)} - {\color{black} \pi_{t, i_t(\boldsymbol{\pi}_t)}} $, and the agent collects a total reward of $s^0_{i_t(\boldsymbol{\pi}_t)} + \pi_{t, i_t(\boldsymbol{\pi}_t)}$. We reiterate that the principal does not observe the agent's reward associated with the chosen action.

\begin{remark} The arithmetic operations performed to compute the exploitation incentives in lines \ref{alg-steps1}-\ref{alg-stepslast} of Algorithm \ref{alg:epsilon} have a complexity of $O(n)$ where $n = |\mathcal{A}|$. This implies that the computational complexity of the principal's bandit algorithm is linear in the dimension of the agent's model. 
\end{remark}

\subsection{Regret Bound}\label{subsec:regret}
We compute the regret of a policy $\Pi_{\epsilon, T} = \{\boldsymbol{\pi}_t\}_{t\in \mathcal{T}}$ generated by Algorithm \ref{alg:epsilon} by comparing it with an \textit{oracle} incentive policy with respect to the cumulative expected net reward obtained by the principal. An oracle incentive policy is defined as the policy with perfect knowledge of all the reward expectations $\boldsymbol{\theta}^0$ and $\mathbf{s}^0$. 
Let $\mathbf{c}(\boldsymbol{\theta}^0, \mathbf{s}^0)$ be the constant oracle incentives that maximize the principal's expected net reward at each time step over the time horizon $\mathcal{T}$. The oracle incentives are computed in a similar way to the computation of the exploitation incentives in Algorithm \ref{alg:epsilon}. We first solve for the minimum incentives required to make an action $j \in \mathcal{A}$ the reward-maximizer action of the agent and compute the associated expected net reward value $\widetilde{V}(j, \mathbf{s}^0;\boldsymbol{\theta}^0)$ as follows: 
\begin{align}
    &\widetilde{c}^j_j = \left(\max_{a \in \mathcal{A}} s^0_a \right) - s^0_j \allowdisplaybreaks  \\
    &\widetilde{c}^j_a = 0, \quad \forall a \in \mathcal{A}, \ a \neq j \allowdisplaybreaks  \\
    &\widetilde{V}(j, \mathbf{s}^0;\boldsymbol{\theta}^0) = \theta^0_j - {\color{black} \widetilde{c}^j_j} = \theta^0_j - \left(\max_{a \in \mathcal{A}} s^0_a \right) + s^0_j
\end{align}
Then, the oracle policy chooses the set of incentives corresponding to the agent action $j^{*,0}$ that has the highest $\widetilde{V}(j, \mathbf{s}^0;\boldsymbol{\theta}^0)$ value, that is $j^{*,0} := \argmax_{j \in \mathcal{A}} \widetilde{V}(j, \mathbf{s}^0;\boldsymbol{\theta}^0)$. We note that by construction of the oracle incentives, this action is same as the action that maximizes the agent's total reward after the incentives, i.e., $j^{*,0} = i(\mathbf{c}(\boldsymbol{\theta}^0, \mathbf{s}^0)) = \argmax_{a \in \mathcal{A}} s^0_a + c_a(\boldsymbol{\theta}^0, \mathbf{s}^0)$ where 
\begin{align}
    &c_{i(\mathbf{c}(\boldsymbol{\theta}^0, \mathbf{s}^0))}(\boldsymbol{\theta}^0, \mathbf{s}^0) = \max_{a \in \mathcal{A}} s^0_a  - s^0_{i(\mathbf{c}^0(\boldsymbol{\theta}^0, \mathbf{s}^0))} + \varsigma \allowdisplaybreaks  \label{eq:oracleincentive1} \\
    &c_a(\boldsymbol{\theta}^0, \mathbf{s}^0) =  0, \quad \forall a \neq i(\mathbf{c}(\boldsymbol{\theta}^0, \mathbf{s}^0)). \label{eq:oracleincentive2}\allowdisplaybreaks
\end{align}
for a sufficiently small constant $\varsigma > 0$ which helps avoiding the occurrence of multiple maximizer actions for the agent. Then, the principal's expected net reward at any time step under the oracle policy is given as 
\begin{align}
    V(\mathbf{c}(\boldsymbol{\theta}^0, \mathbf{s}^0); \boldsymbol{\theta}^0) &= \theta^0_{i(\mathbf{c}(\boldsymbol{\theta}^0, \mathbf{s}^0))} - \max_{a \in \mathcal{A}} s^0_a  + s^0_{i(\mathbf{c}(\boldsymbol{\theta}^0, \mathbf{s}^0))} - \varsigma. \label{eq:deter-oracle} \allowdisplaybreaks 
\end{align}
Similarly, we compute $V_t(\boldsymbol{\pi}_t; \boldsymbol{\theta}^0)$ as the expected net reward of the principal at time $t$ under the incentives generated by Algorithm \ref{alg:epsilon} as  
\begin{align}
    V_t(\boldsymbol{\pi}_t; \boldsymbol{\theta}^0) = \theta^0_{i_t(\boldsymbol{\pi}_t)} - {\color{black}  \pi_{t,i_t(\boldsymbol{\pi}_t)} }\label{eq:deter-epsilon}
\end{align}
where $i_t(\boldsymbol{\pi}_t)$ is as given in line (\ref{alg-action}) of Algorithm \ref{alg:epsilon}.
Lastly, we define the regret of a policy $\Pi_{\epsilon, T} = \{\boldsymbol{\pi}_t\}_{t\in \mathcal{T}}$ with respect to the cumulative expected net reward obtained by the principal. 
\begin{align}
    \mathrm{Regret}\left(\Pi_{\epsilon, T} \right) &= \sum_{t \in \mathcal{T}} V(\mathbf{c}(\boldsymbol{\theta}^0, \mathbf{s}^0); \boldsymbol{\theta}^0) - V_t(\boldsymbol{\pi}_t; \boldsymbol{\theta}^0)  \label{eq:regretdefn}
\end{align}

We provide a rigorous regret bound for the principal's $\epsilon$-greedy algorithm in Theorem \ref{thm:regret}. We next present several intermediate theoretical results that will be used to prove our regret bound. 
\begin{lemma} \label{lemma4regret}
Let $\mathcal{T}^{\mathrm{xplore}} \in \mathcal{T}$ and $\mathcal{T}^{\mathrm{xploit}} \in \mathcal{T}$ be the set of random time steps that Algorithm \ref{alg:epsilon} performs exploration (lines \ref{alg-explore1}-\ref{alg-explorelast}) and exploitation (lines \ref{alg-exploit1}-\ref{alg-exploitlast}), respectively. Then, the following probability bound holds at any $t \in \mathcal{T}^{\mathrm{xploit}}$:
\begin{multline}
        \mathbb{P}\left(\max_{a \in \mathcal{A}} \widehat{s}_{t,a} - \widehat{s}_{t,j^*_t} + 2\beta_t \geq \max_{a \in \mathcal{A}} s^0_{a} - s^0_{j^*_t} \right) \\ > 1 - \color{black} \exp \left(- \alpha (\eta(1, t)-1) \beta_t^2 - n \log \beta_t + n \log (2R_{\max} - 2R_{\min}) \right) \allowdisplaybreaks
\end{multline}
where $\eta(1, t) = | \{\tau: 1 \leq \tau \leq t-1, \tau \in \mathcal{T}^{\mathrm{xplore}} \} |$ as introduced in Theorem \ref{thm:concen}. 
\end{lemma}

The main observation required for the proof of this lemma is that the desired event is implied by the event $\|\mathbf{s}^0 - \widehat{\mathbf{s}}_t \|_\infty \leq \beta_t$. Hence, the lower bound on the probability that the desired event holds is directly obtained by using the result of Corollary \ref{cor:concen}. 
\begin{proposition} \label{prop4regret}
At any time $t \in \mathcal{T}^{\mathrm{xploit}}$, the probability that the agent will pick arm $j^*_t$ after the exploitation incentives $\mathbf{c}(\widehat{\boldsymbol{\theta}}_t, \widehat{\mathbf{s}}_t)$ is bounded by 
\begin{align}
    \mathbb{P} \left(j^*_t = i_t(\mathbf{c}(\widehat{\boldsymbol{\theta}}_t, \widehat{\mathbf{s}}_t))\right)  
    &> 1 - \color{black} \exp \left(- \alpha (\eta(1, t)-1) \beta_t^2 - n \log \beta_t + n \log (2R_{\max} - 2R_{\min}) \right)\color{black}. \allowdisplaybreaks 
\end{align}
\end{proposition}

We recall that the principal estimates that the action $j^*_t$ will yield the highest expected net reward to themselves, and hence desires that $j^*_t$ will be chosen by the agent after observing the exploitation incentives. From this perspective, the implication of the last result is that the exploitation incentives are successful in making $j^*_t$ the total reward maximizer action for the agent with high probability. This result is proved in a straightforward way by using the definition of our exploitation incentives and the result of Lemma \ref{lemma4regret}. 
\begin{proposition}\label{prop4regret-2}
Suppose $\beta_t = \sqrt{\frac{\log (\eta(1, t)-1)}{\alpha (\eta(1, t)-1)}}$ for all $t \in \mathcal{T}$. Then, we have
\begin{align}
    \mathbb{P}\left(i(\mathbf{c}(\boldsymbol{\theta}^0, \mathbf{s}^0)) \neq i_t(\mathbf{c}(\widehat{\boldsymbol{\theta}}_t, \widehat{\mathbf{s}}_t))\right) &\leq \frac{4n}{\eta(1, t)-1} + \color{black}  \frac{2n(2R_{\max} - 2R_{\min})^n \alpha^{n/2} }{[(\eta(1, t)-1) \log (\eta(1, t)-1)]^{n/2}} \color{black}. \allowdisplaybreaks 
\end{align}
\end{proposition}

This result shows a decreasing (over time) upper bound on the probability that the action selected by the agent under the exploitation incentives $\mathbf{c}(\widehat{\boldsymbol{\theta}}_t, \widehat{\mathbf{s}}_t)$ will not be the true reward-maximizer action that would be selected by the agent under the oracle incentives $\mathbf{c}(\boldsymbol{\theta}^0, \mathbf{s}^0)$. The proof follows by mainly using the finite-sample concentration bounds for the principal's estimates $\widehat{\boldsymbol{\theta}}_t$ and $\widehat{\mathbf{s}}_t$ and the result of Proposition \ref{prop4regret}. 

\begin{theorem} \textbf{\textit{(Finite-Sample Regret Bound)}}\label{thm:regret}
The regret of a policy $\Pi_{\epsilon, T}$ computed by the Principal's $\epsilon$-Greedy Algorithm (\ref{alg:epsilon}) is bounded by
\begin{align}
     \mathrm{Regret}\left(\Pi_{\epsilon, T} \right) &\leq \frac{8}{\sqrt{\alpha}} \sqrt{T \log T} + \color{black} 4n \color{black} \left(\color{black} 2 \color{black}\overline{C} - \color{black} 2 \color{black} \underline{C} + \mathrm{diam}(\Theta)\right)\color{black} (2R_{\max} - 2R_{\min})^n \alpha^{n/2} \color{black} \sqrt{T} \notag \\
     &\quad + {\color{black} (m + 4n) \left(\overline{C}  - \underline{C} + \mathrm{diam}(\Theta)\right)}\log T \notag \\ 
     &\quad + m\left( {\color{black} (\overline{C}  - \underline{C})} + \mathrm{diam}(\Theta)\right) + B_1 + B_2 \allowdisplaybreaks
\end{align}
where $B_1 + B_2 = \frac{4}{\sqrt{\alpha}}\sqrt{\frac{\log(m-1)}{m-1}} + \frac{2n \left(2(\overline{C} - \underline{C}) + \mathrm{diam}(\Theta)\right)\color{black} (2R_{\max} - 2R_{\min})^n \alpha^{n/2} \color{black}}{\sqrt{m-2}} + \frac{4n\left(\overline{C} - \underline{C} + \mathrm{diam}(\Theta)\right)}{m-1}$ and $\mathrm{diam}(\Theta) = \max_{a, a' \in \mathcal{A}} \theta^0_a - \theta^0_{a'}$ are finite and strictly positive constants. 
\end{theorem}
\begin{remark}
    This finite-sample regret bound corresponds to an asymptotic regret at a rate of order $O(\sqrt{T \log T})$ for the proposed adaptive incentive design framework. 
\end{remark} 
The proof details for all the results in this section can be found in Appendix \ref{appendix2}. 
\section{Agent's Information Rent} \label{sec:agent}
In this section, we present a discussion of our sequential principal-agent game from the agent's perspective. {\color{black} As in all adverse selection models, the information asymmetry underlying the hidden-rewards setting implies the existence of an inherent, strictly positive information rent for the agent \citep{martimort2009theory}. Given this fact, the proposed approach focuses on addressing the principal’s problem of minimizing the cumulative payments to the agent beyond this unavoidable minimum rent.} 

The way we design the principal's exploitation incentives in (\ref{eq:incentive1})-(\ref{eq:incentive3}) allows the principal to achieve this goal. Assuming that the agent picks their actions with respect to a fixed expected reward vector (that is only known by the agent), \textcolor{black}{we show in Proposition \ref{prop4regret} that the designed incentives ensures a high probability that the agent picks the arm that the principal wants them to pick}. However, the agent could just pretend that their true rewards $\mathbf{s}^0$ are different from the beginning of the sequential game, and pick all their actions in accordance with these ``pretended'' rewards in order to extract a higher information rent from the principal and maximize their total rewards. Under the hidden rewards setting, there is no way for the principal to prohibit the agent from this misbehavior which allows them to maximize the information rent they collect from the principal as we show in this section. We also note that avoiding this extra information rent could be possible in other principal-agent designs where more information about the agent's reward model is accessible by the principal. For instance, the principal could know in advance the discrete set of the agent's mean reward values without necessarily knowing which value belongs to which action. Analyzing such settings in which the principal would be able to offer incentives that get the agent to reveal their true preferences is beyond the scope of this paper, yet it stands as an interesting future research direction. 

From the standpoint of the reward-maximizer agent, we can formalize the agent's problem as an optimization model that maximizes the information rent they are extracting from the principal. The main observation here is that the maximum possible value of the agent's information rent is finite and can be achieved by a sophisticated agent who is also knowledgeable about the principal's rewards \textcolor{black}{and policy structure}. 
Recall that the principal offers the incentives that will induce the agent to pick the action which would yield the highest net expected reward to the principal. Assuming that the agent is informed about $\boldsymbol{\theta}^0$ and $\mathbf{s}^0$, they could demand extra payment from the principal by taking their actions with respect to a fixed pretended mean reward vector $\check{\mathbf{s}}(\mathbf{s}^0, \boldsymbol{\theta}^0)$ throughout the entire time horizon. We next formalize this idea in the following optimization problem.
\begin{equation}\label{agentproblem}
\begin{aligned}
    \check{\mathbf{s}}(\mathbf{s}^0, \boldsymbol{\theta}^0) \hspace{0.5em} \in  \hspace{0.5em} &\argmax_{\mathbf{s}, \boldsymbol{\pi}} \hspace{0.5em} s^0_a + \pi_a \allowdisplaybreaks \\
    &\hspace{0.2em} \mathrm{s.t.} \hspace{0.2em} a = \argmax_{a'  \in \mathcal{A}} \theta^0_{a'} - \pi_{a'} \allowdisplaybreaks \\
    &\hspace{1.7em} \pi_{a} > 0, \ \pi_a \in \mathcal{C}, \ \pi_{a'} = 0 \ \forall a' \in \mathcal{A} \setminus \{a\} \allowdisplaybreaks  \\
    &\hspace{1.7em} b = \argmax_{a' \in \mathcal{A}} s_{a'}  + \pi_{a'} \allowdisplaybreaks  \\
    &\hspace{1.7em} a = b  \allowdisplaybreaks 
\end{aligned}
\end{equation}
The objective function of this optimization problem maximizes the agent's true expected reward (after the incentives) obtained from selecting an action $a$, which is further specified by the constraints. The first constraint implies that action $a$ maximizes the principal's expected net reward when the incentives $\boldsymbol{\pi}$ are selected as given in the second set of constraints. Then, the third and last constraints ensure that the incentives are designed in such a way that action $a$ is also the reward-maximizer for the agent who pretend their rewards as $\check{\mathbf{s}}$. 

\begin{proposition} \label{prop:informationrent}
    The agent's optimization problem (\ref{agentproblem}) is feasible, and the agent can maximize their information rent by choosing its solution $\check{\mathbf{s}}(\mathbf{s}^0, \boldsymbol{\theta}^0)$ as their ``pretended'' fixed mean reward vector during the course of their repeated play with the principal. 
\end{proposition}

The complete proof of this proposition is provided in Appendix \ref{appendix3}. Recall that in Section \ref{subsec:regret}, we show that when the agent plays in accordance with their true mean reward vector $\mathbf{s}^0$ and the principal follows the oracle incentive policy $\mathbf{c}(\boldsymbol{\theta}^0, \mathbf{s}^0)$, then the agent gets their minimum possible expected total reward. We start the proof by showing that this solution is feasible to the problem (\ref{agentproblem}), yet it yields the worst-case result for the agent. We continue by proving the existence of other feasible solutions which use mean reward vectors that are different than $\mathbf{s}^0$ and return higher information rents to the agent. These feasible solutions are proposed for two mutually exclusive cases based on whether the maximizer actions of the principal's and the agent's mean rewards, $\boldsymbol{\theta}^0$ and $\mathbf{s}^0$, are the same as each other or not. We next present two numerical examples that illustrate the feasible solutions proposed in the proof for each of these two cases.   

\begin{example}
Consider a model with three actions $\mathcal{A} = \{1, 2, 3\}$. Let the agent's true mean reward vector be $\mathbf{s}^0 = (s^0_1, s^0_2, s^0_3) = (0, 4, 3)$ and the principal's true mean reward vector be $\boldsymbol{\theta}^0 = (\theta^0_1, \theta^0_2, \theta^0_3) = (1, 8, 2)$. Notice that the principal does not need to incentivize the agent in this case because the reward-maximizer actions for both parties are the same with each other. The principal can just offer the incentives $\boldsymbol{\pi} = (0, 0, 0)$ that yield the highest possible expected net reward to them (which is 8) and the worst-case expected total reward to the agent (which is 4). 

Now, suppose that the agent plays according to the rewards $\mathbf{s} = (0, 4, 9.5)$. In that case, if the principal offers the same incentives $\boldsymbol{\pi}$, then the agent will pick the third action and the principal's expected net reward will be 2. However, the principal can obtain a relatively higher expected net reward by offering a different set of incentives that will get the agent to pick the second action. Suppose that the principal gives the incentives $\widetilde{\boldsymbol{\pi}} = (0, \color{black}5.5\color{black}, 0)$, which is a feasible solution to the agent's optimization problem together with the chosen $\mathbf{s}$. Then, the expected net reward of the principal becomes $\theta^0_2 -\widetilde{\pi}_2 = 2.5$ whereas the agent's expected total reward jumps to $s^0_2 + \widetilde{\pi}_2 = 9.5$. As a result, the agent collects an extra information rent of $9.5 - 4 = 5.5$ which is the difference between their expected total rewards when they are playing with $\mathbf{s}^0$ and when they are pretending their rewards are $\mathbf{s}$. 
\end{example}

\begin{example}
Consider a model with four actions $\mathcal{A} = \{1, 2, 3, 4\}$. Let the agent's true mean reward vector be $\mathbf{s}^0 = (s^0_1, s^0_2, s^0_3, s^0_4) = (0, 4, 3, 6)$ and the principal's true mean reward vector be $\boldsymbol{\theta}^0 = (\theta^0_1, \theta^0_2, \theta^0_3, \theta^0_4) = (1, 8, 7, 2)$. If the agent plays in accordance with their true rewards, then $\mathbf{s} = \mathbf{s}^0$ and $\boldsymbol{\pi} = (0, 2.1, 0, 0)$ will yield a feasible solution to (\ref{agentproblem}) with $a = b = 2$. With this solution, the principal's expected net reward will be $\theta^0_2 - \pi_2 = 8 - 2.1 = 5.9$ and the agent's expected total reward will be $s^0_2 + \pi_2 = 4 + 2.1 = 6.1$. On the other hand, consider the rewards $\mathbf{s} = (0, 4, 3, 7.8)$ and the incentives $\widetilde{\boldsymbol{\pi}} = (0, 3.9, 0, 0)$. These vectors result in another feasible solution in which the principal's expected net reward decreases to $\theta^0_2 - \widetilde{\pi}_2 = 8 - 3.9 = 4.1$ whereas the agent's expected total reward rises to $s^0_2 + \widetilde{\pi}_2 = 4 + 3.9 = 7.9$. As can be seen, the agent gains a higher information rent in this case by pretending their rewards are $\mathbf{s}$ and capturing an extra amount of $1.8$ from the principal's expected profits. 
\end{example}
As stated above, achieving the maximum information rent would require a significant amount of sophistication from the agent (requiring complete knowledge of $\boldsymbol{\theta}^0$ \textcolor{black}{and the policy structure}), which is often not the case in practice. Since the agent is typically less knowledgeable about the principal's model, they are unlikely to obtain information rent exceeding the inherent minimum amount. However, regardless of their level of knowledge, we show that the agent benefits \textcolor{black}{from playing the game} based on a fixed vector of mean rewards. \textcolor{black}{Whether the vector is the true one or a pretended one, the resulting actions remain consistent with that reward vector throughout the entire time horizon. This aligns with the underlying assumption of our model, which is also the standard assumption in MAB formulations \citep{auer2002finite, bubeck2012regret, lattimore2020bandit}. Therefore, we highlight that our learning framework is designed to minimize the principal's cumulative regret at the end of the game subject to some information rent that the principal must give up.}  
\section{Numerical Experiments}\label{sec:numerical}
\color{black} In this section, we aim to support our theoretical results with numerical experiments in which the proposed data-driven incentives are compared with the derived oracle ones. Our experimental setting is based on an instance of the personalized medical adherence incentive design problem introduced in Section \ref{subsubsec:medical}. In particular, our design builds on the case study of \cite{lobo2017using}, where they investigate the adherence interventions for statin treatment that are shown to be crucial for type 2 diabetes, yet characterized by persistently low long-term adherence rates. 

\subsection{Experimental Setting}
In their framework, patient behavior is modeled using a Markov decision process (MDP) where adherence decisions evolve stochastically over time and determine both health benefits and downstream medical costs. We adapt this setting into our repeated game, where the medical provider (principal) offers adherence-improving incentives to influence the patient’s (agent’s) sequential adherence decisions. In this reformulation, the provider’s utility is derived from reduced long-term health care expenditures net of incentive payments, while the patient’s utility balances the perceived health benefits of adherence minus its associated burdens. This transformation preserves the clinical realism of the original statin adherence MDP while enabling us to analyze adaptive incentive design and learning dynamics through the lens of regret minimization.

We assume that adherence is reviewed and incentives are delivered on a weekly basis. This choice is motivated by the clinical context: although statins are prescribed as daily medications and typically refilled on a monthly cycle, adherence monitoring (via pill counts, electronic devices, or self-reports) is often conducted more frequently over shorter windows (such as the proportion of days covered within a week) to capture dynamic fluctuations that would be obscured at the monthly level \citep{Gibson2006, curtis2020prescription}. Accordingly, we define each period $t$ as one week, with the time horizon set to $T = 2100$ weeks ($\approx$ 40 years). This corresponds to the interval between age 40—the reported optimal initiation age for patients regardless of gender or health status \citep{mason2012optimizing}—and age 80, after which interventions are typically discontinued.

Following \cite{lobo2017using}, we discretize adherence into four levels corresponding to four patient arms $\mathcal{A}=\{\text{NON} ([0,10]\%), \text{LOW} ([10,40]\%), \text{MED} ([40,80]\%), \text{HIGH}([80,100]\%)\}$, with mid-point anchors
$\delta_{\text{NON}}=0.05$, $\delta_{\text{LOW}}=0.25$, $\delta_{\text{MED}}=0.60$, $\delta_{\text{HIGH}}=0.90$. At week $t$, the provider offers incentives $\boldsymbol{\pi}_t$, and the patient selects adherence level $i_t(\boldsymbol{\pi}_t)$, yielding the provider an expected total cost of $ - \theta_{i_t(\boldsymbol{\pi}_t)} - \pi_{t, i_t(\boldsymbol{\pi}_t)}$, where $\theta$ represents expected downstream health expenditures (e.g., hospitalizations, cardiovascular complications). (Our setup also accommodates stochastic costs by treating expected rewards as the negative of costs.) To ensure non-negativity while preserving sub-Gaussian properties, stochastic costs for each adherence arm $a$ are modeled with a zero-truncated Normal distribution, $\mathcal{N}_{+}(\theta^0_a, (\sigma^{\text{provider}}_a)^2)$, a standard specification in health econometrics for estimating resource costs \citep{Kurz2017, Lambert2008, Mihaylova2011, French2004}. We calibrate these parameters using data from \cite{pittman2011adherence}, which reports total healthcare costs across various statin adherence levels through a large cohort study. We detail our calibration procedure and present the obtained provider parameters in Appendix \ref{appendix4-1}. 

On the patient side, we follow the design of \cite{lobo2017using} (see their Appendix A.1 for further discussion) and model patient's expected reward for adherence $i_t(\boldsymbol{\pi}_t)$ as $r_{i_t(\boldsymbol{\pi}_t)} = \mathbb{E}k^\mathrm{BNF} \times \delta(i_t(\boldsymbol{\pi}_t)) - k^\mathrm{DIS} \times  \delta^2(i_t(\boldsymbol{\pi}_t))$. Here, the first term reflects expected perceived monetary benefit from adherence while the latter captures the disutility from treatment burden and its side effects. For the disutility part, while \cite{lobo2017using} consider only the direct cost of medication, we aim to capture the overall disutility of adherence, encompassing both monetary costs and non-monetary opportunity costs (e.g., time, effort). We therefore adopt a more general quadratic disutility function with a constant of $k^\mathrm{DIS} = $212/\text{year} = $4.07/\text{week}$, as reported in \cite{lobo2017using}. On the benefit side, we define $\mathbb{E}k^\mathrm{BNF} = \mathrm{WTP} \times \mathbb{E}q_t$, where WTP denotes the willingness-to-pay per quality-adjusted life year (QALY) gained, and $\mathbb{E}q_t$ is the expected marginal health gain from adherence (in QALYs). Following standard convention, we treat WTP as fixed and known \citep{evans2004use}, while patient heterogeneity is captured through variation in $\mathbb{E}q_t$. Patients differ substantially in their perceived benefit from adherence: some strongly value an additional life year, while others assign little or even negative value. This heterogeneity is also visible in the steady-state adherence distribution reported by \cite{mason2012optimizing}, where more than 45\% of the population remained in the NON adherence state after 20 years, indicating persistently poor long-term adherence to statin therapy.

To calibrate this data into our model, we suppose that a single patient's $\mathbb{E}q_t$ is drawn from a Normal distribution representing the population's statin adherence perception distribution. This yields $\mathbb{E}k^\mathrm{BNF} \sim \mathcal{N}(2.0808, (12.9817)^2)$ dollars per unit adherence per week. The detailed explanation of our calibration process is given in Appendix \ref{appendix4-2}. Notably, the relatively large variance ensures consistency with the observed heterogeneity in long-run adherence shares. Although this distribution allows a non-trivial probability of negative perceived benefits, this reflects that a sizable fraction of patients in the data persist in low or non-adherence despite potential health gains.

This implies that the patient's (unknown) expected perceived benefit from adherence (i.e., $\mathbb{E}k^\mathrm{BNF}$) governs how strongly provider can influence adherence choices. To evaluate this performance, we conduct experiments in renewal episodes. Each episode begins with a new iid patient, whose $\mathbb{E}k^\mathrm{BNF}$ is drawn from the calibrated population (with the same provider parameters $\boldsymbol{\theta}$), and runs until the fixed lifetime horizon $T$. At termination, the environment resets, and a new patient starts immediately, producing a continuous sequence of patients under the same per-period data-generating process. For each experiment, we generate a sequence of ten iid patients and replicate the experiment (with the same sequence) ten times. The specific model parameters used in the experiments are reported in Appendix \ref{appendix4-3}. Policy performance statistics are computed at the episode level and then aggregated across episodes and replications.

\subsection{Results and Discussion} 
The main contribution of adopting our proposed framework to this well-known long-run medical adherence problem is to explicitly examine the trade-off between the provider’s perspective of minimizing the cost of interventions, medication, and adverse events, while effectively changing patient's perceived net benefit. As can be seen from the cited works along with many others, although this long-term adherence problem has been well-studied to understand the stochastic adherence and intervention trajectories, the potential learning dynamics that could inform those sequential interventions is missing. Our results show a promising fit of learning-based incentive frameworks to this context to better navigate the provider's trade-off.  

\subsubsection{Finite-Sample Concentration Results (Theorem \ref{thm:concen})}

Although the regret performance presented in the next subsection indicates consistent learning behavior of the proposed algorithm, we start by providing an  explicit assessment of the validity of the concentration bound in Theorem \ref{thm:concen}. 

\begin{figure}
\centering
\includegraphics[width=.9\textwidth]{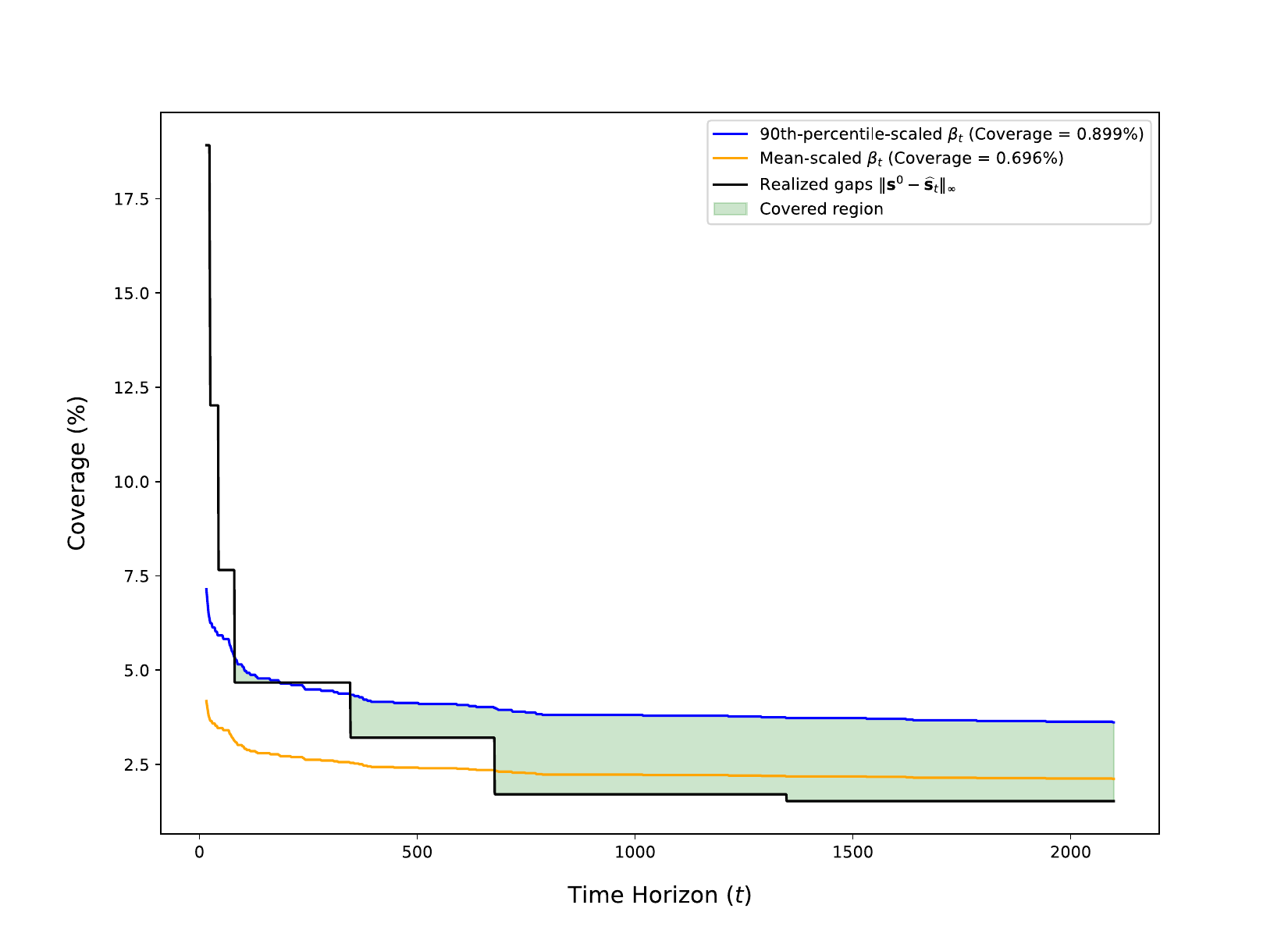}
\caption{\color{black} Realized estimation gaps compared with scaled theoretical tolerances, illustrating the rate, conservativeness, and coverage of the concentration bound. Shaded areas indicate regions where the realized gaps lie below the scaled tolerance. }
\label{fig:beta}
\end{figure}

Figure \ref{fig:beta} displays the trajectories of the realized estimation gaps $\|\mathbf{s}^0 - \widehat{\mathbf{s}}_t \|_\infty$ alongside the theoretical tolerance trajectory $\beta_t$ given in line~\ref{alg:beta}, after rescaling the latter for comparability. This analysis highlights three points. First, the sequence of our estimates $\{\widehat{\mathbf{s}}_t\}_{t \in \mathcal{T}}$ consistently converges toward the true rewards $\mathbf{s}^0$ as the treatment proceeds. Second, because the theoretical bound is conservative in magnitude, we rescale $\beta_t$ using two global factors: one based on the mean of the ratios $\|\mathbf{s}^0 - \widehat{\mathbf{s}}_t \|_\infty / \beta_t$, which reflects the typical discrepancy between realized errors and tolerance, and one based on the 90th percentile of these ratios, which provides a high-coverage adjustment without being dominated by a single extreme outlier. These scaling constants quantify the conservativeness of the bound. After accounting for these constant-level adjustments, the figure shows that our concentration bound captures the correct rate of error decay. Third, we report the \textit{coverage} levels, defined as the fraction of treatment weeks in which $\|\mathbf{s}^0 - \widehat{\mathbf{s}}_t \|_\infty \leq \text{scaled-}\beta_t$ under each scaling scheme. These coverage results demonstrate that the realized errors remain well controlled relative to the scaled tolerance, supporting the practical reliability of the proven concentration inequality. 
\begin{figure}
\centering
\includegraphics[width=.9\textwidth]{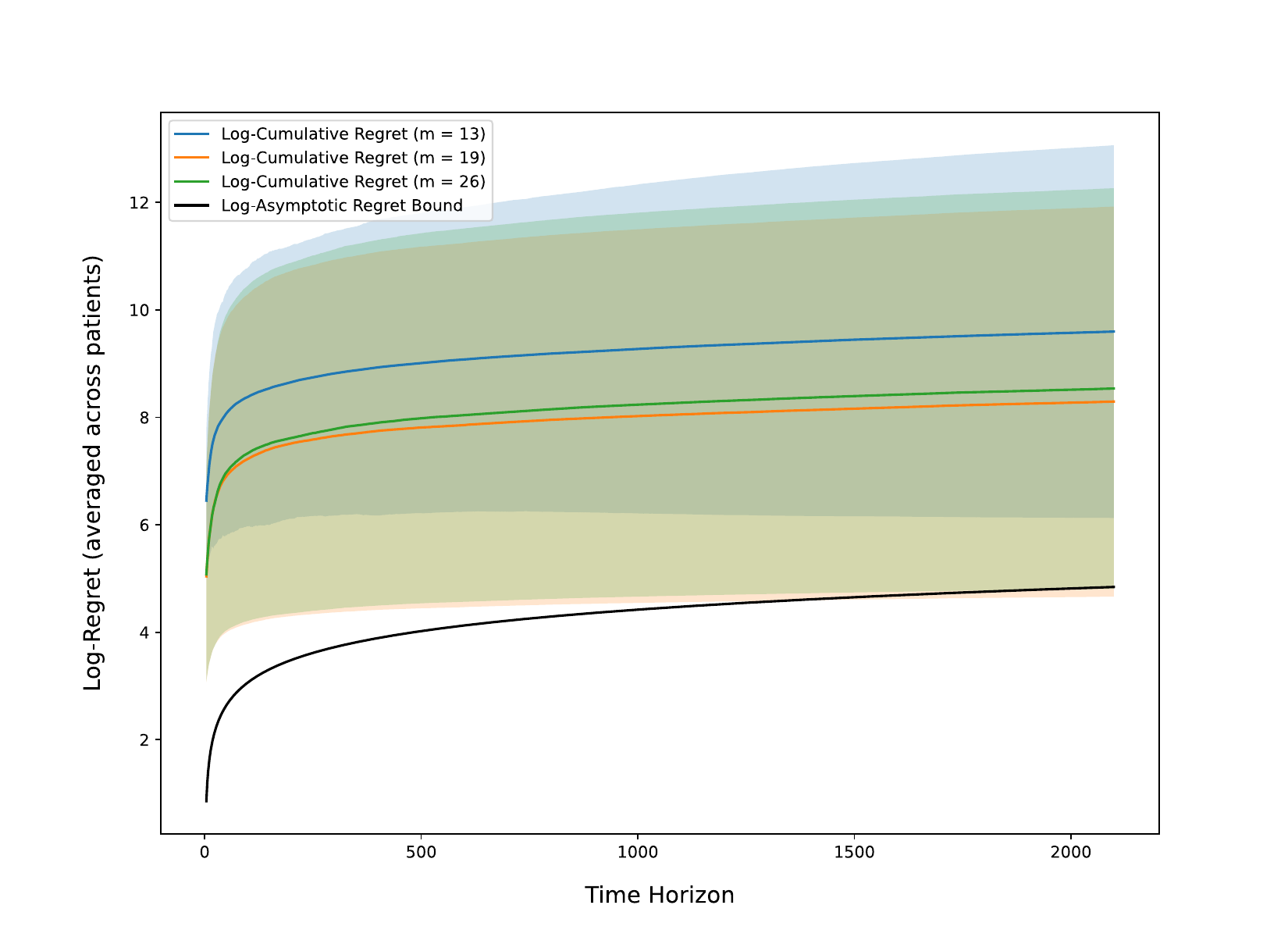}
\caption{\color{black} The provider's log-cumulative regret averaged across iid patients with exploration rates $m \in \{13,19,26\}$. The shaded regions represent the standard error over all replications.}
\label{fig:regret}
\end{figure}
\subsubsection{Finite-Sample Regret Results (Theorem \ref{thm:regret})}
As described in the beginning, the provider's problem embodies a classic exploration–exploitation trade-off. To examine how our algorithm balances this trade-off, we repeat the experiments under different values of the exploration rate parameter $m$ in the schedule $\epsilon_t = \min\{1,\, m/t\}$ (see line \ref{alg-m} of Algorithm \ref{alg:epsilon}). By definition, $m$ corresponds to the number of weeks during which the provider engages in pure exploration at the start of the treatment horizon (after initialization), and thus directly controls the mentioned trade-off. We conduct sensitivity analysis with $m \in \{13, 19, 26\}$, corresponding to the initial exploration phases of approximately 3, 4.5, and 6 months. These values are motivated by the statin adherence context: because prescription refills are typically monthly, $m=13$ weeks span about three refill cycles, allowing the provider to observe whether incentive responses persist across cycles, whereas exploration periods longer than six months may become ethically and operationally burdensome. 

Since the proven theoretical upper bound $O(\sqrt{T \log T})$ in Theorem \ref{thm:regret} characterizes the asymptotic growth of cumulative regret, it is not directly comparable to the finite-time regret accrued per patient in our experiments with $T = 2100$. To facilitate comparison, we plot regrets on a log scale, as is common in MAB studies \citep{besbes2019optimal, ghosh2022breaking, huang2024sublinear}. Figure \ref{fig:regret} yields two main insights. First, in all cases, although the bound is not expected to be tight for short horizons, the log scale reveals that the empirical regret exhibits a growth trend consistent with the theoretical order of growth. Second, we find that increasing the exploration phase from 13 to 19 weeks yields a substantial reduction in regret (approximately 71\%), while extending it further to 26 weeks offers no additional benefit and even worsens outcomes by raising provider intervention costs unnecessarily. These results suggest that increasing $m$ can improve performance up to a point in an instance-dependent manner, but beyond that threshold the algorithm’s learning performance remains robust, and longer pure-exploration phases might become counterproductive.

\subsubsection{Results on Accuracy of Chosen Arms (Propositions \ref{prop4regret}-\ref{prop4regret-2})} \label{subsec:experiment-arms}

\begin{figure}
\centering
\includegraphics[width=.9\textwidth]{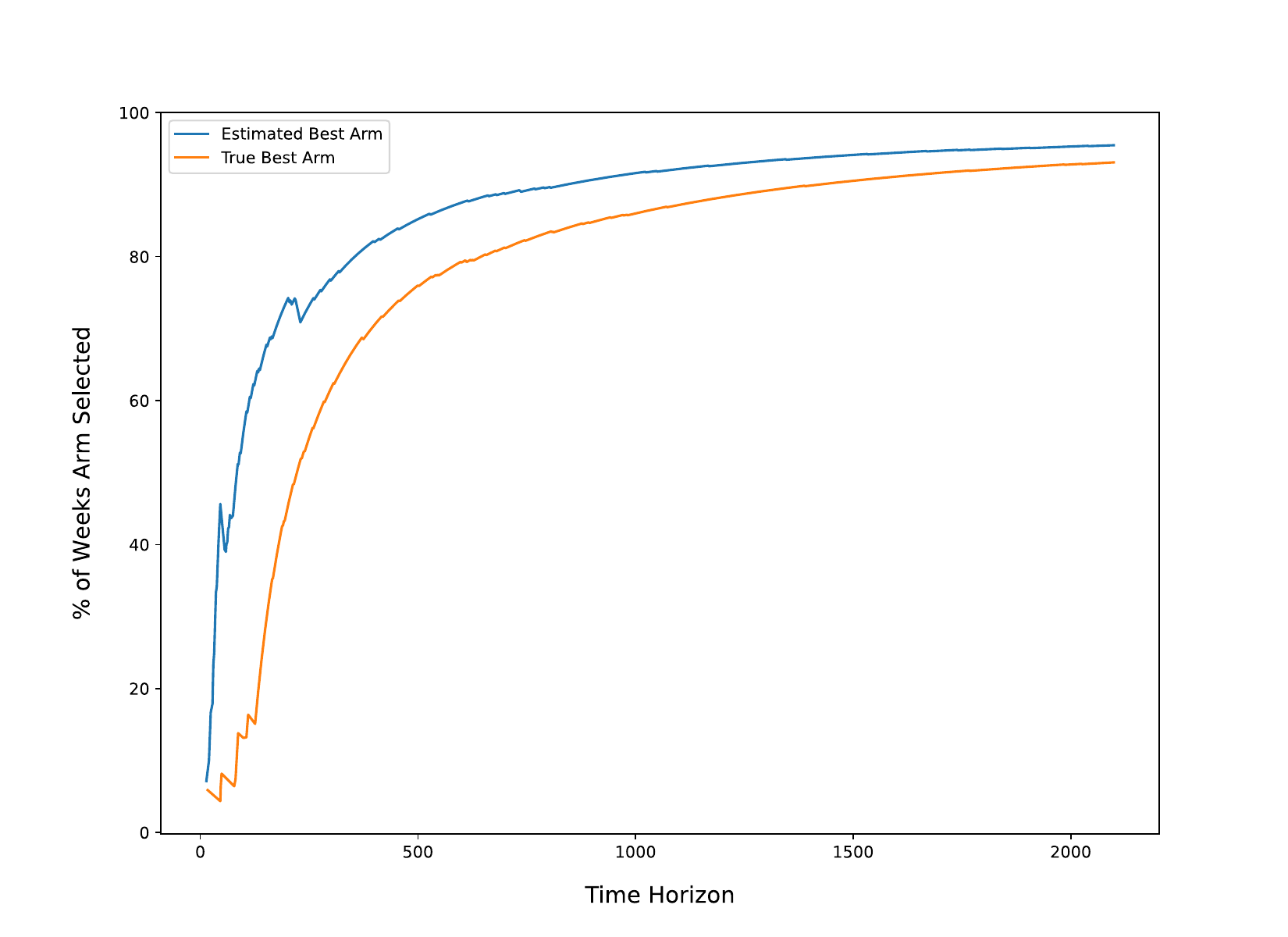}
\caption{\color{black} Cumulative percentage of weeks in which the algorithm steers the patient toward i) the estimated best adherence level, and ii) the true best level.}
\label{fig:arms}
\end{figure}

Lastly, we provide further insight into the internal dynamics of the algorithm by evaluating the provider’s success in (i) inducing the patient to adhere at the desired level under the calculated incentives, and (ii) correctly identifying the true cost-minimizing adherence level. The proof of the regret bound (see Appendix~\ref{appendix2}) relies on the occurrence of these two events, as formalized in Propositions~\ref{prop4regret}–\ref{prop4regret-2}. Figure~\ref{fig:arms} reports the cumulative proportion of weeks in which the patient chooses (i)the provider’s estimated best adherence level $j^*_t$, and (ii) the true best adherence level $i(\mathbf{c}(\boldsymbol{\theta}^0, \mathbf{s}^0))$ that would minimize expected costs under oracle knowledge. The realized frequencies of these two events across our repeated experiments provide numerical support for our theoretical arguments.

We note that the highest adherence is not necessarily optimal for the provider once incentive costs are considered. For patients with high treatment disutility (e.g., severe side effects or financial burden), sustaining full adherence would require prohibitively high incentives that outweigh savings from improved adherence. Moreover, in early learning stages, provider's inaccurate estimates are more likely to lead to misidentifying the desired levels. Together, these factors explain potential deviations of the provider’s desired levels, especially early on and for certain patients. This mirrors the provider’s trade-off between clinical and economic objectives in medical adherence context.

Despite these challenges, Figure~\ref{fig:arms} shows that the algorithm steers patients toward the provider’s desired adherence level in most periods, and that this level increasingly coincides with the true optimum. This reinforces the relevance of our model to statin adherence, where providers often face uncertainty in both the benefits of adherence and the effectiveness of interventions. Our findings suggest that even with imperfect information, a learning-based incentive scheme can not only increase the likelihood that patients follow provider guidance but also ensure that this guidance aligns with the true long-term clinical and economic outcomes. 
\color{black}
\section{Conclusion and Future Directions}\label{sec:conclusion}
We conclude by summarizing our primary contributions to the principal-agent theory and data-driven contract design literature. In this paper, we introduce a novel repeated principal-agent setting which has not been explored in earlier studies even though it is applicable to many real-life problems. In particular, we analyze an adverse selection model where the principal can solely observe the agent's decisions while the agent's true preferences and rewards stay hidden from the principal. To enhance the practical relevance of our theoretical studies, we keep our model as generic as possible. The two integrated dimensions of the considered research problem are: i) estimation of the agent's unknown reward model, and ii) design of adaptive incentives that will maximize the principal's cumulative net rewards over a finite time horizon. We first introduce our novel estimator and prove its identifiability and finite-sample concentration bound. Then, we formalize the principal's data-driven incentives and unite them with our estimator in an $\epsilon$-greedy bandit algorithm. We conduct a rigorous regret analysis for this algorithm and support our theoretical results \textcolor{black}{by evaluating the performance of our approach in simulations of a personalized medical adherence setting, calibrated using real-world data}.

We also highlight possible future work directions pointed by our paper. In this current work, we assume that the reward-maximizing agent has full knowledge of their reward model and is able to take the true reward-maximizer action at every period. A more challenging model would consider an agent with imperfect knowledge of their model. In that case, our analyses will also need to involve the learning process of the agent who trains their algorithm on top of the principal's learning process. As can be expected, the dynamic interactions between these two learning parties will add substantial complexity both to the estimation and the incentive design problems. Yet, we believe that analyzing this setting would be useful for studying certain practical problems such as the medical adherence application discussed in Section \ref{subsubsec:medical}. Another direction would be to consider the collaboration between a principal and multiple reward-maximizer agents. We believe that our model and approach is applicable to a multi-agent setting where the agents collectively work as a team and the principal provides team incentives based on the observed team-level decisions. On the other hand, studying a multi-agent setting where the principal needs to design incentives for each individual selfish-agent (that might be also communicating with other agents) would require a completely different approach and analysis. In addition to these directions, we suppose our paper may be extended to various scenarios that accommodate other common features of repeated principal-agent models observed in practice.

\ACKNOWLEDGMENT{This material is based upon work partially supported by the National Science Foundation under Grant CMMI-184766.}


\bibliographystyle{informs2014} 
\bibliography{PerfectAgents} 

\begin{thebibliography}{103}
\providecommand{\natexlab}[1]{#1}
\providecommand{\url}[1]{\texttt{#1}}
\providecommand{\urlprefix}{URL }

\bibitem[{Abhishek et~al.(2020)Abhishek, Jain, \protect\BIBand{} Gujar}]{abhishek2020designing}
Abhishek K, Jain S, Gujar S (2020) Designing truthful contextual multi-armed bandits based sponsored search auctions. \emph{arXiv preprint arXiv:2002.11349} .

\bibitem[{Abreu et~al.(1990)Abreu, Pearce, \protect\BIBand{} Stacchetti}]{abreu1990toward}
Abreu D, Pearce D, Stacchetti E (1990) Toward a theory of discounted repeated games with imperfect monitoring. \emph{Econometrica: Journal of the Econometric Society} 1041--1063.

\bibitem[{Ahuja \protect\BIBand{} Orlin(2001)}]{ahuja2001inverse}
Ahuja RK, Orlin JB (2001) Inverse optimization. \emph{Operations Research} 49(5):771--783.

\bibitem[{Amin et~al.(2014)Amin, Rostamizadeh, \protect\BIBand{} Syed}]{NIPS2014}
Amin K, Rostamizadeh A, Syed U (2014) Repeated contextual auctions with strategic buyers. \emph{Advances in Neural Information Processing Systems}, volume~27.

\bibitem[{Aswani(2019)}]{aswani2019statistics}
Aswani A (2019) Statistics with set-valued functions: applications to inverse approximate optimization. \emph{Mathematical Programming} 174(1-2):225--251.

\bibitem[{Aswani et~al.(2018)Aswani, Shen, \protect\BIBand{} Siddiq}]{aswani2018inverse}
Aswani A, Shen ZJ, Siddiq A (2018) Inverse optimization with noisy data. \emph{Operations Research} 66(3):870--892.

\bibitem[{Aswani et~al.(2019)Aswani, Shen, \protect\BIBand{} Siddiq}]{aswani2019data}
Aswani A, Shen ZJM, Siddiq A (2019) Data-driven incentive design in the medicare shared savings program. \emph{Operations Research} 67(4):1002--1026.

\bibitem[{Audy et~al.(2012)Audy, Lehoux, D'Amours, \protect\BIBand{} R{\"o}nnqvist}]{audy2012framework}
Audy JF, Lehoux N, D'Amours S, R{\"o}nnqvist M (2012) A framework for an efficient implementation of logistics collaborations. \emph{International transactions in operational research} 19(5):633--657.

\bibitem[{Auer et~al.(2002)Auer, Cesa-Bianchi, \protect\BIBand{} Fischer}]{auer2002finite}
Auer P, Cesa-Bianchi N, Fischer P (2002) Finite-time analysis of the multiarmed bandit problem. \emph{Machine learning} 47(2):235--256.

\bibitem[{Banks \protect\BIBand{} Sundaram(1993)}]{banks1993adverse}
Banks JS, Sundaram RK (1993) Adverse selection and moral hazard in a repeated elections model. \emph{ch} 12:295--311.

\bibitem[{B{\"a}rmann et~al.(2018)B{\"a}rmann, Martin, Pokutta, \protect\BIBand{} Schneider}]{barmann2018online}
B{\"a}rmann A, Martin A, Pokutta S, Schneider O (2018) An online-learning approach to inverse optimization. \emph{arXiv preprint arXiv:1810.12997} .

\bibitem[{Bertsimas et~al.(2015)Bertsimas, Gupta, \protect\BIBand{} Paschalidis}]{bertsimas2015data}
Bertsimas D, Gupta V, Paschalidis IC (2015) Data-driven estimation in equilibrium using inverse optimization. \emph{Mathematical Programming} 153(2):595--633.

\bibitem[{Besbes et~al.(2019)Besbes, Gur, \protect\BIBand{} Zeevi}]{besbes2019optimal}
Besbes O, Gur Y, Zeevi A (2019) Optimal exploration--exploitation in a multi-armed bandit problem with non-stationary rewards. \emph{Stochastic Systems} 9(4):319--337.

\bibitem[{Bhat et~al.(2019)Bhat, Jain, Gujar, \protect\BIBand{} Narahari}]{bhat2019optimal}
Bhat S, Jain S, Gujar S, Narahari Y (2019) An optimal bidimensional multi-armed bandit auction for multi-unit procurement. \emph{Annals of Mathematics and Artificial Intelligence} 85(1):1--19.

\bibitem[{Biswas et~al.(2015)Biswas, Jain, Mandal, \protect\BIBand{} Narahari}]{biswas2015truthful}
Biswas A, Jain S, Mandal D, Narahari Y (2015) A truthful budget feasible multi-armed bandit mechanism for crowdsourcing time critical tasks. \emph{AAMAS}, 1101--1109.

\bibitem[{Bolton \protect\BIBand{} Dewatripont(2004)}]{bolton2004contract}
Bolton P, Dewatripont M (2004) \emph{Contract theory} (MIT press).

\bibitem[{Bosworth(2010)}]{bosworth2010medication}
Bosworth HB (2010) Medication adherence. \emph{Improving patient treatment adherence}, 68--94 (Springer).

\bibitem[{Boucheron et~al.(2013)Boucheron, Lugosi, \protect\BIBand{} Massart}]{boucheron2013concentration}
Boucheron S, Lugosi G, Massart P (2013) \emph{Concentration inequalities: A nonasymptotic theory of independence} (Oxford university press).

\bibitem[{Braverman et~al.(2019)Braverman, Mao, Schneider, \protect\BIBand{} Weinberg}]{braverman2019multi}
Braverman M, Mao J, Schneider J, Weinberg SM (2019) Multi-armed bandit problems with strategic arms. \emph{Conference on Learning Theory}, 383--416 (PMLR).

\bibitem[{Brown et~al.(2016)Brown, Bussell, Dutta, Davis, Strong, \protect\BIBand{} Mathew}]{brown2016medication}
Brown MT, Bussell J, Dutta S, Davis K, Strong S, Mathew S (2016) Medication adherence: truth and consequences. \emph{The American journal of the medical sciences} 351(4):387--399.

\bibitem[{Bubeck et~al.(2012)Bubeck, Cesa-Bianchi et~al.}]{bubeck2012regret}
Bubeck S, Cesa-Bianchi N, et~al. (2012) Regret analysis of stochastic and nonstochastic multi-armed bandit problems. \emph{Foundations and Trends{\textregistered} in Machine Learning} 5(1):1--122.

\bibitem[{Busch \protect\BIBand{} Ferretti-Gallon(2017)}]{busch2017drives}
Busch J, Ferretti-Gallon K (2017) What drives deforestation and what stops it? a meta-analysis. \emph{Review of Environmental Economics and Policy} .

\bibitem[{Chade \protect\BIBand{} Swinkels(2019)}]{chade2019disentangling}
Chade H, Swinkels J (2019) Disentangling moral hazard and adverse selection. Technical report, Working Paper, Arizona State University.

\bibitem[{Chan et~al.(2022)Chan, Eberg, Forster, Holloway, Ieraci, Shalaby, \protect\BIBand{} Yousefi}]{chan2022inverse}
Chan TC, Eberg M, Forster K, Holloway C, Ieraci L, Shalaby Y, Yousefi N (2022) An inverse optimization approach to measuring clinical pathway concordance. \emph{Management Science} 68(3):1882--1903.

\bibitem[{Chan et~al.(2019)Chan, Lee, \protect\BIBand{} Terekhov}]{chan2019inverse}
Chan TC, Lee T, Terekhov D (2019) Inverse optimization: Closed-form solutions, geometry, and goodness of fit. \emph{Management Science} 65(3):1115--1135.

\bibitem[{Conitzer \protect\BIBand{} Garera(2006)}]{conitzer2006}
Conitzer V, Garera N (2006) Learning algorithms for online principal-agent problems (and selling goods online). \emph{Proceedings of the 23rd International Conference on Machine Learning}, 209–216, ICML '06.

\bibitem[{Curtis et~al.(2020)Curtis, Walker, MacKenna, Croker, \protect\BIBand{} Goldacre}]{curtis2020prescription}
Curtis HJ, Walker AJ, MacKenna B, Croker R, Goldacre B (2020) Prescription of suboptimal statin treatment regimens: a retrospective cohort study of trends and variation in english primary care. \emph{British Journal of General Practice} 70(697):e525--e533.

\bibitem[{Devanur \protect\BIBand{} Kakade(2009)}]{Devanur2009ThePO}
Devanur NR, Kakade SM (2009) The price of truthfulness for pay-per-click auctions. \emph{EC '09}.

\bibitem[{Dionne \protect\BIBand{} Lasserre(1985)}]{dionne85}
Dionne G, Lasserre P (1985) {Adverse Selection, Repeated Insurance Contracts and Announcement Strategy}. \emph{The Review of Economic Studies} 52(4):719--723, \urlprefix\url{http://dx.doi.org/10.2307/2297743}.

\bibitem[{Dong et~al.(2018)Dong, Chen, \protect\BIBand{} Zeng}]{dong2018generalized}
Dong C, Chen Y, Zeng B (2018) Generalized inverse optimization through online learning. \emph{Advances in Neural Information Processing Systems} 31.

\bibitem[{Dong \protect\BIBand{} Zeng(2020)}]{dong2020inverse}
Dong C, Zeng B (2020) Inverse multiobjective optimization through online learning. \emph{arXiv preprint arXiv:2010.06140} .

\bibitem[{Early(2011)}]{early_2011}
Early C (2011) Delivering greener logistics. \urlprefix\url{https://www.iema.net/articles/delivering-greener-logistics}.

\bibitem[{Engel et~al.(2016)}]{engel2016devil}
Engel S, et~al. (2016) The devil in the detail: a practical guide on designing payments for environmental services. \emph{International Review of Environmental and Resource Economics} 9(1--2):131--177.

\bibitem[{Ergun et~al.(2007)Ergun, Kuyzu, \protect\BIBand{} Savelsbergh}]{ergun2007reducing}
Ergun O, Kuyzu G, Savelsbergh M (2007) Reducing truckload transportation costs through collaboration. \emph{Transportation science} 41(2):206--221.

\bibitem[{Esfahani et~al.(2018)Esfahani, Shafieezadeh-Abadeh, Hanasusanto, \protect\BIBand{} Kuhn}]{esfahani2018data}
Esfahani PM, Shafieezadeh-Abadeh S, Hanasusanto GA, Kuhn D (2018) Data-driven inverse optimization with imperfect information. \emph{Mathematical Programming} 167(1):191--234.

\bibitem[{Es{\H{o}} \protect\BIBand{} Szentes(2017)}]{esHo2017dynamic}
Es{\H{o}} P, Szentes B (2017) Dynamic contracting: An irrelevance theorem. \emph{Theoretical Economics} 12(1):109--139.

\bibitem[{Evans et~al.(2004)Evans, Tavakoli, \protect\BIBand{} Crawford}]{evans2004use}
Evans C, Tavakoli M, Crawford B (2004) Use of quality adjusted life years and life years gained as benchmarks in economic evaluations: a critical appraisal. \emph{Health care management science} 7(1):43--49.

\bibitem[{Food \protect\BIBand{} (FAO)(2020)}]{forest}
Food TUN, (FAO) AO (2020) Global forest resources assessment 2020. \urlprefix\url{https://www.fao.org/forest-resources-assessment/2020/en/}.

\bibitem[{French(2004)}]{French2004}
French E (2004) On the distribution and dynamics of health care costs: Extended version. Technical Report 2002-21, Federal Reserve Bank of Chicago.

\bibitem[{Gao et~al.(2022)Gao, Huang, Huang, Xiao, Wu, Sun, \protect\BIBand{} Zhang}]{gao2022combination}
Gao G, Huang S, Huang H, Xiao M, Wu J, Sun YE, Zhang S (2022) Combination of auction theory and multi-armed bandits: Model, algorithm, and application. \emph{IEEE Transactions on Mobile Computing} .

\bibitem[{Gayle \protect\BIBand{} Miller(2015)}]{gayle2015identifying}
Gayle GL, Miller RA (2015) Identifying and testing models of managerial compensation. \emph{The Review of Economic Studies} 82(3):1074--1118.

\bibitem[{Ghamat et~al.(2018)Ghamat, Zaric, \protect\BIBand{} Pun}]{ghamat2018contracts}
Ghamat S, Zaric GS, Pun H (2018) Contracts to promote optimal use of optional diagnostic tests in cancer treatment. \emph{Production and Operations Management} 27(12):2184--2200.

\bibitem[{Ghosh \protect\BIBand{} Sankararaman(2022)}]{ghosh2022breaking}
Ghosh A, Sankararaman A (2022) Breaking the $\sqrt{T}$ barrier: Instance-independent logarithmic regret in stochastic contextual linear bandits. \emph{International Conference on Machine Learning}, 7531--7549 (PMLR).

\bibitem[{Gibson et~al.(2006)Gibson, Mark, McGuigan, Axelsen, \protect\BIBand{} Wang}]{Gibson2006}
Gibson TB, Mark TL, McGuigan KA, Axelsen K, Wang S (2006) The effects of prescription drug cost sharing: A review of the evidence. \emph{The American Journal of Managed Care} 12(9):509--517.

\bibitem[{Gneezy et~al.(2011)Gneezy, Meier, \protect\BIBand{} Rey-Biel}]{gneezy2011and}
Gneezy U, Meier S, Rey-Biel P (2011) When and why incentives (don't) work to modify behavior. \emph{Journal of economic perspectives} 25(4):191--210.

\bibitem[{Gottlieb \protect\BIBand{} Moreira(2022)}]{gottlieb2022simple}
Gottlieb D, Moreira H (2022) Simple contracts with adverse selection and moral hazard. \emph{Theoretical Economics} 17(3):1357--1401.

\bibitem[{Grossman \protect\BIBand{} Hart(1983)}]{grossman1983}
Grossman S, Hart O (1983) An analysis of the principal-agent problem. \emph{Econometrica} 51(1):7--45.

\bibitem[{Guo et~al.(2019)Guo, Tang, Wang, \protect\BIBand{} Zhao}]{guo2019impact}
Guo P, Tang CS, Wang Y, Zhao M (2019) The impact of reimbursement policy on social welfare, revisit rate, and waiting time in a public healthcare system: Fee-for-service versus bundled payment. \emph{Manufacturing \& Service Operations Management} 21(1):154--170.

\bibitem[{Halac et~al.(2016)Halac, Kartik, \protect\BIBand{} Liu}]{halac2016optimal}
Halac M, Kartik N, Liu Q (2016) Optimal contracts for experimentation. \emph{The Review of Economic Studies} 83(3):1040--1091.

\bibitem[{Han et~al.(2020)Han, Zhou, Flores, Ordentlich, \protect\BIBand{} Weissman}]{han2020learning}
Han Y, Zhou Z, Flores A, Ordentlich E, Weissman T (2020) Learning to bid optimally and efficiently in adversarial first-price auctions. \emph{arXiv preprint arXiv:2007.04568} .

\bibitem[{Hart \protect\BIBand{} Holmstr{\"o}m(1987)}]{hart1987theory}
Hart O, Holmstr{\"o}m B (1987) The theory of contracts. \emph{Advances in economic theory: Fifth world congress}, volume~71, 155 (Cambridge).

\bibitem[{Hespanhol \protect\BIBand{} Aswani(2020)}]{hespanhol2020statistical}
Hespanhol P, Aswani A (2020) Statistical consistency of set-membership estimator for linear systems. \emph{IEEE Control Systems Letters} 4(3):668--673.

\bibitem[{Heuberger(2004)}]{heuberger2004inverse}
Heuberger C (2004) Inverse combinatorial optimization: A survey on problems, methods, and results. \emph{Journal of combinatorial optimization} 8(3):329--361.

\bibitem[{Ho et~al.(2016)Ho, Slivkins, \protect\BIBand{} Vaughan}]{ho2016}
Ho CJ, Slivkins A, Vaughan J (2016) Adaptive contract design for crowdsourcing markets: Bandit algorithms for repeated principal-agent problems. \emph{Journal of Artificial Intelligence Research} 55:317--359.

\bibitem[{Holmstr{\"o}m(1979)}]{holmstrom1979moral}
Holmstr{\"o}m B (1979) Moral hazard and observability. \emph{The Bell journal of economics} 74--91.

\bibitem[{Hong et~al.(2021)Hong, Fan, \protect\BIBand{} Luo}]{hong2021review}
Hong LJ, Fan W, Luo J (2021) Review on ranking and selection: A new perspective. \emph{Frontiers of Engineering Management} 8(3):321--343.

\bibitem[{Huang et~al.(2024)Huang, Jia, \protect\BIBand{} Zhou}]{huang2024sublinear}
Huang Y, Jia Y, Zhou XY (2024) Sublinear regret for a class of continuous-time linear-quadratic reinforcement learning problems. \emph{arXiv preprint arXiv:2407.17226} .

\bibitem[{Jain et~al.(2014)Jain, Narayanaswamy, \protect\BIBand{} Narahari}]{jain2014multiarmed}
Jain S, Narayanaswamy B, Narahari Y (2014) A multiarmed bandit incentive mechanism for crowdsourcing demand response in smart grids. \emph{Proceedings of the AAAI Conference on Artificial Intelligence}, volume~28.

\bibitem[{Juan et~al.(2014)Juan, Faulin, Pérez-Bernabeu, \protect\BIBand{} Jozefowiez}]{juan2014}
Juan AA, Faulin J, Pérez-Bernabeu E, Jozefowiez N (2014) Horizontal cooperation in vehicle routing problems with backhauling and environmental criteria. \emph{Procedia - Social and Behavioral Sciences} 111:1133--1141.

\bibitem[{Kaynar \protect\BIBand{} Siddiq(2022)}]{kaynar2022estimating}
Kaynar N, Siddiq A (2022) Estimating effects of incentive contracts in online labor platforms. \emph{Management Science} .

\bibitem[{Keshavarz et~al.(2011)Keshavarz, Wang, \protect\BIBand{} Boyd}]{keshavarz2011imputing}
Keshavarz A, Wang Y, Boyd S (2011) Imputing a convex objective function. \emph{2011 IEEE international symposium on intelligent control}, 613--619 (IEEE).

\bibitem[{Kurz(2017)}]{Kurz2017}
Kurz CF (2017) Tweedie distributions for fitting semicontinuous health care utilization cost data. \emph{BMC Medical Research Methodology} 17(1):171.

\bibitem[{Lagarde et~al.(2007)Lagarde, Haines, \protect\BIBand{} Palmer}]{lagarde2007conditional}
Lagarde M, Haines A, Palmer N (2007) Conditional cash transfers for improving uptake of health interventions in low-and middle-income countries: a systematic review. \emph{Jama} 298(16):1900--1910.

\bibitem[{Lambert et~al.(2008)Lambert, Lipkovich, \protect\BIBand{} Wei}]{Lambert2008}
Lambert D, Lipkovich I, Wei L (2008) Analysis of semicontinuous healthcare cost data. \emph{Health Economics} 17(8):909--929.

\bibitem[{Lattimore \protect\BIBand{} Szepesv{\'a}ri(2020)}]{lattimore2020bandit}
Lattimore T, Szepesv{\'a}ri C (2020) \emph{Bandit algorithms} (Cambridge University Press).

\bibitem[{Lee \protect\BIBand{} Zenios(2012)}]{leezenios2012}
Lee DKK, Zenios SA (2012) An evidence-based incentive system for medicare's end-stage renal disease program. \emph{Management Science} 58(6):1092--1105.

\bibitem[{Li et~al.(2023)Li, Ashlagi, \protect\BIBand{} Lo}]{li2023simple}
Li WD, Ashlagi I, Lo I (2023) Simple and approximately optimal contracts for payment for ecosystem services. \emph{Management Science} 69(12):7821--7837.

\bibitem[{Lobo et~al.(2017)Lobo, Denton, Wilson, Shah, \protect\BIBand{} Smith}]{lobo2017using}
Lobo J, Denton B, Wilson J, Shah ND, Smith S (2017) Using claims data linked with electronic health records to monitor and improve adherence to medication. \emph{IISE Transactions on Healthcare Systems Engineering} 7(4):194--214.

\bibitem[{Long et~al.(2011)Long, Smith, Zhang, Tang, \protect\BIBand{} Garner}]{long2011patient}
Long Q, Smith H, Zhang T, Tang S, Garner P (2011) Patient medical costs for tuberculosis treatment and impact on adherence in china: a systematic review. \emph{BMC public health} 11(1):1--9.

\bibitem[{Maheshwari et~al.(2022)Maheshwari, Kulkarni, Wu, \protect\BIBand{} Sastry}]{maheshwari2022inducing}
Maheshwari C, Kulkarni K, Wu M, Sastry SS (2022) Inducing social optimality in games via adaptive incentive design. \emph{2022 IEEE 61st Conference on Decision and Control (CDC)}, 2864--2869 (IEEE).

\bibitem[{Maheshwari et~al.(2023)Maheshwari, Sasty, Ratliff, \protect\BIBand{} Mazumdar}]{maheshwari2023convergent}
Maheshwari C, Sasty SS, Ratliff L, Mazumdar E (2023) Convergent first-order methods for bi-level optimization and stackelberg games.

\bibitem[{Marques et~al.(2020)Marques, Soares, Santos, \protect\BIBand{} Amorim}]{marques2020integrated}
Marques A, Soares R, Santos MJ, Amorim P (2020) Integrated planning of inbound and outbound logistics with a rich vehicle routing problem with backhauls. \emph{Omega} 92:102172.

\bibitem[{Martimort \protect\BIBand{} Laffont(2009)}]{martimort2009theory}
Martimort D, Laffont JJ (2009) \emph{The Theory of Incentives: The Principal-Agent Model} (Princeton University Press).

\bibitem[{Mason \protect\BIBand{} Plantinga(2013)}]{mason2013additionality}
Mason CF, Plantinga AJ (2013) The additionality problem with offsets: Optimal contracts for carbon sequestration in forests. \emph{Journal of Environmental Economics and Management} 66(1):1--14.

\bibitem[{Mason et~al.(2012)Mason, England, Denton, Smith, Kurt, \protect\BIBand{} Shah}]{mason2012optimizing}
Mason JE, England DA, Denton BT, Smith SA, Kurt M, Shah ND (2012) Optimizing statin treatment decisions for diabetes patients in the presence of uncertain future adherence. \emph{Medical Decision Making} 32(1):154--166.

\bibitem[{Mihaylova et~al.(2011)Mihaylova, Briggs, O'Hagan, \protect\BIBand{} Thompson}]{Mihaylova2011}
Mihaylova B, Briggs A, O'Hagan A, Thompson SG (2011) Review of statistical methods for analysing healthcare resources and costs. \emph{BMC Health Services Research} 11(1):183.

\bibitem[{Mintz et~al.(2023)Mintz, Aswani, Kaminsky, Flowers, \protect\BIBand{} Fukuoka}]{mintz2023behavioral}
Mintz Y, Aswani A, Kaminsky P, Flowers E, Fukuoka Y (2023) Behavioral analytics for myopic agents. \emph{European Journal of Operational Research} .

\bibitem[{Misra et~al.(2005)Misra, Coughlan, \protect\BIBand{} Narasimhan}]{misra2005salesforce}
Misra S, Coughlan AT, Narasimhan C (2005) Salesforce compensation: An analytical and empirical examination of the agency theoretic approach. \emph{Quantitative Marketing and Economics} 3(1):5--39.

\bibitem[{Misra \protect\BIBand{} Nair(2011)}]{misra2011structural}
Misra S, Nair HS (2011) A structural model of sales-force compensation dynamics: Estimation and field implementation. \emph{Quantitative Marketing and Economics} 9(3):211--257.

\bibitem[{Navabi \protect\BIBand{} Nayyar(2018)}]{navabi2018optimal}
Navabi S, Nayyar A (2018) Optimal auction design for flexible consumers. \emph{IEEE Transactions on Control of Network Systems} 6(1):138--150.

\bibitem[{Nazerzadeh et~al.(2008)Nazerzadeh, Saberi, \protect\BIBand{} Vohra}]{Nazerzadeh2008DynamicCM}
Nazerzadeh H, Saberi A, Vohra RV (2008) Dynamic cost-per-action mechanisms and applications to online advertising. \emph{WWW}.

\bibitem[{Osterberg \protect\BIBand{} Blaschke(2005)}]{osterberg2005adherence}
Osterberg L, Blaschke T (2005) Adherence to medication. \emph{New England journal of medicine} 353(5):487--497.

\bibitem[{Pittman et~al.(2011)Pittman, Chen, Bowlin, \protect\BIBand{} Foody}]{pittman2011adherence}
Pittman DG, Chen W, Bowlin SJ, Foody JM (2011) Adherence to statins, subsequent healthcare costs, and cardiovascular hospitalizations. \emph{The American journal of cardiology} 107(11):1662--1666.

\bibitem[{Plambeck \protect\BIBand{} Zenios(2000)}]{plambeck2000performance}
Plambeck EL, Zenios SA (2000) Performance-based incentives in a dynamic principal-agent model. \emph{Manufacturing \& service operations management} 2(3):240--263.

\bibitem[{Radner(1981)}]{radner1981monitoring}
Radner R (1981) Monitoring cooperative agreements in a repeated principal-agent relationship. \emph{Econometrica: Journal of the Econometric Society} 1127--1148.

\bibitem[{Rogerson(1985)}]{rogerson1985repeated}
Rogerson WP (1985) Repeated moral hazard. \emph{Econometrica: Journal of the Econometric Society} 69--76.

\bibitem[{Salzman et~al.(2018)Salzman, Bennett, Carroll, Goldstein, \protect\BIBand{} Jenkins}]{salzman2018global}
Salzman J, Bennett G, Carroll N, Goldstein A, Jenkins M (2018) The global status and trends of payments for ecosystem services. \emph{Nature Sustainability} 1(3):136--144.

\bibitem[{Sannikov(2008)}]{sannikov2008}
Sannikov Y (2008) A continuous- time version of the principal: Agent problem. \emph{The Review of Economic Studies} 75(3):957--984.

\bibitem[{Sannikov(2013)}]{sannikov2013}
Sannikov Y (2013) Contracts: The theory of dynamic principal—agent relationships and the continuous-time approach. \emph{Advances in Economics and Econometrics: Volume 1, Economic Theory: Tenth World Congress}, volume~49, 89.

\bibitem[{Santos et~al.(2021)Santos, Curcio, Amorim, Carvalho, \protect\BIBand{} Marques}]{santos2021bilevel}
Santos MJ, Curcio E, Amorim P, Carvalho M, Marques A (2021) A bilevel approach for the collaborative transportation planning problem. \emph{International Journal of Production Economics} 233:108004.

\bibitem[{Schweppe(1967)}]{schweppe1967}
Schweppe FC (1967) Recursive state estimation: Unknown but bounded errors and system inputs. \emph{Sixth Symposium on Adaptive Processes}, 102--107.

\bibitem[{Seymour \protect\BIBand{} Harris(2019)}]{science}
Seymour F, Harris NL (2019) Reducing tropical deforestation. \emph{Science} 365(6455):756--757.

\bibitem[{Shweta \protect\BIBand{} Sujit(2020)}]{shweta2020multiarmed}
Shweta J, Sujit G (2020) A multiarmed bandit based incentive mechanism for a subset selection of customers for demand response in smart grids. \emph{Proceedings of the AAAI Conference on Artificial Intelligence}, volume~34, 2046--2053.

\bibitem[{Simchowitz \protect\BIBand{} Slivkins(2021)}]{simchowitz2021}
Simchowitz M, Slivkins A (2021) Exploration and incentives in reinforcement learning. \emph{arXiv preprint arXiv:2103.00360} .

\bibitem[{Spear \protect\BIBand{} Srivastava(1987)}]{spear1987repeated}
Spear SE, Srivastava S (1987) On repeated moral hazard with discounting. \emph{The Review of Economic Studies} 54(4):599--617.

\bibitem[{Suen et~al.(2022)Suen, Negoescu, \protect\BIBand{} Goh}]{suen2022design}
Suen Sc, Negoescu D, Goh J (2022) Design of incentive programs for optimal medication adherence in the presence of observable consumption. \emph{Operations Research} .

\bibitem[{Turkensteen \protect\BIBand{} Hasle(2017)}]{turkensteen2017combining}
Turkensteen M, Hasle G (2017) Combining pickups and deliveries in vehicle routing--an assessment of carbon emission effects. \emph{Transportation Research Part C: Emerging Technologies} 80:117--132.

\bibitem[{Van~der Vaart(2000)}]{van2000asymptotic}
Van~der Vaart AW (2000) \emph{Asymptotic statistics}, volume~3 (Cambridge university press).

\bibitem[{Vera-Hernandez(2003)}]{vera2003structural}
Vera-Hernandez M (2003) Structural estimation of a principal-agent model: moral hazard in medical insurance. \emph{RAND Journal of Economics} 670--693.

\bibitem[{Wang et~al.(2022)Wang, Gao, \protect\BIBand{} Huang}]{wang2022}
Wang Z, Gao L, Huang J (2022) Socially-optimal mechanism design for incentivized online learning. \emph{IEEE INFOCOM 2022 - IEEE Conference on Computer Communications}, 1828--1837.

\bibitem[{Warnes et~al.(2023)Warnes, de~Zegher, Iancu, \protect\BIBand{} Plambeck}]{warnes2023area}
Warnes X, de~Zegher JF, Iancu D, Plambeck E (2023) Area conditions and positive incentives: Engaging local communities to protect forests. \emph{Available at SSRN 4609761} .

\bibitem[{WHO(2003)}]{world2003adherence}
WHO (2003) \emph{Adherence to long-term therapies: evidence for action} (World Health Organization).

\bibitem[{Williams(2015)}]{williams2015solvable}
Williams N (2015) A solvable continuous time dynamic principal--agent model. \emph{Journal of Economic Theory} 159:989--1015.

\end{thebibliography}



%
%
%
\begin{APPENDICES} 
\section{Proofs of All Results}
\subsection{Results in Section \ref{sec:estimator}} \label{appendix1}
\proof{\textbf{Proof of Proposition \ref{prop:iden1}.}}
We first note that $\ell\left(\mathbf{s}, i_t(\boldsymbol{\pi}_t), \boldsymbol{\pi}_t\right) = +\infty$ is obtained when the action selected by the agent (the maximizer of $\mathbf{s}^0 + \boldsymbol{\pi}_t$) is not the same as the maximizer of $\mathbf{s} + \boldsymbol{\pi}_t$. Since now we consider the case that $K^0 \cap K = \emptyset$, we already observe different indices for the largest entries of the true normalized rewards $\mathbf{s}^0$ and the considered normalized rewards $\mathbf{s}$ before adding the incentives. Hence, we can observe the desired event ($\ell\left(\mathbf{s}, i_t(\boldsymbol{\pi}_t), \boldsymbol{\pi}_t\right) = +\infty$) by simply choosing the incentive amounts in such a way that the new maximizers after adding the incentives will still belong to the sets $K$ and $K^0$. Suppose we have 
\begin{align}
    &\pi_{t,a} < R_{\min} + \gamma + \beta - d \ \text{for all }  a \in \mathcal{A} \setminus \{\kappa, \kappa^0\} \label{eq:smallc's-part1} \allowdisplaybreaks  \\
    &\pi_{t,a} \geq R_{\min} + \gamma + \beta - d \ \text{for } a \in \{\kappa, \kappa^0\} \label{eq:smallc's-part2} 
\end{align} 
Note that (\ref{eq:smallc's-part1}) and (\ref{eq:smallc's-part2}) are valid conditions according to Assumption \ref{assm1}. Now, recall that a vector $\mathbf{s} \in \mathcal{B}(\mathbf{s}^j, d)$ satisfies $\|\mathbf{s}^0 - \mathbf{s}\|_\infty > \beta$ by definition. We define $\widetilde{\mathbf{s}}^j := \arginf_{\mathbf{s} \in \mathcal{B}(\mathbf{s}^j, d)} \|\mathbf{s}^0 - \mathbf{s}\|_\infty $ as the closest vector (with respect to the $\ell_\infty$-norm) in ball $\mathcal{B}(\mathbf{s}^j, d)$ to the true reward vector $\mathbf{s}^0$. Then, we have $\|\mathbf{s}^0 - \widetilde{\mathbf{s}}^j\|_\infty \geq \beta - d$ by construction, and it follows that 
\begin{align}
    &\mathbb{P} \left(\ell\left(\mathbf{s}, i_t(\boldsymbol{\pi}_t), \boldsymbol{\pi}_t\right) = +\infty\right) \nonumber  \allowdisplaybreaks  \\
    &\geq \mathbb{P}\left( \bigcup_{x \in \mathcal{A}, y \in \mathcal{A}, y \neq x} x = \argmax_{a \in \mathcal{A}}\left(s^0_a  + \pi_{t,a} \right), y = \argmax_{a \in \mathcal{A}}\left(s_a  + \pi_{t,a} \right) \right)   \allowdisplaybreaks  \\
    &\geq \mathbb{P}\left(\kappa = \argmax_{a \in \mathcal{A}} (s_a + \pi_{t,a}),\kappa^0 = \argmax_{a \in \mathcal{A}} (s^0_a + \pi_{t,a})\right) \text{ for any } \kappa \in K, \ \kappa^0 \in K^0  \allowdisplaybreaks  \\
    & \geq \mathbb{P}\left(\kappa = \argmax_{a \in \mathcal{A}} (s_a + \pi_{t,a}),\kappa^0 = \argmax_{a \in \mathcal{A}} (s^0_a + \pi_{t,a}) \bigg | (\ref{eq:smallc's-part1}), (\ref{eq:smallc's-part2})\right) \mathbb{P}\left((\ref{eq:smallc's-part1}), (\ref{eq:smallc's-part2})\right)  \allowdisplaybreaks  \\
    & = \mathbb{P}\left(s_{\kappa^0} - s_{\kappa} < \pi_{t,\kappa} - \pi_{t,\kappa^0} < s^0_{\kappa^0} - s^0_{\kappa} \right) \hspace{-3mm}
    \prod_{a \in \{\kappa, \kappa^0 \}} \hspace{-3.5mm} \mathbb{P}\left(\pi_{t,a} \geq R_{\min} + \gamma + \beta - d \right) \hspace{-4mm} \prod_{a \in \mathcal{A} \setminus \{\kappa, \kappa^0 \}} \hspace{-4.5mm} \mathbb{P} \left(\pi_{t,a} < R_{\min} + \gamma + \beta - d \right)   \label{eq:byindepc's}\allowdisplaybreaks  \\
    &\geq \mathbb{P}\left(s_{\kappa^0} - s_{\kappa} < \pi_{t,\kappa} - \pi_{t,\kappa^0} < s^0_{\kappa^0} - s^0_{\kappa} \right) 
    \prod_{a \in \{\kappa, \kappa^0 \}} \hspace{-2mm} \mathbb{P}\left(\pi_{t,a} \geq R_{\min} + \gamma + \beta - d \right) \prod_{a \in \mathcal{A} \setminus \{\kappa, \kappa^0 \}} \hspace{-2mm} \mathbb{P} \left(\pi_{t,a} \leq R_{\min} + \gamma \right)   \allowdisplaybreaks  \\
    &= \mathbb{P}\left(s_{\kappa^0} - s_{\kappa} < \pi_{t,\kappa} - \pi_{t,\kappa^0} < s^0_{\kappa^0} - s^0_{\kappa} \right) \prod_{a \in \{\kappa, \kappa^0 \}} \left(1 - \frac{R_{\min} + \gamma + \beta - d - \underline{C}}{\overline{C} - \underline{C}} \right)  \prod_{a \in \mathcal{A} \setminus \{\kappa, \kappa^0 \}} \frac{R_{\min} + \gamma - \underline{C}}{\overline{C} - \underline{C}}   \label{eq:byuniformc's} \allowdisplaybreaks  \\
    &= \mathbb{P}\left(s_{\kappa^0} - s_{\kappa} < \pi_{t,\kappa} - \pi_{t,\kappa^0} < s^0_{\kappa^0} - s^0_{\kappa} \right)  \prod_{a \in \{\kappa, \kappa^0 \}} \left(1 - \frac{\gamma + \beta - d}{\overline{C} - \underline{C}} \right)  \prod_{a \in \mathcal{A} \setminus \{\kappa, \kappa^0 \}} \frac{\gamma}{\overline{C} - \underline{C}}   \label{eq:byrangeofc's} \allowdisplaybreaks  
\end{align}
where (\ref{eq:byindepc's}) follows since $\pi_{t,a}$'s are considered to be independent random variables, (\ref{eq:byuniformc's}) follows since $\pi_{t,a} \sim \mathcal{U}(\underline{C}, \overline{C}), \forall a \in \mathcal{A}$, and (\ref{eq:byrangeofc's}) follows since $\underline{C} = R_{\min}$ by Assumption \ref{assm1}. For the first term in (\ref{eq:byrangeofc's}), notice that the case that $s_{\kappa^0} - s_{\kappa} = s^0_{\kappa^0} - s^0_{\kappa} = 0$ cannot occur. This can only happen if $\kappa^0 \in K$ and $\kappa \in K^0$ which contradicts with the condition $K^0 \cap K_t = \emptyset$. Similarly,  $\mathbf{s}^0$ cannot be the all-zeros vector under the given condition $K^0 \cap K_t = \emptyset$. Thus, the following always holds under the given condition: $s^0_{\kappa^0} - s^0_{\kappa} > 0$,  $s_{\kappa^0} - s_{\kappa} < 0$, and $\mathbf{s}^0 \neq \mathbf{0}_n$. Then, we obtain 
\begin{align}
    (\ref{eq:byrangeofc's}) \geq \mathbb{P}\left(0 \leq \pi_{t,\kappa} - \pi_{t,\kappa^0} < s^0_{\kappa^0} - s^0_{\kappa} \right)  \left(1 - \frac{\gamma + \beta - d}{\overline{C} - \underline{C}} \right)^2  \left(\frac{\gamma}{\overline{C} - \underline{C}}\right)^{n-2}  \label{eq:combine1}
\end{align}
The probability term in the last inequality can be computed by using the cumulative distribution function (cdf) of $\pi_{t,a} - \pi_{t,a'}$ – which is the difference of two identically and independently distributed (iid) Uniform random variables. The difference $\pi_{t,a} - \pi_{t,a'}$ follows a triangular distribution whose cdf can be explicitly computed as follows. 
\begin{align}
    \mathbb{P}\left(\pi_{t,a} - \pi_{t,a'} \leq \Delta \right)  \label{eq:delta}
  &= \left\{\begin{array}{ll}
        0, & \text{for } \Delta < \underline{C}-\overline{C} \allowdisplaybreaks  \\
        \int\limits_{\underline{C}}^{\overline{C}+\Delta}  \int\limits_{\pi_{t,a} - \Delta}^{\overline{C}} \frac{1}{(\overline{C}-\underline{C})^2} d\pi_{t,a} d\pi_{t,a'}, & \text{for } \underline{C}-\overline{C} \leq \Delta < 0 \allowdisplaybreaks  \\
        1 - \int\limits_{\underline{C} + \Delta}^{\overline{C}}  \int\limits_{\underline{C}}^{\pi_{t,a} - \Delta} \frac{1}{(\overline{C}-\underline{C})^2} d\pi_{t,a} d\pi_{t,a'} , & \text{for } 0 \leq \Delta \leq \overline{C}-\underline{C} \allowdisplaybreaks  \\
        1 , & \text{for } \Delta \geq \overline{C}-\underline{C} \allowdisplaybreaks  \\
        \end{array}\right\} \allowdisplaybreaks  \\
    &= \left\{\begin{array}{ll}
        0, & \text{for } \Delta < \underline{C}-\overline{C} \allowdisplaybreaks  \\
        \frac{(\Delta + \overline{C} - \underline{C})^2 }{2(\overline{C}-\underline{C})^2}, & \text{for } \underline{C}-\overline{C} \leq \Delta < 0 \allowdisplaybreaks  \\
        1-\frac{\left(\Delta + \underline{C} - \overline{C}\right)^2 }{2(\overline{C}-\underline{C})^2} , & \text{for } 0 \leq \Delta \leq \overline{C}-\underline{C} \allowdisplaybreaks  \\
        1 , & \text{for } \Delta \geq \overline{C}-\underline{C} \allowdisplaybreaks  \\
        \end{array}\right\} \allowdisplaybreaks   \label{eq:cdf}
\end{align}
Since by construction we have $\mathcal{R} \subseteq [\underline{C}, \overline{C}]$, we know that $0 < s^0_{\kappa^0} - s^0_{\kappa} \leq \overline{C}-\underline{C}$ holds. Thus, we have
\begin{align}
   \mathbb{P}\left(0 \leq \pi_{t,\kappa} - \pi_{t,\kappa^0} < s^0_{\kappa^0} - s^0_{\kappa} \right) &= 1-\frac{\left(s^0_{\kappa^0} - s^0_{\kappa} + \underline{C} - \overline{C}\right)^2 }{2(\overline{C}-\underline{C})^2} -  \frac{1}{2} = \frac{1}{2} -\frac{\left(\overline{C} - \underline{C} - s^0_{\kappa^0} +  s^0_{\kappa}\right)^2 }{2(\overline{C}-\underline{C})^2} > 0 \allowdisplaybreaks 
\end{align}
Combining this last result with (\ref{eq:combine1}), we obtain the desired result and conclude.  
\Halmos \endproof

\proof{\textbf{Proof of Proposition \ref{prop:iden2}.}} 
Recall that the event $\ell\left(\mathbf{s}, i_t(\boldsymbol{\pi}_t), \boldsymbol{\pi}_t\right) = +\infty$ is observed when the maximizer entries of the total reward vectors $\mathbf{s}^0 + \boldsymbol{\pi}_t$ and $\mathbf{s} + \boldsymbol{\pi}_t$ are different from each other. Hence, to prove the lower bound in (\ref{eq:identf-2}), we will consider the case when $\argmax_{a \in \mathcal{A}}\left(s_a  + \pi_{t,a} \right) = 1$ and $\argmax_{a \in \mathcal{A}}\left(s^0_a  + \pi_{t,a} \right) = b$ because we know that $b \neq 1$. As we have $s_1 = s^0_1 = 0$ by construction, having $b = 1$ would imply that $\mathbf{s}^0 = \mathbf{s} = \mathbf{0}_n$. However, this contradicts with the fact that $\mathbf{s} \in \mathcal{B}(\mathbf{s}^j, d) \in \mathcal{F}$ which means $\|\mathbf{s}^0 - \mathbf{s}\|_\infty = |s^0_b - s_b| > \beta$ must be satisfied. 

With this consideration, let $\omega = \sup_{\mathbf{s} \in \mathcal{B}(\mathbf{s}^j, d)} \max_{a \in \mathcal{A}} \{|s^0_a|, |s_a| \}$ be the largest absolute value observed among the entries of $\mathbf{s}^0$ and of all vectors in $\mathcal{B}(\mathbf{s}^j, d)$. 
Then, suppose we have 
\begin{align}
    & \pi_{t,a} < R_{\min} + \gamma + \beta - d \ \text{for all }  a \in \mathcal{A} \setminus \{1, b\} \label{eq:smallc's-2-part1} \allowdisplaybreaks  \\
    & \pi_{t,a} \geq R_{\min} + \gamma + \omega  \ \text{for } a \in \{1, b\}. \label{eq:smallc's-2-part2} 
\end{align} 
Note that (\ref{eq:smallc's-2-part1}) and (\ref{eq:smallc's-2-part2}) are consistent with Assumption \ref{assm1}. Further, they imply that the indices in the sets $K^0$ and $K$ are no more maximizers after adding the incentives in (\ref{eq:smallc's-2-part1})-(\ref{eq:smallc's-2-part2}). To restate, we now have $s^0_{\kappa^0} + \pi_{t, \kappa^0} < s^0_a + \pi_{t, a}$ and $s_{\kappa} + \pi_{t, \kappa} < s_{a} + \pi_{t, a}$ for any $\kappa^0 \in K^0$, $\kappa \in K$, $a \in \{1, b\}$. Further, if the events $s^0_{1} + \pi_{t, 1} < s^0_{b} + \pi_{t, b}$ and $s_b + \pi_{t, b} < s_1 + \pi_{t, 1}$ also hold, then we will obtain the desired case (that is $\argmax_{a \in \mathcal{A}} (s_a + \pi_{t,a}) = 1$ and $\argmax_{a \in \mathcal{A}} (s^0_a + \pi_{t,a}) = b$). Our proof will be based on this observation.

Since $|s^0_b - s_b| > \beta$ by definition, we know that $|s^0_b - s_b| > |s^0_1 - s_1| = 0$. Suppose that without loss of generality, we have $s^0_b - s_b > s^0_1 - s_1 = 0$ and $s^0_b - s_b > \beta$. Then, we get
\begin{align}
    &\mathbb{P} \left(\ell\left(\mathbf{s}, i_t(\boldsymbol{\pi}_t), \boldsymbol{\pi}_t\right) = +\infty\right) \nonumber \allowdisplaybreaks  \\
    &\geq \mathbb{P}\left( \bigcup_{x \in \mathcal{A}, y \in \mathcal{A}, y \neq x} x = \argmax_{a \in \mathcal{A}}\left(s^0_a  + \pi_{t,a} \right), y = \argmax_{a \in \mathcal{A}}\left(s_a  + \pi_{t,a} \right) \right) \allowdisplaybreaks  \\
    &\geq \mathbb{P}\left(1 = \argmax_{a \in \mathcal{A}} (s_a + \pi_{t,a}), b = \argmax_{a \in \mathcal{A}} (s^0_a + \pi_{t,a})\right) \allowdisplaybreaks  \\
    &\geq \mathbb{P}\left(1 = \argmax_{a \in \mathcal{A}} (s_a + \pi_{t,a}), b = \argmax_{a \in \mathcal{A}} (s^0_a + \pi_{t,a}) \bigg |  (\ref{eq:smallc's-2-part1}), (\ref{eq:smallc's-2-part2}) \right) \mathbb{P}\left((\ref{eq:smallc's-2-part1}), (\ref{eq:smallc's-2-part2}) \right) \allowdisplaybreaks  \\ 
    &= \mathbb{P}\left(s_b - s_1 <  \pi_{t, 1} - \pi_{t, b} < s^0_{b} - s^0_{1}\right) \prod_{a \in \{1, b\}} \hspace{-2mm}  \mathbb{P} \left(\pi_{t,a} \geq R_{\min} + \gamma + \omega \right) 
    \prod_{a \in \mathcal{A} \setminus \{1, b\}} \mathbb{P} \left(\pi_{t,a} < R_{\min} + \gamma + \beta - d \right) \allowdisplaybreaks  \\ 
    &\geq \mathbb{P}\left(s_b - s_1 < \pi_{t, 1} - \pi_{t, b} < s^0_{b} - s^0_{1}\right)  \prod_{a \in \{1, b\}} \hspace{-2mm}  \mathbb{P} \left(\pi_{t,a} \geq R_{\min} + \gamma + \omega \right) 
    \prod_{a \in \mathcal{A} \setminus \{1, b\}} \mathbb{P} \left(\pi_{t,a} \leq R_{\min} + \gamma \right) \label{eq:byindepc's-2-1} \allowdisplaybreaks 
     \allowdisplaybreaks  \\ 
    &= \mathbb{P}\left(s_b - s_1 < \pi_{t, 1} - \pi_{t, b} < s^0_{b} - s^0_{1}\right) \prod_{a \in \{1, b\}} \left(1 - \frac{\gamma + \omega}{\overline{C} - \underline{C}} \right) \prod_{a \in \mathcal{A} \setminus \{1, b\}}  \frac{\gamma}{\overline{C} - \underline{C}}  \label{eq:byindepc's-2-2}  \allowdisplaybreaks 
\end{align}
where (\ref{eq:byindepc's-2-1}) and (\ref{eq:byindepc's-2-2}) follow since $\underline{C} = R_{\min}$ by Assumption \ref{assm1} and $\pi_{t,a}$'s are independent random variables with $\pi_{t,a} \sim \mathcal{U}(\underline{C}, \overline{C}), \forall a \in \mathcal{A}$.

We next compute a lower bound for the first term in (\ref{eq:byindepc's-2-2}).
\begin{align}
    \mathbb{P}\left(s_b - s_1 < \pi_{t, 1} - \pi_{t, b} < s^0_{b} - s^0_{1}\right) &= \mathbb{P}\left(s_b - s_1 + s^0_b - s^0_b  < \pi_{t, 1} - \pi_{t, b} < s^0_{b} - s^0_{1}\right)  \label{eq:combine2-0} \allowdisplaybreaks  \\
     &\geq \mathbb{P}\left(s^0_{b} - \beta - s_1 < \pi_{t, 1} - \pi_{t, b} < s^0_{b} - s^0_{1} \right) \allowdisplaybreaks  \\
     &= \mathbb{P}\left(s^0_{b} - \beta < \pi_{t, 1} - \pi_{t, b} < s^0_{b} \right)   \allowdisplaybreaks  
\end{align}
We can compute the probability in the last line above by using the cdf derived in (\ref{eq:cdf}). Since the cdf is a piecewise function, we need to consider the two disjoint cases given as:
\begin{itemize}
    \item[--] \textit{Case 1:} $\underline{C}-\overline{C} \leq s^0_{b} < 0$
    \item[--] \textit{Case 2:} $0 \leq s^0_{b} \leq \overline{C}-\underline{C}$ 
\end{itemize}
We also consider the following subcases to derive the probability bounds for the two cases above.
\begin{itemize}
    \item[--] \textit{Subcase 1:} $\underline{C}-\overline{C} \leq s^0_{b} < 0$ and $\underline{C}-\overline{C} \leq s^0_{b} - \beta < 0$
    \item[--] \textit{Subcase 2:} $0 \leq s^0_{b} \leq \overline{C}-\underline{C}$ and $0 \leq s^0_{b} - \beta \leq \overline{C}-\underline{C}$
\end{itemize}
We can bound $\mathbb{P}\left(s^0_{b} - \beta < \pi_{t, 1} - \pi_{t, b} < s^0_{b} \right)$ from below under \textit{Subcase 1} and \textit{Subcase 2} as follows.
\begin{align}
    \mathbb{P} \left(s^0_{b} - \beta \leq \pi_{t, 1} - \pi_{t, b} < s^0_{b}, \textit{Subcase 1} \right) &=  \frac{(s^0_{b} + \overline{C} - \underline{C})^2 }{2(\overline{C}-\underline{C})^2} - \frac{(s^0_{b} - \beta + \overline{C} - \underline{C})^2 }{2(\overline{C}-\underline{C})^2} \allowdisplaybreaks  \\
    &= \frac{(s^0_{b})^2 - (s^0_{b} - \beta)^2 + 2(s^0_{b} - (s^0_{b} - \beta))(\overline{C} - \underline{C})}{2(\overline{C}-\underline{C})^2}\allowdisplaybreaks  \\
    &= \frac{(s^0_{b})^2 - (s^0_{b} - \beta)^2 + 2\beta(\overline{C} - \underline{C})}{2(\overline{C}-\underline{C})^2} \allowdisplaybreaks  \\
    &= \frac{-\beta^2 + 2\beta(s^0_b  + \overline{C} - \underline{C})}{2(\overline{C}-\underline{C})^2} \allowdisplaybreaks  \\
    &\geq \frac{-\beta^2 + 2\beta^2}{2(\overline{C}-\underline{C})^2} \allowdisplaybreaks  \\
    &= \frac{\beta^2}{2(\overline{C}-\underline{C})^2} \allowdisplaybreaks 
\end{align}
where second to the last line follows since we have $0 < \beta \leq s^0_b + \overline{C} - \underline{C}$ in this subcase. 
\begin{align}
    \mathbb{P} \left(s^0_{b} - \beta \leq \pi_{t, 1} - \pi_{t, b} < s^0_{b}, \textit{Subcase 2} \right) &=  1 - \frac{(s^0_{b} + \underline{C} - \overline{C})^2 }{2(\overline{C}-\underline{C})^2} - 1 + \frac{(s^0_{b} - \beta + \underline{C} - \overline{C})^2 }{2(\overline{C}-\underline{C})^2} \allowdisplaybreaks  \\
    &=  \frac{(\overline{C} - \underline{C} - s^0_{b} + \beta)^2 }{2(\overline{C}-\underline{C})^2} - \frac{(\overline{C} - \underline{C} - s^0_{b})^2 }{2(\overline{C}-\underline{C})^2} \allowdisplaybreaks  \\
    &= \frac{\beta^2 + 2\beta(\overline{C} - \underline{C} - s^0_{b})}{2(\overline{C}-\underline{C})^2} \allowdisplaybreaks  \\
    &\geq \frac{\beta^2}{2(\overline{C}-\underline{C})^2} 
\end{align}
where the last inequality follows since we have $\overline{C} - \underline{C} - s^0_{b} \geq 0$ and $\beta > 0$ by definition. Now, since \textit{Case 1} and \textit{Case 2} are mutually exclusive events, we combine everything and obtain
\begin{align}
    &\mathbb{P} \left(s^0_{b} - \beta \leq \pi_{t, 1} - \pi_{t, b} < s^0_{b} \right) \nonumber \allowdisplaybreaks  \\
    &= \mathbb{P} \left(s^0_{b} - \beta \leq \pi_{t, 1} - \pi_{t, b} < s^0_{b}, \textit{Case 1} \right) + \mathbb{P} \left(s^0_{b} - \beta \leq \pi_{t, 1} - \pi_{t, b} < s^0_{b},\textit{Case 2}\right) \allowdisplaybreaks  \\
    &\geq  \mathbb{P} \left(s^0_{b} - \beta \leq \pi_{t, 1} - \pi_{t, b} < s^0_{b}, \textit{Subcase 1} \right) + \mathbb{P} \left(s^0_{b} - \beta \leq \pi_{t, 1} - \pi_{t, b} < s^0_{b},\textit{Subcase 2}\right)  \allowdisplaybreaks  \\
    &\geq \frac{\beta^2}{(\overline{C}-\underline{C})^2} \allowdisplaybreaks  \label{eq:combine2-last}
\end{align}
Combining this last result with (\ref{eq:byindepc's-2-2}), we obtain 
\begin{align}
    \mathbb{P} \left(\ell\left(\mathbf{s}, i_t(\boldsymbol{\pi}_t), \boldsymbol{\pi}_t\right) = +\infty\right) &\geq \frac{\beta^2}{(\overline{C}-\underline{C})^2} \left(1 - \frac{\gamma + \omega}{\overline{C} - \underline{C}} \right)^2 \left( \frac{\gamma}{\overline{C} - \underline{C}}\right)^{n-2} \allowdisplaybreaks   
\end{align}
\Halmos \endproof

\proof{\textbf{Proof of Proposition \ref{prop:iden3}.}}
We follow a mainly similar argument as in the proof of Proposition \ref{prop:iden2}. Recall that we know $b \neq 1$ since $\|\mathbf{s}^0 - \mathbf{s}\|_\infty > \beta$ by construction as explained in the previous proof, and that either $s_b >0$ or $s^0_b >0$ holds. We also have 
$\omega = \sup_{\mathbf{s} \in \mathcal{B}(\mathbf{s}^j, d)} \max_{a \in \mathcal{A}} \{|s^0_a|, |s_a| \}$ as before. Now, consider the following conditions on the incentives 
\begin{align}
    &\pi_{t,a} < R_{\min} + \gamma + \beta - d \ \text{for all }  a \in \mathcal{A} \setminus \{1, b\} \label{eq:smallc's-3-part1} \allowdisplaybreaks  \\
    &\pi_{t,b} \geq R_{\min} + \gamma + \beta - d \label{eq:smallc's-3-part2}  \allowdisplaybreaks  \\
    &\pi_{t,1} \geq R_{\min} + \gamma + \omega \label{eq:smallc's-3-part3} 
\end{align} 
which are compatible with Assumption \ref{assm1}. Now, since $|s^0_b - s_b| > \beta$ by definition of $\mathbf{s}$, we know that $|s^0_b - s_b| > |s^0_1 - s_1| = 0$. Suppose that without loss of generality, we have $s^0_b - s_b > s^0_1 - s_1 = 0$ and $s^0_b - s_b > \beta$. Then, we obtain
\begin{align}
    & \mathbb{P} \left(\ell\left(\mathbf{s}, i_t(\boldsymbol{\pi}_t), \boldsymbol{\pi}_t\right) = +\infty\right) \nonumber \allowdisplaybreaks  \\
    &\geq \mathbb{P}\left( \bigcup_{x \in \mathcal{A}, y \in \mathcal{A}, y \neq x} x = \argmax_{a \in \mathcal{A}}\left(s^0_a  + \pi_{t,a} \right), y = \argmax_{a \in \mathcal{A}}\left(s_a  + \pi_{t,a} \right) \right) \allowdisplaybreaks  \\
    &\geq \mathbb{P}\left(1 =\argmax_{a \in \mathcal{A}} (s_a + \pi_{t,a}), b = \argmax_{a \in \mathcal{A}} (s^0_a + \pi_{t,a})\right) \allowdisplaybreaks  \\
    &\geq \mathbb{P}\left(1 = \argmax_{a \in \mathcal{A}} (s_a + \pi_{t,a}), b = \argmax_{a \in \mathcal{A}} (s^0_a + \pi_{t,a}) \bigg |  (\ref{eq:smallc's-3-part1}) - (\ref{eq:smallc's-3-part3}) \right) \mathbb{P}\left((\ref{eq:smallc's-3-part1}) - (\ref{eq:smallc's-3-part3}) \right) \allowdisplaybreaks  \\ 
    &=  \mathbb{P}\left(s_b - s_1 <  \pi_{t, 1} - \pi_{t, b} < s^0_{b} - s^0_{1} \right) \mathbb{P}\left(\ref{eq:smallc's-3-part2}\right)\mathbb{P}\left(\ref{eq:smallc's-3-part3}\right) \prod_{a \in \mathcal{A} \setminus \{1, b\}}  \mathbb{P}\left(\pi_{t,a} < R_{\min} + \gamma + \beta - d \right)  \label{eq:byindepc's-3} \allowdisplaybreaks  \\ 
    &\geq  \mathbb{P}\left(s_b - s_1 < \pi_{t, 1} - \pi_{t, b} < s^0_{b} - s^0_{1}\right) \mathbb{P}\left(\ref{eq:smallc's-3-part2}\right)\mathbb{P}\left(\ref{eq:smallc's-3-part3}\right) \prod_{a \in \mathcal{A} \setminus \{1, b\}} \mathbb{P}\left(\pi_{t,a} \leq R_{\min} + \gamma\right) \allowdisplaybreaks  \\ 
    &=  \mathbb{P}\left(s_b - s_1 < \pi_{t, 1} - \pi_{t, b} < s^0_{b} - s^0_{1}\right) \left(1 - \frac{\gamma + \beta - d}{\overline{C} - \underline{C}} \right) \left(1 - \frac{\gamma + \omega}{\overline{C} - \underline{C}} \right) \prod_{a \in \mathcal{A} \setminus \{1, b\}} \frac{\gamma}{\overline{C} - \underline{C}}   \label{eq:combine3} \allowdisplaybreaks 
\end{align}
where (\ref{eq:byindepc's-3}) follows as $\pi_{t,a}$'s are independent random variables and (\ref{eq:combine3}) follows since $\underline{C} = R_{\min}$ by Assumption \ref{assm1} and $\pi_{t,a} \sim \mathcal{U}(\underline{C}, \overline{C}), \forall a \in \mathcal{A}$. Then, we obtain the following lower bound for the first term in (\ref{eq:combine3})
\begin{align}
     \mathbb{P}\left(s_b - s_1 < \pi_{t, 1} - \pi_{t, b} < s^0_{b} - s^0_{1}\right) \geq \mathbb{P}\left(s^0_{b} - \beta < c_{t, 1} - c_{t, b} < s^0_{b} \right)  \geq \frac{\beta^2}{(\overline{C}-\underline{C})^2}  \allowdisplaybreaks 
\end{align}
by using similar arguments as in (\ref{eq:combine2-0})-(\ref{eq:combine2-last}) from the proof of Proposition \ref{prop:iden2}. Lastly, combining this result with (\ref{eq:combine3}), we obtain 
\begin{align}
    \mathbb{P} \left(\ell\left(\mathbf{s}, i_t(\boldsymbol{\pi}_t), \boldsymbol{\pi}_t\right) = +\infty\right) &\geq \frac{\beta^2}{(\overline{C}-\underline{C})^2} \left(1 - \frac{\gamma + \beta - d}{\overline{C} - \underline{C}} \right) \left(1 - \frac{\gamma + \omega}{\overline{C} - \underline{C}} \right) \left( \frac{\gamma}{\overline{C} - \underline{C}}  \right)^{n-2} \allowdisplaybreaks  
\end{align}
\Halmos \endproof

\proof{\textbf{Proof of Proposition \ref{prop:iden4}.}}
Notice that the following three conditions are mutually exclusive events: 
\begin{enumerate} 
    \item[\textit{i.}] $K^0 \cap K = \emptyset$
    \item[\textit{ii.}] $K^0 \cap K \neq \emptyset$ and $b \notin K^0 \cap K$ 
    \item[\textit{iii.}] $K^0 \cap K \neq \emptyset$ and $b \in K^0 \cap K$ 
\end{enumerate}
Hence, we can unite the results of Propositions \ref{prop:iden1}, \ref{prop:iden2}, and \ref{prop:iden3} and obtain
\begin{align}
   \mathbb{P} \left(\ell\left(\mathbf{s}, i_t(\boldsymbol{\pi}_t), \boldsymbol{\pi}_t\right) = +\infty\right) &= \sum_{j \in \{i, ii, iii\}} \mathbb{P}\left(\ell\left(\mathbf{s}, i_t(\boldsymbol{\pi}_t), \boldsymbol{\pi}_t\right) = +\infty, \mathrm{\textit{j}}  \right) \allowdisplaybreaks  \\
   &\geq \alpha \beta^2 \allowdisplaybreaks 
\end{align}
for some constant $\alpha > 0$.
\Halmos \endproof

\proof{\textbf{Proof of Theorem \ref{thm:concen}.}} Recall that we define an open ball $\mathcal{B}(\mathbf{s}^j, d) := \{\mathbf{s} : \| \mathbf{s} - \mathbf{s}^j \|_\infty < d\}$ centered around a vector $\mathbf{s}^j$ with diameter $d >0$. Since $\mathcal{F} = \{\mathbf{s} \in \mathcal{S}^n: \| \mathbf{s} - \mathbf{s}^0 \|_\infty > \beta\}$ is compact, there is a finite subcover $\{\mathcal{B}(\mathbf{s}^j, d) : \mathbf{s}^j \in \mathcal{F}\}_{j = 1}^q$ of a collection of open balls covering $\mathcal{F}$ where $d < \beta$. Further, we define $ \overline{\mathbf{s}}^j_t := \arginf_{\mathbf{s} \in \mathcal{B}(\mathbf{s}^j, d)} L\left(\mathbf{s}, I_t(\boldsymbol{\Pi}_t), \boldsymbol{\Pi}_t\right)$. Now, since $\mathcal{F} \subseteq \bigcup_{j = 1}^q \mathcal{B}(\mathbf{s}^j, d)$, we have 
\begin{align}
    \inf_{\mathbf{s} \in \mathcal{F}} L\left(\mathbf{s}, I_t(\boldsymbol{\Pi}_t), \boldsymbol{\Pi}_t\right) = \inf_{\mathbf{s} \in \mathcal{F} }  \sum_{\tau = 1}^{t-1} \ell\left(\mathbf{s}, i_\tau(\boldsymbol{\pi}_\tau), \boldsymbol{\pi}_\tau\right)   
    &\geq \min_{j \in [q] } \inf_{\mathbf{s} \in \mathcal{B}(\mathbf{s}^j, d)} \sum_{\tau = 1}^{t-1} \ell\left(\mathbf{s}, i_\tau(\boldsymbol{\pi}_\tau), \boldsymbol{\pi}_\tau\right) \allowdisplaybreaks  \\
    &\geq \min_{j \in [q] } \sum_{\tau = 1}^{t-1} \ell\left(\overline{\mathbf{s}}^j_t, i_\tau(\boldsymbol{\pi}_\tau), \boldsymbol{\pi}_\tau\right) \allowdisplaybreaks  \\ 
    &\geq \min_{j \in [q] } \sum_{\tau \in \Lambda(1, t)} \ell\left(\overline{\mathbf{s}}^j_t, i_\tau(\boldsymbol{\pi}_\tau), \boldsymbol{\pi}_\tau\right) \label{eq:conc1} \allowdisplaybreaks 
\end{align}
where $[q] = \{1, \ldots, q\}$. We then follow by 
\begin{align}
    \mathbb{P}\left(\inf_{\mathbf{s} \in \mathcal{F}}  L\left(\mathbf{s}, I_t(\boldsymbol{\Pi}_t), \boldsymbol{\Pi}_t\right) < +\infty\right) &\leq \mathbb{P}\left(\min_{j \in [q] } \sum_{\tau \in \Lambda(1, t)} \ell\left(\overline{\mathbf{s}}^j_t, i_\tau(\boldsymbol{\pi}_\tau), \boldsymbol{\pi}_\tau\right) < +\infty \right) \allowdisplaybreaks  \\
    &\leq \mathbb{P}\left(\bigcup_{j \in [q]} \sum_{\tau \in \Lambda(1, t)} \ell\left(\overline{\mathbf{s}}^j_t, i_\tau(\boldsymbol{\pi}_\tau), \boldsymbol{\pi}_\tau\right) < +\infty \right) \allowdisplaybreaks  \\
    &\leq \sum_{j \in [q]} \mathbb{P}\left(\sum_{\tau \in \Lambda(1, t)} \ell\left(\overline{\mathbf{s}}^j_t, i_\tau(\boldsymbol{\pi}_\tau), \boldsymbol{\pi}_\tau\right) < +\infty \right) \label{eq:unionn} \allowdisplaybreaks  \\
    &= \sum_{j \in [q]} \mathbb{P}\left( \ell\left(\overline{\mathbf{s}}^j_t, i_\tau(\boldsymbol{\pi}_\tau), \boldsymbol{\pi}_\tau\right) < +\infty, \ \forall \tau \in \Lambda(1, t) \right)  \allowdisplaybreaks  \\
    &= \sum_{j \in [q]} \prod_{\tau \in \Lambda(1, t)}\mathbb{P}\left( \ell\left(\overline{\mathbf{s}}^j_t, i_\tau(\boldsymbol{\pi}_\tau), \boldsymbol{\pi}_\tau\right) < +\infty \right)   \label{eq:indept's}\allowdisplaybreaks  \\
    &= \sum_{j \in [q]} \prod_{\tau \in \Lambda(1, t)} \left[ 1 - \mathbb{P}\left( \ell\left(\overline{\mathbf{s}}^j_t, i_\tau(\boldsymbol{\pi}_\tau), \boldsymbol{\pi}_\tau\right)= +\infty \right) \right] \allowdisplaybreaks  \\
     &\leq \sum_{j \in [q]} \prod_{\tau \in \Lambda(1, t)} \left( 1 - \alpha \beta^2 \right)   \label{eq:byidentifiability} \allowdisplaybreaks 
     \end{align}
where the first inequality follows by (\ref{eq:conc1}), (\ref{eq:unionn}) follows by the Boole's inequality (a.k.a. union bound), (\ref{eq:indept's}) follows by the assumption of independence of the time steps, and  (\ref{eq:byidentifiability}) follows by the identifiability condition provided in Proposition \ref{prop:iden4}. Note that we prove Proposition \ref{prop:iden4} for any vector $\mathbf{s} \in \mathcal{B}(\mathbf{s}^j, d)$, and hence, it also holds for $\overline{\mathbf{s}}^j_t$. We continue as
\begin{align}
    (\ref{eq:byidentifiability}) &= \sum_{j \in [q]} \left( 1 - \alpha \beta^2 \right)^{\eta(1, t)-1} \allowdisplaybreaks  \\
    &= \sum_{j \in [q]} \exp\left( (\eta(1, t)-1) \log \left( 1 - \alpha \beta^2 \right) \right)  \allowdisplaybreaks  \\
    &\leq \sum_{j \in [q]}  \exp\left(- \alpha (\eta(1, t)-1) \beta^2 \right)   \label{eq:meanvalue} \allowdisplaybreaks  \\
    &= q \exp\left(- \alpha (\eta(1, t)-1) \beta^2 \right)  \label{eq:conc2}
\end{align}
where (\ref{eq:meanvalue}) follows by an upper bound on natural logarithm: $\log x \leq x - 1 $ for $x > 0$, which can be proven by the Mean Value Theorem and works by selecting $x = 1 - \alpha \beta^2$ in our case.

Next, we provide an upper bound for the covering number $q$ by using the volume ratios. Recall that $\mathcal{S} = [R_{\min} - R_{\max}, R_{\max} - R_{\min}]$ by definition, and hence,
\begin{align}
    q = \mathcal{N}(d, \mathcal{F}, \|\cdot\|) \leq \frac{\mathrm{vol} (\mathcal{F} \color{black} \oplus \mathcal{B}(\mathbf{0}_n,d) \color{black})}{\mathrm{vol} (\mathcal{B}(\mathbf{s}^j, d))} \leq  \frac{\mathrm{vol} (\mathcal{S}^n \color{black} \oplus \mathcal{B}(\mathbf{0}_n,d) \color{black})}{\mathrm{vol} (\mathcal{B}(\mathbf{s}^j, d))} \leq \frac{(R_{\max} - R_{\min} \color{black} + d \color{black})^n}{d^n}
\end{align}
Suppose we have \color{black} $d = \beta/2$\color{black}. Then, combining everything, we obtain \color{black}
\begin{align} 
     \mathbb{P}\left(\inf_{\mathbf{s} \in \mathcal{F}}  L\left(\mathbf{s}, I_t(\boldsymbol{\Pi}_t), \boldsymbol{\Pi}_t\right) < +\infty\right) &\leq  \frac{(2R_{\max} - 2R_{\min} + \beta)^n}{\beta^n} \exp\left(- \alpha (\eta(1, t)-1) \beta^2 \right)  \allowdisplaybreaks  \\
     &\leq  \frac{(2R_{\max} - 2R_{\min} + 1)^n}{\beta^n} \exp\left(- \alpha (\eta(1, t)-1) \beta^2 \right)  \allowdisplaybreaks  \\
     &= \exp \left(- \alpha (\eta(1, t)-1) \beta^2 - n \log \beta + n \log (2R_{\max} - 2R_{\min}) \right) 
\end{align} \color{black}
\Halmos \endproof

\proof{\textbf{Proof of Corollary \ref{cor:concen}.}}
We first highlight that the result of Theorem \ref{thm:concen} is proven for any normalized reward vector $\mathbf{s} \in \mathcal{F} \subset \mathcal{S}^n$ that satisfies $\|\mathbf{s}^0 - \mathbf{s}\|_\infty > \beta$ by definition. Also, recall that the principal's estimator $\widehat{\mathbf{s}}_t \in \mathcal{S}^n$ is defined in (\ref{prblm:estimator}) such that it satisfies $L \left(\widehat{\mathbf{s}}_t, I_t(\boldsymbol{\Pi}_t), \boldsymbol{\Pi}_t\right) < +\infty$. Then, we have the following implication 
\begin{align}
    \left\{\|\mathbf{s}^0 - \widehat{\mathbf{s}}_t \|_\infty > \beta\right\} \subseteq \left\{\exists  \mathbf{s} : \|\mathbf{s}^0 - \mathbf{s}\|_\infty > \beta \text{ and }  L \left(\mathbf{s}, I_t(\boldsymbol{\Pi}_t), \boldsymbol{\Pi}_t\right) < +\infty\right\}
\end{align}
which gives us the desired bound as 
\begin{align}
    \mathbb{P} \left(\|\mathbf{s}^0 - \widehat{\mathbf{s}}_t \|_\infty > \beta \right) &\leq \mathbb{P}\left(\inf_{\mathbf{s} \in \mathcal{F}} L \left(\mathbf{s}, I_t(\boldsymbol{\Pi}_t), \boldsymbol{\Pi}_t\right) < +\infty\right)  \allowdisplaybreaks  \\
    &\leq \color{black} \exp \left(- \alpha (\eta(1, t)-1) \beta^2 - n \log \beta + n \log (2R_{\max} - 2R_{\min}) \right) 
\end{align}
where the last inequality follows by Theorem \ref{thm:concen}. 
\Halmos \endproof

\subsection{Results in Section \ref{sec:learning}} \label{appendix2}
\proof{\textbf{Proof of Lemma \ref{lemma4regret}.}}
We start by defining the indices $\kappa_t \in \argmax_{a \in \mathcal{A}} \widehat{s}_{t,a} $ and $\kappa^0 \in \argmax_{a \in \mathcal{A}} s^0_a $ for notational convenience. Then, we can rewrite the given probability as 
\begin{align}
    \mathbb{P}\left(\max_{a \in \mathcal{A}} \widehat{s}_{t,a} - \widehat{s}_{t,j^*_t} + 2\beta_t \geq \max_{a \in \mathcal{A}} s^0_{a} - s^0_{j^*_t} \right) &= \mathbb{P}\left(\widehat{s}_{t,\kappa_t} - \widehat{s}_{t,j^*_t} + 2\beta_t \geq s^0_{\kappa^0} - s^0_{j^*_t} \right) \allowdisplaybreaks  \\ 
    &= \mathbb{P}\left(2\beta_t \geq s^0_{\kappa^0} - \widehat{s}_{t,\kappa_t} +  \widehat{s}_{t,j^*_t} - s^0_{j^*_t} \right)   \allowdisplaybreaks  \\  
    &= \mathbb{P}\left(2\beta_t \geq s^0_{\kappa^0} - \widehat{s}_{t,\kappa_t} +  \widehat{s}_{t,j^*_t} - s^0_{j^*_t} + \widehat{s}_{t,\kappa^0} - \widehat{s}_{t,\kappa^0} \right)   \allowdisplaybreaks  \\ 
    &= \mathbb{P}\left(2\beta_t \geq \left(s^0_{\kappa^0} - \widehat{s}_{t,\kappa^0}\right) +  \left(\widehat{s}_{t,\kappa^0} - \widehat{s}_{t,\kappa_t}\right) + \left(\widehat{s}_{t,j^*_t} - s^0_{j^*_t} \right) \right)    
\end{align}
where the second term inside the parenthesis satisfies 
\begin{align}
    0 \geq \widehat{s}_{t,\kappa^0} - \widehat{s}_{t,\kappa_t} \label{eq:incentive-proof-2}
\end{align} 
Further, if $\|\mathbf{s}^0 - \widehat{\mathbf{s}}_t \|_\infty \leq \beta_t$, then we have
\begin{align}
    \beta_t &\geq s^0_{\kappa^0} - \widehat{s}_{t,\kappa^0}  \label{eq:incentive-proof-3}\allowdisplaybreaks  \\
     \beta_t &\geq \widehat{s}_{t,j^*_t} - s^0_{j^*_t}  \label{eq:incentive-proof-4}
\end{align}
Hence, we have
\begin{align}
    (\ref{eq:incentive-proof-2})-(\ref{eq:incentive-proof-4}) \Longrightarrow  2\beta_t \geq \left(s^0_{\kappa^0} - \widehat{s}_{t,\kappa^0}\right) +  \left(\widehat{s}_{t,\kappa^0} - \widehat{s}_{t,\kappa_t}\right) + \left(\widehat{s}_{t,j^*_t} - s^0_{j^*_t} \right) \label{eq:incentive-proof-5}
\end{align}
when $\|\mathbf{s}^0 - \widehat{\mathbf{s}}_t \|_\infty \leq \beta_t$ holds. Combining this result with Corollary \ref{cor:concen}, we conclude the proof.
\begin{align}
    &\mathbb{P}\left(2\beta_t \geq \left(s^0_{\kappa^0} - \widehat{s}_{t,\kappa^0}\right) +  \left(\widehat{s}_{t,\kappa^0} - \widehat{s}_{t,\kappa_t}\right) + \left(\widehat{s}_{t,j^*_t} - s^0_{j^*_t} \right) \right) \allowdisplaybreaks  \\
    &\geq \mathbb{P} \left(\|\mathbf{s}^0 - \widehat{\mathbf{s}}_t \|_\infty \leq \beta_t \right) \allowdisplaybreaks  \\
    & \color{black} > 1 - \exp \left(- \alpha (\eta(1, t)-1) \beta_t^2 - n \log \beta_t + n \log (2R_{\max} - 2R_{\min}) \right) 
\end{align}
\Halmos \endproof

\proof{\textbf{Proof of Proposition \ref{prop4regret}.}}
By construction of Algorithm \ref{alg:epsilon}, the arm that the agent picks at time $t \in \mathcal{T}^{\mathrm{xploit}}$ is defined as $i_t(\mathbf{c}(\widehat{\boldsymbol{\theta}}_t, \widehat{\mathbf{s}}_t)) = \argmax_{a \in \mathcal{A}}s^0_a  + c_{a}(\widehat{\boldsymbol{\theta}}_t, \widehat{\mathbf{s}}_t)$. This implies
\begin{align}
    s^0_{i_t(\mathbf{c}(\widehat{\boldsymbol{\theta}}_t, \widehat{\mathbf{s}}_t))}  + c_{i_t(\mathbf{c}(\widehat{\boldsymbol{\theta}}_t, \widehat{\mathbf{s}}_t))}(\widehat{\boldsymbol{\theta}}_t, \widehat{\mathbf{s}}_t) &> s^0_a  + c_{a}(\widehat{\boldsymbol{\theta}}_t, \widehat{\mathbf{s}}_t) \quad \forall a \in \mathcal{A} \setminus \left\{i_t(\mathbf{c}(\widehat{\boldsymbol{\theta}}_t, \widehat{\mathbf{s}}_t)\right\}
\end{align}
Then, the probability that the agent picks arm $j^*_t$ at time $t$ is bounded by
\begin{align}
    \mathbb{P} \left(j^*_t = i_t(\mathbf{c}(\widehat{\boldsymbol{\theta}}_t, \widehat{\mathbf{s}}_t))\right) &= \mathbb{P} \left(s^0_{j^*_t}  + c_{j^*_t}(\widehat{\boldsymbol{\theta}}_t, \widehat{\mathbf{s}}_t) > s^0_a  + c_{a}(\widehat{\boldsymbol{\theta}}_t, \widehat{\mathbf{s}}_t), \ \forall a \in \mathcal{A} \setminus \{j^*_t\} \right)  \allowdisplaybreaks  \\
    &= \mathbb{P} \left(s^0_{j^*_t} +\left(\max_{a \in \mathcal{A}} \widehat{s}_{t,a}\right) - \widehat{s}_{t,j^*_t} + 2\beta_t > s^0_a, \ \forall a \in \mathcal{A} \setminus \{j^*_t\} \right)  \allowdisplaybreaks  \\
    &\geq  \mathbb{P} \left(s^0_{j^*_t} +\left(\max_{a \in \mathcal{A}} \widehat{s}_{t,a}\right) - \widehat{s}_{t,j^*_t} + 2\beta_t \geq \max_{a \in \mathcal{A}} s^0_{a} \right)  \allowdisplaybreaks  \\
    &\color{black} > 1 - \exp \left(- \alpha (\eta(1, t)-1) \beta_t^2 - n \log \beta_t + n \log (2R_{\max} - 2R_{\min}) \right) 
\end{align}
where the last inequality follows by Lemma \ref{lemma4regret}. 
\Halmos \endproof

\proof{\textbf{Proof of Proposition \ref{prop4regret-2}.}}
First, recall that we define the true reward-maximizer action under the oracle incentives in Section \ref{subsec:regret} as 
\begin{equation}
    i(\mathbf{c}(\boldsymbol{\theta}^0, \mathbf{s}^0)) = \argmax_{j \in \mathcal{A}} \widetilde{V}(j, \mathbf{s}^0;\boldsymbol{\theta}^0) = \argmax_{j \in \mathcal{A}} \theta^0_j - \left(\max_{a \in \mathcal{A}} s^0_a \right) + s^0_j
\end{equation} 
Then, we introduce the set $\mathcal{A}_t = \mathcal{A} \setminus \{i_t(\mathbf{c}(\widehat{\boldsymbol{\theta}}_t, \widehat{\mathbf{s}}_t))\}$ for notational convenience and obtain
\begin{align}
     \mathbb{P}\left(i(\mathbf{c}(\boldsymbol{\theta}^0, \mathbf{s}^0)) \neq i_t(\mathbf{c}(\widehat{\boldsymbol{\theta}}_t, \widehat{\mathbf{s}}_t))\right) &\leq \sum_{a \in \mathcal{A}_t} \mathbb{P}\left( \widetilde{V}(i_t(\mathbf{c}(\widehat{\boldsymbol{\theta}}_t, \widehat{\mathbf{s}}_t)), \mathbf{s}^0;\boldsymbol{\theta}^0) < \widetilde{V}(a, \mathbf{s}^0;\boldsymbol{\theta}^0)\right)  \allowdisplaybreaks  \\ 
    &= \sum_{a \in \mathcal{A}_t} \mathbb{P}\left( \theta^0_{i_t(\mathbf{c}(\widehat{\boldsymbol{\theta}}_t, \widehat{\mathbf{s}}_t))} - \theta^0_a < s^0_a - s^0_{i_t(\mathbf{c}(\widehat{\boldsymbol{\theta}}_t, \widehat{\mathbf{s}}_t))}\right) \allowdisplaybreaks  \label{eq:prop6-1}
\end{align} 
We continue by conditioning on whether the action picked by the agent under the exploitation incentives is same as the action with the highest estimated net reward to the principal ($j^*_t$). 
\begin{align}
    (\ref{eq:prop6-1}) &= \sum_{a \in \mathcal{A}_t} \mathbb{P}\left( \theta^0_{i_t(\mathbf{c}(\widehat{\boldsymbol{\theta}}_t, \widehat{\mathbf{s}}_t))} - \theta^0_a < s^0_a - s^0_{i_t(\mathbf{c}(\widehat{\boldsymbol{\theta}}_t, \widehat{\mathbf{s}}_t))} \Big | j^*_t = i_t(\mathbf{c}(\widehat{\boldsymbol{\theta}}_t, \widehat{\mathbf{s}}_t))\right)\mathbb{P}\left(j^*_t = i_t(\mathbf{c}(\widehat{\boldsymbol{\theta}}_t, \widehat{\mathbf{s}}_t))\right) \notag \allowdisplaybreaks \\
    &\quad \quad + \mathbb{P}\left( \theta^0_{i_t(\mathbf{c}(\widehat{\boldsymbol{\theta}}_t, \widehat{\mathbf{s}}_t))} - \theta^0_a < s^0_a - s^0_{i_t(\mathbf{c}(\widehat{\boldsymbol{\theta}}_t, \widehat{\mathbf{s}}_t))} \Big | j^*_t \neq i_t(\mathbf{c}(\widehat{\boldsymbol{\theta}}_t, \widehat{\mathbf{s}}_t))\right)\mathbb{P}\left(j^*_t \neq i_t(\mathbf{c}(\widehat{\boldsymbol{\theta}}_t, \widehat{\mathbf{s}}_t))\right)\allowdisplaybreaks \\
    &\leq \sum_{a \in \mathcal{A}_t} \mathbb{P}\left( \theta^0_{i_t(\mathbf{c}(\widehat{\boldsymbol{\theta}}_t, \widehat{\mathbf{s}}_t))} - \theta^0_a < s^0_a - s^0_{i_t(\mathbf{c}(\widehat{\boldsymbol{\theta}}_t, \widehat{\mathbf{s}}_t))} \Big | j^*_t = i_t(\mathbf{c}(\widehat{\boldsymbol{\theta}}_t, \widehat{\mathbf{s}}_t))\right) + \mathbb{P}\left(j^*_t \neq i_t(\mathbf{c}(\widehat{\boldsymbol{\theta}}_t, \widehat{\mathbf{s}}_t))\right)\allowdisplaybreaks \\
    &\leq \sum_{a \in \mathcal{A}_t} \mathbb{P}\left( \theta^0_{j^*_t} - \theta^0_a < s^0_a - s^0_{j^*_t} \right) + \color{black} \exp \left(- \alpha (\eta(1, t)-1) \beta_t^2 - n \log \beta_t + n \log (2R_{\max} - 2R_{\min}) \right) \label{eq:prop6-2} \allowdisplaybreaks 
\end{align}
where the last inequality follows by Proposition \ref{prop4regret}. Now, by definition of $j^*_t$, we have $ \widetilde{V}(a, \widehat{\mathbf{s}}_t;\widehat{\boldsymbol{\theta}}) < \widetilde{V}(j^*_t, \widehat{\mathbf{s}}_t;\widehat{\boldsymbol{\theta}}), \ \forall a \in \mathcal{A}, a \neq j^*_t$ which implies $\widehat{\theta}_{t,a} - \widehat{\theta}_{t,j^*_t} < \widehat{s}_{t,j^*_t} - \widehat{s}_{t,a}, \forall a \in \mathcal{A}, a \neq j^*_t$. Combining this inequality with the first term in (\ref{eq:prop6-2}), we have
\begin{align}
    &\mathbb{P}\left( \theta^0_{j^*_t} - \theta^0_a < s^0_a - s^0_{j^*_t} \right) \notag \allowdisplaybreaks \\
    &= \mathbb{P}\left((\theta^0_{j^*_t} - \widehat{\theta}_{t,j^*_t}) + (\widehat{\theta}_{t,a} - \theta^0_a) < (s^0_a - \widehat{s}_{t,a} ) + ( \widehat{s}_{t,j^*_t}- s^0_{j^*_t}) \right) \allowdisplaybreaks \\
    &= \mathbb{P}\left((\theta^0_{j^*_t} - \widehat{\theta}_{t,j^*_t}) + (\widehat{\theta}_{t,a} - \theta^0_a) < (s^0_a - \widehat{s}_{t,a}) + ( \widehat{s}_{t,j^*_t}- s^0_{j^*_t}) \Big | \|\mathbf{s}^0 - \widehat{\mathbf{s}}_t \|_\infty \leq \beta_t \right)  \mathbb{P}\left(\|\mathbf{s}^0 - \widehat{\mathbf{s}}_t \|_\infty \leq \beta_t \right) \notag \allowdisplaybreaks  \\
    &\quad + \mathbb{P}\left((\theta^0_{j^*_t} - \widehat{\theta}_{t,j^*_t}) + (\widehat{\theta}_{t,a} - \theta^0_a) < (s^0_a - \widehat{s}_{t,a}) + ( \widehat{s}_{t,j^*_t}- s^0_{j^*_t}) \Big | \|\mathbf{s}^0 - \widehat{\mathbf{s}}_t \|_\infty > \beta_t \right)  \mathbb{P}\left(\|\mathbf{s}^0 - \widehat{\mathbf{s}}_t \|_\infty > \beta_t \right) \allowdisplaybreaks  \\
    &\leq \mathbb{P}\left((\theta^0_{j^*_t} - \widehat{\theta}_{t,j^*_t}) + (\widehat{\theta}_{t,a} - \theta^0_a) < (s^0_a - \widehat{s}_{t,a}) + ( \widehat{s}_{t,j^*_t}- s^0_{j^*_t}) \Big | \|\mathbf{s}^0 - \widehat{\mathbf{s}}_t \|_\infty \leq \beta_t \right) + \mathbb{P}\left(\|\mathbf{s}^0 - \widehat{\mathbf{s}}_t \|_\infty > \beta_t \right) \allowdisplaybreaks  \\
    &\leq \mathbb{P}\left((\theta^0_{j^*_t} - \widehat{\theta}_{t,j^*_t}) + (\widehat{\theta}_{t,a} - \theta^0_a) < 2\beta_t \right) + \color{black} \exp \left(- \alpha (\eta(1, t)-1) \beta_t^2 - n \log \beta_t + n \log (2R_{\max} - 2R_{\min}) \right) \allowdisplaybreaks \label{eq:pgg}
\end{align}
where the last line follows by the finite-sample concentration bound in Corollary \ref{cor:concen}. Next, we bound the first term above as follows.  
\begin{align}
     &\mathbb{P}\left(\theta^0_{j^*_t} - \widehat{\theta}_{t,j^*_t} < 2\beta_t -(\widehat{\theta}_{t,a} - \theta^0_a) \right) \notag \allowdisplaybreaks \\
     &=  \mathbb{P}\left(\theta^0_{j^*_t} - \widehat{\theta}_{t,j^*_t} < 2\beta_t -(\widehat{\theta}_{t,a} - \theta^0_a) \Big | \widehat{\theta}_{t,a} - \theta^0_a < 3\beta_t \right) \mathbb{P}\left(\widehat{\theta}_{t,a} - \theta^0_a < 3\beta_t \right)  \notag \allowdisplaybreaks \\
     &\quad + \mathbb{P}\left(\theta^0_{j^*_t} - \widehat{\theta}_{t,j^*_t} < 2\beta_t -(\widehat{\theta}_{t,a} - \theta^0_a) \Big | \widehat{\theta}_{t,a} - \theta^0_a \geq 3\beta_t \right) \mathbb{P}\left(\widehat{\theta}_{t,a} - \theta^0_a \geq 3\beta_t \right) \allowdisplaybreaks \\
     &\leq \mathbb{P}\left(\widehat{\theta}_{t,j^*_t} - \theta^0_{j^*_t} > \beta_t \right) + \mathbb{P}\left(\widehat{\theta}_{t,a} - \theta^0_a \geq 3\beta_t \right) \allowdisplaybreaks \\
     &\leq \mathbb{P}\left(\widehat{\theta}_{t,j^*_t} - \theta^0_{j^*_t} > \beta_t \right) + \mathbb{P}\left(\widehat{\theta}_{t,a} - \theta^0_a \geq \beta_t \right) \allowdisplaybreaks \label{eq:pgg2}
\end{align}
Notice that we bound the two probability terms in the last line in the same way by definition of $\widehat{\theta}_{t,a}$'s (\ref{eq:theta-hat}). For any $a \in \mathcal{A}$, let $\overline{T}(a, t) = \left|\{\tau \in \Lambda(1, t): i_\tau(\boldsymbol{\pi}_\tau) = a\} \right|$ be the number of exploration steps up to time $t$ at which the agent's reward-maximizer arm is action $a$. Thus, $\overline{T}(a, t)$ is the sum of $\eta(1, t)$ independent Bernoulli random variables with success probabilities $\mathbb{P} \left(a = \argmax_{a' \in \mathcal{A}} {s^0_{a'} + \pi_{\tau, a'}} \right)$. Then, 
\begin{align}
    \mathbb{E} \overline{T}(a, t) = \sum_{\tau \in \Lambda(1, t)} \mathbb{P} \left(a = \argmax_{a' \in \mathcal{A}} {s^0_{a'} + \pi_{\tau, a'}} \right) &\geq \sum_{\tau \in \Lambda(1, t)} \left[ \sum_{a' \in \mathcal{A}} \frac{(s^0_{a} - s^0_{a'} + \overline{C}- \underline{C})^2}{2(\overline{C}- \underline{C})^2} \right] \label{eq:prop6-3} \allowdisplaybreaks \\
    &= n \frac{(s^0_{a} + \gamma)^2}{2(\overline{C}- \underline{C})^2}  \eta(1, t)\label{eq:prop6-4}
\end{align}
where the term in the square brackets in (\ref{eq:prop6-3}) follows by using the cdf derived in (\ref{eq:cdf}). Since the cdf is defined as a piecewise function, it suffices to only consider the case when $\underline{C} - \overline{C} \leq s^0_{a} - s^0_{a'} < 0$ holds to find a lower bound on the probability $\mathbb{P} \left(a = \argmax_{a' \in \mathcal{A}} {s^0_{a'} + \pi_{\tau, a'}} \right)$. Further, (\ref{eq:prop6-4}) follows since by definition we know that $s^0_{a'} \leq R_{\max} - R_{\min} = \overline{C}- \underline{C} - \gamma$ for all $a' \in \mathcal{A}$. 

Now, observing that $\overline{T}(a, t) \leq T(a, t)$, we can use Hoeffding's Inequality \citep{boucheron2013concentration} to proceed. For any $a \in \mathcal{A}$, 
\begin{align}
    &\mathbb{P}\left(\widehat{\theta}_{t,a} - \theta^0_{a} > \beta_t \right)  \notag \allowdisplaybreaks \\
    &\hspace{-.1cm}= \mathbb{P}\left(\widehat{\theta}_{t,a} - \theta^0_{a} > \beta_t \big | \overline{T}(a, t) > \mathbb{E} \overline{T}(a, t) \textstyle \frac{4\alpha(R_{\max} - R_{\min})^2(\overline{C}- \underline{C})^2}{n(s^0_{a} + \gamma)^2} \right)\mathbb{P}\left(\overline{T}(a, t) > \mathbb{E} \overline{T}(a, t) \textstyle \frac{4\alpha(R_{\max} - R_{\min})^2(\overline{C}- \underline{C})^2}{n(s^0_{a} + \gamma)^2}\right)  \notag \allowdisplaybreaks \\
    &\hspace{-.1cm}+ \mathbb{P}\left(\widehat{\theta}_{t,a} - \theta^0_{a} > \beta_t \big | \overline{T}(a, t) \leq \mathbb{E} \overline{T}(a, t)\textstyle \frac{4\alpha(R_{\max} - R_{\min})^2(\overline{C}- \underline{C})^2}{n(s^0_{a} + \gamma)^2}\right)\mathbb{P}\left(\overline{T}(a, t) \leq \mathbb{E} \overline{T}(a, t)\textstyle \frac{4\alpha(R_{\max} - R_{\min})^2(\overline{C}- \underline{C})^2}{n(s^0_{a} + \gamma)^2}\right) \allowdisplaybreaks \\
    &\leq \mathbb{P}\left(\widehat{\theta}_{t,a} - \theta^0_{a} > \beta_t \big | \overline{T}(a, t) > \mathbb{E} \overline{T}(a, t)\textstyle \frac{4\alpha(R_{\max} - R_{\min})^2(\overline{C}- \underline{C})^2}{n(s^0_{a} + \gamma)^2}\right) \notag \allowdisplaybreaks \\
    &\ + \mathbb{P}\left(\overline{T}(a, t) \leq \mathbb{E} \overline{T}(a, t) \textstyle\frac{4\alpha(R_{\max} - R_{\min})^2(\overline{C}- \underline{C})^2}{n(s^0_{a} + \gamma)^2}\right) \allowdisplaybreaks \\
    &\leq \exp \left(-\frac{2\textstyle\frac{4\alpha(R_{\max} - R_{\min})^2(\overline{C}- \underline{C})^2}{n(s^0_{a} + \gamma)^2}\mathbb{E} \overline{T}(a, t) \beta^2_{t}}{4\alpha(R_{\max} - R_{\min})^2}\right) + \exp \left(- 2\eta(1, t) \left(\textstyle\frac{4\alpha(R_{\max} - R_{\min})^2(\overline{C}- \underline{C})^2}{n(s^0_{a} + \gamma)^2}\right)^2\mathbb{E} \overline{T}(a, t)^2\right)  \allowdisplaybreaks \\
    &\leq \exp \left(- \eta(1, t) \frac{\log (\eta(1, t)-1)}{\eta(1, t)-1}\right) + \exp(-\eta(1, t)^3)  \allowdisplaybreaks \label{plugin}\\
    &\leq \frac{1}{\eta(1, t)-1} + \frac{1}{\eta(1, t)^3}  \allowdisplaybreaks \\
    &\leq \frac{2}{\eta(1, t)-1} \allowdisplaybreaks 
\end{align}
where (\ref{plugin}) follows by substituting the lower bound in (\ref{eq:prop6-4}) and $\beta_t = \sqrt{\frac{\log (\eta(1, t)-1)}{\alpha (\eta(1, t)-1)}}$. Lastly, we combine this result with (\ref{eq:prop6-2}), (\ref{eq:pgg}), and (\ref{eq:pgg2}) and conclude our proof.  
\begin{align}
   &\mathbb{P}\left(i(\mathbf{c}(\boldsymbol{\theta}^0, \mathbf{s}^0)) \neq i_t(\mathbf{c}(\widehat{\boldsymbol{\theta}}_t, \widehat{\mathbf{s}}_t))\right) \notag  \allowdisplaybreaks  \\
   &\leq \frac{4n}{\eta(1, t)-1} + 2n \color{black} \exp \left(- \alpha (\eta(1, t)-1) \beta_t^2 - n \log \beta_t + n \log (2R_{\max} - 2R_{\min}) \right)  \allowdisplaybreaks  \\
   &= \frac{4n}{\eta(1, t)-1} + 2n \color{black} \exp\left(- \log (\eta(1, t)-1) - n\log \sqrt{\frac{\log (\eta(1, t)-1)}{\alpha (\eta(1, t)-1)}}+ n \log (2R_{\max} - 2R_{\min}) \right)  \allowdisplaybreaks  \\
    &= \frac{4n}{\eta(1, t)-1} + \color{black}  \frac{2n(2R_{\max} - 2R_{\min})^n \alpha^{n/2} }{[(\eta(1, t)-1) \log (\eta(1, t)-1)]^{n/2}} \allowdisplaybreaks 
\end{align} 
\Halmos \endproof

\proof{\textbf{Proof of Theorem \ref{thm:regret}.}}
The expected net reward of the principal defined in (\ref{eq:deter-epsilon}) has two main components: cost incurred due to the offered incentives and mean reward collected through the arm chosen by the agent. Accordingly, we decompose our regret notion (\ref{eq:regretdefn}) into two main parts as follows.
\begin{align}
     \mathrm{Regret}\left(\Pi_{\epsilon, T} \right) &= \sum_{t \in \mathcal{T}} V(\mathbf{c}(\boldsymbol{\theta}^0, \mathbf{s}^0); \boldsymbol{\theta}^0) - V_t(\boldsymbol{\pi}_t; \boldsymbol{\theta}^0)  \allowdisplaybreaks  \\
    &= \sum_{t \in \mathcal{T}} \left[ {\color{black}\pi_{t, i_t(\boldsymbol{\pi}_t)} - c_{i(\mathbf{c}(\boldsymbol{\theta}^0, \mathbf{s}^0))} (\boldsymbol{\theta}^0, \mathbf{s}^0) } \right] + \sum_{t \in \mathcal{T}} \left[\theta^0_{i(\mathbf{c}(\boldsymbol{\theta}^0, \mathbf{s}^0))} -\theta^0_{i_t(\boldsymbol{\pi}_t)} \right] \allowdisplaybreaks  \label{eq:divide-regret}
\end{align}
First, we provide an upper bound for the first part of (\ref{eq:divide-regret}).
\begin{align} \color{black}
    \sum_{t \in \mathcal{T}} \left[ \pi_{t, i_t(\boldsymbol{\pi}_t)} - c_{i(\mathbf{c}(\boldsymbol{\theta}^0, \mathbf{s}^0)} (\boldsymbol{\theta}^0, \mathbf{s}^0) \right] &\color{black}\leq \sum_{t \in \mathcal{T}^{\mathrm{xplore}}} \left[\pi_{t, i_t(\boldsymbol{\pi}_t)} - c_{i(\mathbf{c}(\boldsymbol{\theta}^0, \mathbf{s}^0)} (\boldsymbol{\theta}^0, \mathbf{s}^0)\right] \notag \\
    &\color{black} \quad + \sum_{t \in \mathcal{T}^{\mathrm{xploit}}} \left[c_{j^*_t}(\widehat{\boldsymbol{\theta}}_t, \widehat{\mathbf{s}}_t) - c_{i(\mathbf{c}(\boldsymbol{\theta}^0, \mathbf{s}^0)}(\boldsymbol{\theta}^0, \mathbf{s}^0) \right] \label{eq:deter-regret1}
\end{align}
Notice that the cardinalities $|\mathcal{T}^{\mathrm{xplore}}|$ and $| \mathcal{T}^{\mathrm{xploit}}|$ are random variables. Then, 
\begin{align}
     \mathbb{E} \bigg[\sum_{t \in \mathcal{T}^{\mathrm{xplore}}} \left[ {\color{black} \pi_{t, i_t(\boldsymbol{\pi}_t)} - c_{i(\mathbf{c}(\boldsymbol{\theta}^0, \mathbf{s}^0)} (\boldsymbol{\theta}^0, \mathbf{s}^0) }\right] \Big| \mathcal{T}^{\mathrm{xplore}} \bigg] &\leq \color{black} (\overline{C}  - \underline{C}) |\mathcal{T}^{\mathrm{xplore}}| 
\end{align}
Taking the expectation of both sides of the last inequality, we obtain the following upper bound for the first summation term in (\ref{eq:deter-regret1}). 
\begin{align}
    \sum_{t \in \mathcal{T}^{\mathrm{xplore}}} \left[ {\color{black} \pi_{t, i_t(\boldsymbol{\pi}_t)} - c_{i(\mathbf{c}(\boldsymbol{\theta}^0, \mathbf{s}^0)} (\boldsymbol{\theta}^0, \mathbf{s}^0) }\right] \leq (\overline{C}  - \underline{C}) \mathbb{E} |\mathcal{T}^{\mathrm{xplore}}|  
    &= (\overline{C}  - \underline{C})  \sum_{t=1}^T \min\left\{1, \frac{m}{t}\right\}  \label{eq:deter-regret1-0}  \allowdisplaybreaks  \\
    &\leq (\overline{C}  - \underline{C})  \sum_{t=1}^T \frac{m}{t}  \allowdisplaybreaks  \\
    &\leq (\overline{C}  - \underline{C}) \left(m + \int_{t=1}^T \frac{m}{t} \right)   \allowdisplaybreaks  \\
    &= m (\overline{C}  - \underline{C}) (1 + \log T)  \label{eq:deter-regret1-1}
\end{align}
where the last equality follows by the finite sum formula of the harmonic series. Next, we bound the second part of (\ref{eq:deter-regret1}) as follows. 
\begin{align}
    &\mathbb{E} \left[ \sum_{t \in \mathcal{T}^{\mathrm{xploit}}} \left[ {\color{black} c_{j^*_t}(\widehat{\boldsymbol{\theta}}_t, \widehat{\mathbf{s}}_t) - c_{i(\mathbf{c}(\boldsymbol{\theta}^0, \mathbf{s}^0)}(\boldsymbol{\theta}^0, \mathbf{s}^0) } \right] \Big| \mathcal{T}^{\mathrm{xploit}} \right] \nonumber  \allowdisplaybreaks  \\ 
    &= \sum_{t \in \mathcal{T}^{\mathrm{xploit}}} \mathbb{E} \left[{\color{black} c_{j^*_t}(\widehat{\boldsymbol{\theta}}_t, \widehat{\mathbf{s}}_t) - c_{i(\mathbf{c}(\boldsymbol{\theta}^0, \mathbf{s}^0)}(\boldsymbol{\theta}^0, \mathbf{s}^0) }  \Big| \| \mathbf{s}^0 - \widehat{\mathbf{s}}_t \|_\infty \leq \beta_t \right] \mathbb{P} \left(\|\mathbf{s}^0 - \widehat{\mathbf{s}}_t \|_\infty \leq \beta_t  \right) \nonumber  \allowdisplaybreaks  \\ 
    &\quad + \sum_{t \in \mathcal{T}^{\mathrm{xploit}}}  \mathbb{E} \left[{\color{black} c_{j^*_t}(\widehat{\boldsymbol{\theta}}_t, \widehat{\mathbf{s}}_t) - c_{i(\mathbf{c}(\boldsymbol{\theta}^0, \mathbf{s}^0)}(\boldsymbol{\theta}^0, \mathbf{s}^0) } \Big| \| \mathbf{s}^0 - \widehat{\mathbf{s}}_t \|_\infty > \beta_t \right] \mathbb{P} \left(\|\mathbf{s}^0 - \widehat{\mathbf{s}}_t \|_\infty > \beta_t \right) \allowdisplaybreaks  \\
    &\leq \sum_{t \in \mathcal{T}^{\mathrm{xploit}}}  \mathbb{E} \left[{\color{black} c_{j^*_t}(\widehat{\boldsymbol{\theta}}_t, \widehat{\mathbf{s}}_t) - c_{i(\mathbf{c}(\boldsymbol{\theta}^0, \mathbf{s}^0)}(\boldsymbol{\theta}^0, \mathbf{s}^0) }  \Big| \| \mathbf{s}^0 - \widehat{\mathbf{s}}_t \|_\infty \leq \beta_t \right] \nonumber  \allowdisplaybreaks  \\ 
    &\quad + \sum_{t \in \mathcal{T}^{\mathrm{xploit}}} \left(\overline{C} - \underline{C} \right) \color{black} \exp \left(- \alpha (\eta(1, t)-1) \beta_t^2 - n \log \beta_t + n \log (2R_{\max} - 2R_{\min}) \right) \label{eq:deter-regret1-2-0} \allowdisplaybreaks 
\end{align}
where the last line follows by Corollary \ref{cor:concen}. To compute an upper bound for the first term in the last inequality, we proceed as
\begin{align}
       &\sum_{t \in \mathcal{T}^{\mathrm{xploit}}} \mathbb{E} \left[{\color{black} c_{j^*_t}(\widehat{\boldsymbol{\theta}}_t, \widehat{\mathbf{s}}_t) - c_{i(\mathbf{c}(\boldsymbol{\theta}^0, \mathbf{s}^0)}(\boldsymbol{\theta}^0, \mathbf{s}^0) }  \Big| \| \mathbf{s}^0 - \widehat{\mathbf{s}}_t \|_\infty \leq \beta_t \right] \nonumber  \allowdisplaybreaks  \\
       &= \sum_{t \in \mathcal{T}^{\mathrm{xploit}}} \mathbb{E} \left[{\color{black} c_{j^*_t}(\widehat{\boldsymbol{\theta}}_t, \widehat{\mathbf{s}}_t) - c_{i(\mathbf{c}(\boldsymbol{\theta}^0, \mathbf{s}^0)}(\boldsymbol{\theta}^0, \mathbf{s}^0) } \Big| \| \mathbf{s}^0 - \widehat{\mathbf{s}}_t \|_\infty \leq \beta_t, j^*_t = i(\mathbf{c}(\boldsymbol{\theta}^0, \mathbf{s}^0)) \right] \mathbb{P} \left(j^*_t = i(\mathbf{c}(\boldsymbol{\theta}^0, \mathbf{s}^0)) \right) \nonumber  \allowdisplaybreaks  \\
       &\hspace{2.2cm} + \mathbb{E} \left[{\color{black} c_{j^*_t}(\widehat{\boldsymbol{\theta}}_t, \widehat{\mathbf{s}}_t) - c_{i(\mathbf{c}(\boldsymbol{\theta}^0, \mathbf{s}^0)}(\boldsymbol{\theta}^0, \mathbf{s}^0) }  \Big| \| \mathbf{s}^0 - \widehat{\mathbf{s}}_t \|_\infty \leq \beta_t, j^*_t \neq i(\mathbf{c}(\boldsymbol{\theta}^0, \mathbf{s}^0))\right]\mathbb{P} \left(j^*_t \neq i(\mathbf{c}(\boldsymbol{\theta}^0, \mathbf{s}^0)) \right)  \allowdisplaybreaks  \\
       &\leq  \sum_{t \in \mathcal{T}^{\mathrm{xploit}}} \left( \max_{a \in \mathcal{A}} \widehat{s}_{t,a} - \widehat{s}_{t,j^*_t}  + 2\beta_t - \max_{a \in \mathcal{A}} s^0_a  + s^0_{i(\mathbf{c}(\boldsymbol{\theta}^0, \mathbf{s}^0))} \Big| \| \mathbf{s}^0 - \widehat{\mathbf{s}}_t \|_\infty \leq \beta_t, j^*_t = i(\mathbf{c}(\boldsymbol{\theta}^0, \mathbf{s}^0)) \right) \nonumber  \allowdisplaybreaks  \\
       &\hspace{1.8cm} + \left(\overline{C} - \underline{C} \right) \mathbb{P} \left(j^*_t \neq i(\mathbf{c}(\boldsymbol{\theta}^0, \mathbf{s}^0))\right)  
       \intertext{As before, we use the indices $\kappa_t \in \argmax_{a \in \mathcal{A}} \widehat{s}_{t,a}$ and $\kappa^0 \in \argmax_{a \in \mathcal{A}} s^0_a $ for notational convenience.}
       &= \sum_{t \in \mathcal{T}^{\mathrm{xploit}}} \left( \widehat{s}_{t,\kappa_t} - \widehat{s}_{t,j^*_t} + 2\beta_t - s^0_{\kappa^0}  + s^0_{i(\mathbf{c}(\boldsymbol{\theta}^0, \mathbf{s}^0))} \Big| \| \mathbf{s}^0 - \widehat{\mathbf{s}}_t \|_\infty \leq \beta_t, j^*_t = i(\mathbf{c}(\boldsymbol{\theta}^0, \mathbf{s}^0)) \right) \nonumber  \allowdisplaybreaks  \\
       &\hspace{1.8cm} + \left(\overline{C} - \underline{C} \right) \mathbb{P} \left(j^*_t \neq i(\mathbf{c}(\boldsymbol{\theta}^0, \mathbf{s}^0)) \right)  \allowdisplaybreaks  \\
       &= \sum_{t \in \mathcal{T}^{\mathrm{xploit}}} \left( (\widehat{s}_{t,\kappa_t}  - s^0_{\kappa_t}) + (s^0_{\kappa_t} - s^0_{\kappa^0}) + (s^0_{i(\mathbf{c}(\boldsymbol{\theta}^0, \mathbf{s}^0))} - \widehat{s}_{t,j^*_t}) + 2\beta_t \Big| \| \mathbf{s}^0 - \widehat{\mathbf{s}}_t \|_\infty \leq \beta_t, j^*_t = i(\mathbf{c}(\boldsymbol{\theta}^0, \mathbf{s}^0)) \right) \nonumber  \allowdisplaybreaks  \\
       &\hspace{1.8cm} + \left(\overline{C} - \underline{C} \right) \mathbb{P} \left(j^*_t \neq i(\mathbf{c}(\boldsymbol{\theta}^0, \mathbf{s}^0)) \right)  \allowdisplaybreaks  \\
       &\leq \sum_{t \in \mathcal{T}^{\mathrm{xploit}}} 4\beta_t + \left(\overline{C} - \underline{C} \right) \mathbb{P} \left(j^*_t \neq i(\mathbf{c}(\boldsymbol{\theta}^0, \mathbf{s}^0)) \right)  \label{eq:ggg}
\end{align}
At this step, we continue by observing that
\begin{align}
    \mathbb{P} \left(j^*_t = i(\mathbf{c}(\boldsymbol{\theta}^0, \mathbf{s}^0)) \right) &\geq  \mathbb{P} \left( j^*_t = i_t(\mathbf{c}(\widehat{\boldsymbol{\theta}}_t, \widehat{\mathbf{s}}_t)), \ i_t(\mathbf{c}(\widehat{\boldsymbol{\theta}}_t, \widehat{\mathbf{s}}_t)) = i(\mathbf{c}(\boldsymbol{\theta}^0, \mathbf{s}^0)) \right) 
\end{align}
which implies 
\begin{align}
    &\mathbb{P} \left(j^*_t \neq i(\mathbf{c}(\boldsymbol{\theta}^0, \mathbf{s}^0)) \right) \nonumber  \allowdisplaybreaks  \\
    &\leq 1 - \mathbb{P} \left( j^*_t = i_t(\mathbf{c}(\widehat{\boldsymbol{\theta}}_t, \widehat{\mathbf{s}}_t)), \ i_t(\mathbf{c}(\widehat{\boldsymbol{\theta}}_t, \widehat{\mathbf{s}}_t)) = i(\mathbf{c}(\boldsymbol{\theta}^0, \mathbf{s}^0)) \right)  \allowdisplaybreaks  \\
    &= 1 - \left[1 -  \mathbb{P} \left(j^*_t \neq i_t(\mathbf{c}(\widehat{\boldsymbol{\theta}}_t, \widehat{\mathbf{s}}_t)) \  \bigcup \ i_t(\mathbf{c}(\widehat{\boldsymbol{\theta}}_t, \widehat{\mathbf{s}}_t)) \neq i(\mathbf{c}(\boldsymbol{\theta}^0, \mathbf{s}^0)) \right) \right] \label{eq:probidentity} \allowdisplaybreaks  \\
    &\leq \mathbb{P} \left(j^*_t \neq i_t(\mathbf{c}(\widehat{\boldsymbol{\theta}}_t, \widehat{\mathbf{s}}_t)) \right)  + \mathbb{P} \left(i_t(\mathbf{c}(\widehat{\boldsymbol{\theta}}_t, \widehat{\mathbf{s}}_t)) \neq i(\mathbf{c}(\boldsymbol{\theta}^0, \mathbf{s}^0)) \right) \label{eq:unionbound} \allowdisplaybreaks  \\
    &\leq \color{black} \exp \left(- \alpha (\eta(1, t)-1) \beta_t^2 - n \log \beta_t + n \log (2R_{\max} - 2R_{\min}) \right) \nonumber \allowdisplaybreaks \\
    &\quad + \frac{4n}{\eta(1, t)-1} +  \color{black} \frac{2n(2R_{\max} - 2R_{\min})^n \alpha^{n/2} }{[(\eta(1, t)-1) \log (\eta(1, t)-1)]^{n/2}} 
\end{align}
where (\ref{eq:probidentity}) follows by the fact that $\mathbb{P} (\cap_i A_i) = 1 - \mathbb{P} (\cup_i \overline{A}_i)$ for a set of events $A_i$'s, (\ref{eq:unionbound}) follows by the Boole's inequality (a.k.a. union bound), and the last inequality follows by Propositions \ref{prop4regret} and \ref{prop4regret-2}.

Combining the last result with (\ref{eq:deter-regret1-2-0}) and (\ref{eq:ggg}) for $\beta_t = \sqrt{\frac{\log (\eta(1, t)-1)}{\alpha (\eta(1, t)-1)}}$, we obtain
\begin{align}
    &\mathbb{E} \left[ \sum_{t \in \mathcal{T}^{\mathrm{xploit}}} \sum_{a \in \mathcal{A}} \left[c_{a}(\widehat{\boldsymbol{\theta}}_t, \widehat{\mathbf{s}}_t) - c_{a}(\boldsymbol{\theta}^0, \mathbf{s}^0) \right] \Big| \mathcal{T}^{\mathrm{xploit}} \right] \nonumber  \allowdisplaybreaks  \\
    &\leq \sum_{t \in \mathcal{T}^{\mathrm{xploit}}}4\beta_t + \left(\overline{C} - \underline{C} \right) \Bigg(2\color{black} \exp \left(- \alpha (\eta(1, t)-1) \beta_t^2 - n \log \beta_t + n \log (2R_{\max} - 2R_{\min}) \right) \nonumber  \allowdisplaybreaks  \\
    &\hspace{5cm} + \frac{4n}{\eta(1, t)-1} + \color{black} \frac{2n(2R_{\max} - 2R_{\min})^n \alpha^{n/2} }{[(\eta(1, t)-1) \log (\eta(1, t)-1)]^{n/2}} \color{black} \Bigg) \label{eq:continue0}\allowdisplaybreaks  \\
    &= \sum_{t \in \mathcal{T}^{\mathrm{xploit}}} 4\sqrt{\frac{\log (\eta(1, t)-1)}{\alpha (\eta(1, t)-1)}} + \sum_{t \in \mathcal{T}^{\mathrm{xploit}}} \frac{4n\left(\overline{C} - \underline{C} \right)\color{black} (2R_{\max} - 2R_{\min})^n \alpha^{n/2}}{\color{black}[(\eta(1, t)-1) \log (\eta(1, t)-1)]^{n/2}} + \sum_{t \in \mathcal{T}^{\mathrm{xploit}}} \frac{4n\left(\overline{C} - \underline{C} \right)}{\eta(1, t)-1}   \allowdisplaybreaks \\
    &\leq \sum_{t \in \mathcal{T}^{\mathrm{xploit}}} 4\sqrt{\frac{\log (\eta(1, t)-1)}{\alpha (\eta(1, t)-1)}} + \sum_{t \in \mathcal{T}^{\mathrm{xploit}}} \frac{4n\left(\overline{C} - \underline{C} \right)\color{black} (2R_{\max} - 2R_{\min})^n \alpha^{n/2}}{\sqrt{(\eta(1, t)-1) \log (\eta(1, t)-1)}} + \sum_{t \in \mathcal{T}^{\mathrm{xploit}}} \frac{4n\left(\overline{C} - \underline{C} \right)}{\eta(1, t)-1}   \allowdisplaybreaks \allowdisplaybreaks \label{eq:continue1-pre}\\
    &\leq \sum_{t \in \mathcal{T}^{\mathrm{xploit}}} 4\sqrt{\frac{\log (\eta(1, t)-1)}{\alpha (\eta(1, t)-1)}} + \sum_{t \in \mathcal{T}^{\mathrm{xploit}}} \frac{4n\left(\overline{C} - \underline{C} \right)\color{black} (2R_{\max} - 2R_{\min})^n \alpha^{n/2}}{\sqrt{\eta(1, t)-2}} + \sum_{t \in \mathcal{T}^{\mathrm{xploit}}} \frac{4n\left(\overline{C} - \underline{C} \right)}{\eta(1, t)-1}   \allowdisplaybreaks \label{eq:continue1}
\end{align}
where the second term in \textcolor{black}{(\ref{eq:continue1-pre}) follows as $n \geq 1$ and} (\ref{eq:continue1}) follows by the following bound on the natural logarithm: $1 - 1/x \leq \log x$ for $x > 0$. Now, recall that the principal's $\epsilon$-Greedy Algorithm (\ref{alg:epsilon}) performs pure exploration over the first $m$ steps of the finite time horizon $\mathcal{T}$. This implies $\eta(1, t) \geq m$ and $t \geq m + 1$ for any $t \in \mathcal{T}^{\mathrm{xploit}}$. Then, because the terms of the three summations in (\ref{eq:continue1}) are monotone decreasing functions of $\eta(1, t)$ for $m \geq 4$, we can bound these finite summations with the corresponding definite integrals plus the first terms of these series.
\begin{align}
    (\ref{eq:continue1}) &\leq \frac{4}{\sqrt{\alpha}} \int_{x=m}^{|\mathcal{T}^{\mathrm{xploit}}|} \sqrt{\frac{\log(x-1)}{x-1}}dx + \int_{t=m}^{|\mathcal{T}^{\mathrm{xploit}}|} \frac{4n\left(\overline{C} - \underline{C} \right)\color{black} (2R_{\max} - 2R_{\min})^n \alpha^{n/2}}{\sqrt{x-2}}dx \notag \allowdisplaybreaks\\
    &\quad + \int_{t=m}^{|\mathcal{T}^{\mathrm{xploit}}|} \frac{4n\left(\overline{C} - \underline{C} \right)}{x-1}dx + B_1 \allowdisplaybreaks
    \intertext{where $B_1 = \frac{4}{\sqrt{\alpha}}\sqrt{\frac{\log(m-1)}{m-1}} + \frac{4n\left(\overline{C} - \underline{C} \right)\color{black} (2R_{\max} - 2R_{\min})^n \alpha^{n/2}}{\sqrt{m-2}} + \frac{4n\left(\overline{C} - \underline{C} \right)}{m-1}$,}
    &\leq \frac{8}{\sqrt{\alpha}} \sqrt{(|\mathcal{T}^{\mathrm{xploit}}|-1) \log (|\mathcal{T}^{\mathrm{xploit}}|-1)} + 8n\left(\overline{C} - \underline{C} \right)\color{black} (2R_{\max} - 2R_{\min})^n \alpha^{n/2} \color{black}\sqrt{|\mathcal{T}^{\mathrm{xploit}}|-2} \nonumber  \allowdisplaybreaks  \\
    &\quad + 4n\left(\overline{C} - \underline{C} \right) \log (|\mathcal{T}^{\mathrm{xploit}}|-1) + B_1 \allowdisplaybreaks  \\
    &\leq \frac{8}{\sqrt{\alpha}} \sqrt{T \log T} + 8n\left(\overline{C} - \underline{C} \right)\color{black} (2R_{\max} - 2R_{\min})^n \alpha^{n/2} \color{black}\sqrt{T} + 4n\left(\overline{C} - \underline{C} \right) \log T + B_1 \allowdisplaybreaks \label{eq:continue2}
 \end{align}
By taking the expectation of the last result, we have 
\begin{multline}
\sum_{t \in \mathcal{T}^{\mathrm{xploit}}} \sum_{a \in \mathcal{A}} \left[c_{a}(\widehat{\boldsymbol{\theta}}_t, \widehat{\mathbf{s}}_t) - c_{a}(\boldsymbol{\theta}^0, \mathbf{s}^0) \right]  \leq \frac{8}{\sqrt{\alpha}} \sqrt{T \log T} + 8n\left(\overline{C} - \underline{C} \right)\color{black} (2R_{\max} - 2R_{\min})^n \alpha^{n/2} \color{black}\sqrt{T} \\ + 4n\left(\overline{C} - \underline{C} \right) \log T + B_1   \label{eq:deter-regret1-2}
\end{multline}
Combining the results in (\ref{eq:deter-regret1-1}) and (\ref{eq:deter-regret1-2}) with (\ref{eq:deter-regret1}), we obtain the following upper bound for the first part of our regret bound in (\ref{eq:divide-regret}).
\begin{multline}
    \sum_{t \in \mathcal{T}} \sum_{a \in \mathcal{A}} \left[\pi_{t,a} - c_{a}(\boldsymbol{\theta}^0, \mathbf{s}^0) \right]  \leq {\color{black} m (\overline{C}  - \underline{C}) (1 + \log T)} + \frac{8}{\sqrt{\alpha}} \sqrt{T \log T} \\ + 8n\left(\overline{C} - \underline{C} \right)\color{black} (2R_{\max} - 2R_{\min})^n \alpha^{n/2} \color{black}\sqrt{T} + 4n\left(\overline{C} - \underline{C} \right) \log T + B_1 \label{eq:deter-regret1-final}
\end{multline}
Next, we consider the second part of our regret bound in (\ref{eq:divide-regret}).
\begin{align}
    \sum_{t \in \mathcal{T}} \left[\theta^0_{i(\mathbf{c}(\boldsymbol{\theta}^0, \mathbf{s}^0))} - \theta^0_{i_t(\boldsymbol{\pi}_t)} \right] = \sum_{t \in \mathcal{T}^{\mathrm{xplore}}} \left[\theta^0_{i(\mathbf{c}(\boldsymbol{\theta}^0, \mathbf{s}^0))} - \theta^0_{i_t(\boldsymbol{\pi}_t)} \right]  + \sum_{t \in \mathcal{T}^{\mathrm{xploit}}} \left[\theta^0_{i(\mathbf{c}(\boldsymbol{\theta}^0, \mathbf{s}^0))} - \theta^0_{i_t(\mathbf{c}(\widehat{\boldsymbol{\theta}}_t, \widehat{\mathbf{s}}_t))} \right]  \label{eq:deter-regret2}
\end{align}
We recall that the principal's reward expectations $\theta^0_a$ belong to a known compact set $\Theta$ and define $\mathrm{diam}(\Theta) := \max_{a, a' \in \mathcal{A}} \theta^0_a - \theta^0_{a'}$. As earlier, we consider that $|\mathcal{T}^{\mathrm{xplore}}|$ and $|\mathcal{T}^{\mathrm{xploit}}|$ are random variables, and bound the first term in (\ref{eq:deter-regret2}) by following a similar argument as in (\ref{eq:deter-regret1-0})-(\ref{eq:deter-regret1-1}). 
\begin{align}
     \sum_{t \in \mathcal{T}^{\mathrm{xplore}}} \left[\theta^0_{i(\mathbf{c}(\boldsymbol{\theta}^0, \mathbf{s}^0))} - \theta^0_{i_t(\boldsymbol{\pi}_t)} \right]  \leq \mathrm{diam}(\Theta) \mathbb{E} |\mathcal{T}^{\mathrm{xplore}}| \leq \mathrm{diam}(\Theta) m (1 + \log T)  \label{eq:deter-regret2-1-final}
\end{align}
We continue by deriving the upper bound for the second term in (\ref{eq:deter-regret2}). 
\begin{align}
    &\mathbb{E} \left[ \sum_{t \in \mathcal{T}^{\mathrm{xploit}}} \left[ \theta^0_{i(\mathbf{c}(\boldsymbol{\theta}^0, \mathbf{s}^0))} - \theta^0_{i_t(\mathbf{c}(\widehat{\boldsymbol{\theta}}_t, \widehat{\mathbf{s}}_t))} \right] \Big| \mathcal{T}^{\mathrm{xploit}} \right]  \nonumber \allowdisplaybreaks  \\ 
    &= \sum_{t \in \mathcal{T}^{\mathrm{xploit}}} \mathbb{E} \left[\mu_{t, i(\mathbf{c}(\boldsymbol{\theta}^0, \mathbf{s}^0))} - \mu_{t, i_t(\mathbf{c}(\widehat{\boldsymbol{\theta}}_t, \widehat{\mathbf{s}}_t))} \Big | i(\mathbf{c}(\boldsymbol{\theta}^0, \mathbf{s}^0)) \neq i_t(\mathbf{c}(\widehat{\boldsymbol{\theta}}_t, \widehat{\mathbf{s}}_t)) \right] \mathbb{P}\left(i(\mathbf{c}(\boldsymbol{\theta}^0, \mathbf{s}^0)) \neq i_t(\mathbf{c}(\widehat{\boldsymbol{\theta}}_t, \widehat{\mathbf{s}}_t)) \right)   \allowdisplaybreaks  \\
    &\leq \mathrm{diam}(\Theta) \sum_{t \in \mathcal{T}^{\mathrm{xploit}}} \mathbb{P}\left(i(\mathbf{c}(\boldsymbol{\theta}^0, \mathbf{s}^0)) \neq i_t(\mathbf{c}(\widehat{\boldsymbol{\theta}}_t, \widehat{\mathbf{s}}_t)) \right)   \allowdisplaybreaks  \\
    &\leq \mathrm{diam}(\Theta) \sum_{t \in \mathcal{T}^{\mathrm{xploit}}} \frac{4n}{\eta(1, t)-1} + \color{black}  \frac{2n(2R_{\max} - 2R_{\min})^n \alpha^{n/2} }{[(\eta(1, t)-1) \log (\eta(1, t)-1)]^{n/2}}
    \intertext{which follows by Proposition \ref{prop4regret-2}. By following similar arguments as in (\ref{eq:continue0}) - (\ref{eq:continue2}), we obtain}
    &\leq \color{black} 4 \color{black} n\mathrm{diam}(\Theta)\color{black} (2R_{\max} - 2R_{\min})^n \alpha^{n/2} \color{black} \sqrt{T} + 4n \color{black} \mathrm{diam}(\Theta) \color{black} \log T + B_2
\end{align}
where $B_2 = \frac{2n\mathrm{diam}(\Theta)\color{black} (2R_{\max} - 2R_{\min})^n \alpha^{n/2} \color{black}}{\sqrt{m-2}} + \frac{4n\mathrm{diam}(\Theta)}{m-1}$. We then take the expectation of this result and get
\begin{multline}
    \sum_{t \in \mathcal{T}^{\mathrm{xploit}}} \left[\theta^0_{i(\mathbf{c}(\boldsymbol{\theta}^0, \mathbf{s}^0))} - \theta^0_{i_t(\mathbf{c}(\widehat{\boldsymbol{\theta}}_t, \widehat{\mathbf{s}}_t))} \right] \leq \color{black} 4 \color{black} n\mathrm{diam}(\Theta)\color{black} (2R_{\max} - 2R_{\min})^n \alpha^{n/2} \color{black} \sqrt{T} + 4n \color{black} \mathrm{diam}(\Theta) \color{black} \log T + B_2 \label{eq:deter-regret2-2-final}
\end{multline}
Together (\ref{eq:deter-regret2-1-final}) and (\ref{eq:deter-regret2-2-final}) gives the following upper bound for the second part of our regret.
\begin{multline}
    \sum_{t \in \mathcal{T}} \left[\theta^0_{i(\mathbf{c}(\boldsymbol{\theta}^0, \mathbf{s}^0))} - \theta^0_{i_t(\boldsymbol{\pi}_t)} \right] \leq \mathrm{diam}(\Theta) m (1 + \log T) \\ + \color{black} 4 \color{black} n\mathrm{diam}(\Theta)\color{black} (2R_{\max} - 2R_{\min})^n \alpha^{n/2} \color{black} \sqrt{T} + 4n \color{black} \mathrm{diam}(\Theta) \color{black} \log T + B_2 \label{eq:deter-regret2-final}
\end{multline}
Finally, we join the upper bounds in (\ref{eq:deter-regret1-final}) and (\ref{eq:deter-regret2-final}) to achieve the regret bound presented in Theorem \ref{thm:regret}. 
\begin{align}
     \mathrm{Regret}\left(\Pi_{\epsilon, T} \right) &\leq \frac{8}{\sqrt{\alpha}} \sqrt{T \log T} + \color{black} 4n \color{black} \left(\color{black} 2 \color{black}\overline{C} - \color{black} 2 \color{black} \underline{C} + \mathrm{diam}(\Theta)\right)\color{black} (2R_{\max} - 2R_{\min})^n \alpha^{n/2} \color{black} \sqrt{T} \notag \\
     &\quad + {\color{black} (m + 4n) \left(\overline{C}  - \underline{C} + \mathrm{diam}(\Theta)\right)}\log T \notag \\ 
     &\quad + m\left( {\color{black} (\overline{C}  - \underline{C})} + \mathrm{diam}(\Theta)\right) + B_1 + B_2 \allowdisplaybreaks
\end{align}
\Halmos \endproof

\subsection{Results in Section \ref{sec:agent}} \label{appendix3}

\proof{\textbf{Proof of Proposition \ref{prop:informationrent}.}}
First, recall that in Section \ref{subsec:regret}, we show that if the agent behaves in accordance with their true mean reward vector $\mathbf{s}^0$ and the principal follows the oracle incentive policy $\mathbf{c}(\boldsymbol{\theta}^0, \mathbf{s}^0)$, then the agent gets their minimum possible expected total reward (which is equal to $\max_{a' \in \mathcal{A}} s^0_{a'} + \varsigma$ for a sufficiently small constant $\varsigma > 0$). In this proof, we start by demonstrating this result again by using the agent's optimization problem (\ref{agentproblem}). To recall, the oracle incentive policy first computes the maximum net expected reward that the principal can get from the selection of each action $j \in \mathcal{A}$. This amount was computed as: $\widetilde{V}(j, \mathbf{s}^0;\boldsymbol{\theta}^0)$ = (principal's expected reward from $j$) $-$ (the minimum total incentives to make $j$ agent's reward-maximizer action) = $\theta^0_j - \left(\max_{a' \in \mathcal{A}} s^0_{a'} - s^0_j \right)$. Then, we denoted the action corresponding to the highest of these values as $j^{*, 0} = \argmax_{j \in \mathcal{A}} \widetilde{V}(j, \mathbf{s}^0;\boldsymbol{\theta}^0)$ and the agent's true reward maximizer action as $i(\mathbf{c}(\boldsymbol{\theta}^0, \mathbf{s}^0)) = \argmax_{j \in \mathcal{A}} \left( s^0_j + c_j(\boldsymbol{\theta}^0, \mathbf{s}^0) \right)$. Now, in the optimization problem (\ref{agentproblem}), we let $\mathbf{s} = \mathbf{s}^0$ and $\boldsymbol{\pi} = \mathbf{c}(\boldsymbol{\theta}^0, \mathbf{s}^0)$ where $\mathbf{c}(\boldsymbol{\theta}^0, \mathbf{s}^0)$ is as given in (\ref{eq:oracleincentive1})-(\ref{eq:oracleincentive2}). Then, we have $a = j^{*, 0}$ satisfying the first and second constraints and $b = i(\mathbf{c}(\boldsymbol{\theta}^0, \mathbf{s}^0))$ satisfying the third constraint. As discussed in Section \ref{subsec:regret}, the oracle incentives are designed such that $j^{*, 0} = i(\mathbf{c}(\boldsymbol{\theta}^0, \mathbf{s}^0))$, and thus they also satisfy the last constraint of (\ref{agentproblem}). This shows that $\mathbf{s}^0$ and the oracle incentive policy $\mathbf{c}(\boldsymbol{\theta}^0, \mathbf{s}^0)$ together yield a feasible solution to the agent's optimization problem. Under this feasible solution, the principal's expected net reward is $\theta^0_{j^{*, 0}} - c_{j^{*, 0}}(\boldsymbol{\theta}^0, \mathbf{s}^0) =  \theta^0_{j^{*, 0}} - \max_{a' \in \mathcal{A}} s^0_{a'} + s^0_{j^{*, 0}} - \varsigma$, and the agent's expected total reward (i.e., the value of the objective function) is $s^0_{j^{*, 0}} + c_j^{*, 0}(\boldsymbol{\theta}^0, \mathbf{s}^0) = \max_{a' \in \mathcal{A}} s^0_{a'} + \varsigma$.

Second, we show that there exists a different feasible solution to the agent's optimization problem (\ref{agentproblem}) and that this solution yields a higher profit to the agent than the truthful (and worst-case) solution above. To show this, we need to consider two mutually exclusive cases based on the maximizer actions of the principal and the agent: $\kappa^0 := \argmax_{a' \in \mathcal{A}} s^0_{a'}$ and $q^0 := \argmax_{a' \in \mathcal{A}} \theta^0_{a'}$.

\underline{\textit{Case 1:}} $\kappa^0 = q^0$. In this case, notice that the principal does not need to incentivize the agent at all to get them pick the desired action, and thus we have $j^{*, 0} = q^0$. However, the agent can pretend that they have a different reward vector whose reward-maximizer action is different than $j^{*, 0}$. This way, the agent can oblige the principal to offer them positive incentives for selecting $j^{*, 0}$. Let $\overline{q}^0 = \argmax_{a' \in \mathcal{A} \setminus \{q^0\}} \theta^0_{a'}$ be the action associated with the second highest true mean reward of the principal. We define the quantity $Q_1:= \theta^0_{q^0} - \theta^0_{\overline{q}^0}$. Then, we consider the solution $(\mathbf{s}, \boldsymbol{\pi})$ where $\mathbf{s}$ is such that $s_{\overline{q}^0} = s^0_{q^0} + Q_1 - 2\varsigma$ and $s_j = s^0_j, \ \forall j \neq \overline{q}^0$ and $\boldsymbol{\pi}$ is such that $\pi_{q^0} = Q_1 - \varsigma$ and $\pi_j = 0, \ \forall j \neq q^0$, for a sufficiently small constant $\varsigma > 0$. For $a = b = {q^0}$, this solution is feasible to the agent's problem (\ref{agentproblem}) and yields an expected net reward $\theta^0_{q^0} - \pi_{q^0} = \theta^0_{q^0} - Q_1 + \varsigma$ to the principal and an expected total reward $s^0_{q^0} + \pi_{q^0}= \max_{a' \in \mathcal{A}} s^0_{a'} + Q_1 - \varsigma$ to the agent (which is the value of the objective function for this solution). This result shows that the agent can increase their expected total reward by using the considered reward vector $\mathbf{s}$ and extracting an extra amount of $Q_1 - 2\varsigma$ from the principal.  

\underline{\textit{Case 2:}} $\kappa^0 \neq q^0$. Now, we define a new quantity corresponding to the difference between the highest and second highest $\widetilde{V}(j, \mathbf{s}^0;\boldsymbol{\theta}^0)$. Let this quantity be $Q_2 := \widetilde{V}(j^{*, 0}, \mathbf{s}^0;\boldsymbol{\theta}^0) - \max_{j \in \mathcal{A} \setminus \{j^{*, 0}\}} \widetilde{V}(j, \mathbf{s}^0;\boldsymbol{\theta}^0)$. Then, we consider the solution $(\mathbf{s}, \boldsymbol{\pi})$ where $\mathbf{s}$ is such that $s_{\kappa^0} = s_{\kappa^0} + Q_2 - 2\varsigma$ and $s_a = s^0_a, \ \forall a \neq \kappa^0$ and $\boldsymbol{\pi}$ is such that $ \pi_{j^{*, 0}} = s^0_{\kappa^0} - s^0_{j^{*, 0}} + Q_2 - \varsigma$ and $\pi_a = 0, \ \forall a \neq j^{*, 0}$ for a sufficiently small constant $\varsigma > 0$. Note that this solution is feasible to the agent's problem (\ref{agentproblem}) for $a = b = j^{*, 0}$. Then, the principal's expected net reward becomes $\theta^0_{j^{*, 0}} - \pi_{j^{*, 0}} = \theta^0_{j^{*, 0}} - \max_{a \in \mathcal{A}} s^0_a + s^0_{j^{*, 0}} - Q_2 + \varsigma$ and the agent's expected total reward (i.e., the value of the objective function) becomes $s^0_{j^{*, 0}} + \pi_{j^{*, 0}} = \max_{a \in \mathcal{A}} s^0_a + Q_2 - \varsigma$. In other words, there is a feasible solution of (\ref{agentproblem}) that increases the agent's expected reward (and decreases the principal's expected net reward) by $Q_2 - 2\varsigma$ as compared to the worst-case solution above. 

These example solutions prove that the optimization problem given in (\ref{agentproblem}) is feasible and designed to maximize the agent's information rent by the use of an untrue mean reward vector. 
\ \endproof

\color{black}
\section{Data Calibration for Numerical Experiments} \label{appendix4}

\subsection{Calibration of Provider's Expected Rewards}\label{appendix4-1}
Recall that we model provider's costs for each adherence arm $a$ using a zero-truncated Normal distribution, $\mathcal{N}_{+} (\theta^0_a, (\sigma^{\text{provider}}_a)^2)$. We calibrate these two parameters using Table 2 of \cite{pittman2011adherence}, which reports cost savings by adherence level. 

We follow:
\begin{enumerate}
    \item[\textbf{i.}] First, their three adherence bins are mapped to our four arms by splitting the $<60\%$ bin into distinct NON and LOW categories. To capture the diminishing marginal savings observed in their data, we interpolate costs by fitting a quadratic curve $x a^2+ y a + z$ through their three points $(0.30,11102)$, $(0.70,10609)$, $(0.90,10198)$. We then evaluate the fitted curve at $a=\delta_{\text{NON}}= 0.05$ and $a=\delta_{\text{LOW}} = 0.25$ as defined in Section \ref{sec:numerical}, and lastly make conversion to weekly means by dividing by 52.14. 
    \item[\textbf{ii.}] To set the standard deviations, we use the relation $\sigma^{\text{provider}}_a = CV \times \theta^0_a$ by imposing a common coefficient of variation, $\mathrm{CV}=0.30 / \text{month} = 0.625 / \text{week}$, so truncation at zero has negligible effect on moments. 
\end{enumerate}
The resulting provider parameters are reported in Table \ref{tab:providercosts}. 
\begin{table}[h!]
    \centering
    \begin{tabular}{| c | c | c | c | c |}
    \hline
                   a  & NON     & LOW      & MED     & HIGH \\ \hline
      $\theta^0_a (\$)$ & 214.564  & 213.517 & 203.471  &  195.588 \\ \hline
      $\sigma^{\text{provider}}_a (\$)$   & 134.173 &  133.521 & 127.241  &  122.309 \\ 
    \hline 
    \end{tabular}
    \caption{\color{black} Provider's true distributional parameters by adherence arm.}
    \label{tab:providercosts}
\end{table}

\subsection{Calibration of Patient's Expected Rewards}\label{appendix4-2}

Recall that the patient's expected reward at an adherence level $a \in \mathcal{A}$ is given as 
\begin{equation}
    r_{a} =  \mathrm{WTP} \times \mathbb{E}q_t \times \delta(a) - k^\mathrm{DIS} \times  \delta^2(a). \label{eq:patientreward}
\end{equation}

To calibrate the data from \cite{mason2012optimizing} to our model, we suppose that a single patient's $\mathbb{E}q_t$ is drawn from $\mathbb{E}q_t \sim \mathcal{N}(\mu^{\text{pop}}, (\sigma^{\text{pop}})^2)$, representing the population distribution of statin adherence perceptions.

We then conduct the following analysis: 
\begin{enumerate}
    \item[\textbf{i.}] For each pair of adjacent adherence levels $(\delta_a,\delta_{a+1})$, we calculate the cutoff value $\text{Cut}_j = (k^{\mathrm{DIS}}/\mathrm{WTP})(\delta_a+\delta_{a+1})$ where indifference occurs since their expected rewards ($r_{a}$ and $r_{a+1}$) are equal. With the chosen $\mathrm{WTP} = 10^5$ \citep{evans2004use, mason2012optimizing, lobo2017using} and $k^{\mathrm{DIS}} = 4.07$ values, the resulting cutoffs are $\{0.000636,\ 0.001802,\ 0.003180\}$. 
 
    Intuitively, these thresholds partition the support of $\mathbb{E}q_t$ such that: patients with $\mathbb{E}q_t$ below the first cutoff are predicted to choose the arm NON, those between the first and second choose LOW, those between the second and third choose MED, and those above the third choose HIGH. Thus, the stationary adherence shares reported in Table 5 of \cite{mason2012optimizing} (45.8\% NON, 11.0\% LOW, 17.9\% MED, 25.4\% HIGH) correspond exactly to the probabilities that the Normal random variable $\mathbb{E}q_t$ falls into these four intervals. 
    \item[\textbf{ii.}] Letting $P_j$ denote the cumulative shares and $z_j=\Phi^{-1}(P_j)$, we then need to solve these three equations: $\mu+\sigma z_j = \text{Cut}_j$ for $j=1,2,3$. Because this system is over-identified (three equations, two unknowns), we estimate $(\mu^{\text{pop}},\sigma^{\text{pop}})$ by ordinary least squares, i.e., by minimizing $\sum_{j=1}^3\big[\text{Cut}_j-(\mu^{\text{pop}}+\sigma^{\text{pop}} z_j)\big]^2$. This procedure yields $\mu^{\text{pop}} = 2.0808 \approx 10^{-5}$ and $\sigma^{\text{pop}} \approx 1.29817 \times 10^{-4}$.
\end{enumerate} 

Lastly, as a sanity check, we plug these values back into the Normal cdf and obtain the calibrated cumulative shares of 44.5$\%$, 58.7$\%$, and 74.1$\%$, which are sufficiently close to the observed values of 45.8$\%$, 56.8$\%$, and 74.6$\%$ from \cite{mason2012optimizing}.

\subsection{Other Model Parameters}\label{appendix4-3}

For our experiments, we generate a sequence of 10 iid patients whose $\mathbb{E}q_t$ is drawn from the population distribution $\mathcal{N}(\mu^{\text{pop}}, (\sigma^{\text{pop}})^2)$ calibrated above. We then calculate the expected reward for each patient based on (\ref{eq:patientreward}) and present in Table \ref{tab:agentrewards}. 

Based on these generated values, we then compute the remaining parameters as follows: $R_\mathrm{min} = -35$, $R_\mathrm{max} = 22$, $\gamma = R_\mathrm{max} - R_\mathrm{min} - 1 = 56$, and accordingly, the range of provider's incentives is given by $\mathcal{C} = [-35, 78]$. 

\begin{table}[h!]
    \centering
    \begin{tabular}{| c | c |}
    \hline
        Patient & $\mathbf{r}^0 = (r^0_{\text{NON}}, r^0_{\text{LOW}}, r^0_{\text{MED}}, r^0_{\text{HIGH}})$ \\ \hline
      1 & (0, 2, 4, 5)  \\ \hline
      2 & (0, -3, -10, -16)\\  \hline
      3 & (0, 1, 3, 4)  \\  \hline
      4 & (1, 6, 1, 19) \\  \hline
      5 & (0, 0, -1, -3) \\\hline
      6 & (1, 6, 15, 22)  \\ \hline
      7 & (0, 0, 1, 0)\\  \hline
      8 & (-1, -9, -23, -35)  \\  \hline
      9 & (1, 6, 14, 21 \\  \hline
      10 & (0, -1, -4, -7) \\
    \hline 
    \end{tabular}
    \caption{\color{black} The true expected reward vectors for the generated iid patients.}
    \label{tab:agentrewards}
\end{table}




\end{APPENDICES}


\end{document}